\documentclass[a4paper,fleqn]{cas-dc}

\usepackage[numbers]{natbib} 
\usepackage{graphicx} 
\usepackage{todonotes}
\usepackage{algorithm}
\usepackage{algorithmic}
\usepackage{amssymb}
\usepackage{amsmath}
\usepackage{svg}
\usepackage{subcaption}
\usepackage{adjustbox}
\usepackage{array} 
\usepackage{arydshln}
\usepackage{appendix}
\usepackage{float}
\usepackage{tabularx} 
\usepackage{setspace}
\usepackage{placeins}

\usepackage{lineno}
\usepackage{hyperref}

\def\tsc#1{\csdef{#1}{\textsc{\lowercase{#1}}\xspace}}
\tsc{WGM}
\tsc{QE}
\makeatletter
\def\printPreprint#1{}
\def\@printorcidline#1{}
\makeatother
\begin{document}

\let\WriteBookmarks\relax
\onehalfspacing

\shorttitle{Wheat Spike Volume Estimation}    

\shortauthors{}  
\title [mode = title]{Fine-Tuned Vision Transformers Capture Complex Wheat Spike Morphology for Volume Estimation from RGB Images}  

\author[1]{Olivia Zumsteg}
\cormark[1]
\ead{olivia.zumsteg@usys.ethz.ch}
\credit{Writing – review \& editing, Writing – original draft, Visualisation, Validation, Software, Methodology.}

\author[2]{Nico Graf}
\credit{Writing – review \& editing, Visualisation, Software, Methodology.}

\author[3]{Aaron Häusler}
\fnmark[1]
\credit{Writing – review \& editing, Visualisation, Software, Methodology.}

\author[1]{Norbert Kirchgessner}
\credit{Writing – review \& editing, Software, Methodology.}

\author[1]{Nicola Storni}
\credit{Writing – review \& editing, Software, Methodology.}

\author[1]{Lukas Roth}
\credit{Writing – review \& editing, Methodology, Supervision.}

\author[1]{Andreas Hund}
\credit{Writing – review \& editing, Methodology, Supervision, Conceptualisation, Funding acquisition}

\affiliation[1]{organization={ETH Zurich, Institute of Agricultural Science},
            addressline={Universitätsstrasse 2}, 
            city={Zurich},
            postcode={8092}, 
            country={Switzerland}}

\affiliation[2]{organization={ETH Zurich, Department of Computer Science},
            addressline={Universitätsstrasse 6}, 
            city={Zurich},
            postcode={8902}, 
            country={Switzerland}}

\affiliation[3]{organization={ETH Zurich, Department of Mechanical and Process Engineering},
            addressline={Leonhardstrasse 21}, 
            city={Zurich},
            postcode={8092}, 
            country={Switzerland}}

\cortext[1]{Corresponding author}

\begin{abstract}
 Estimating three-dimensional morphological traits such as volume from two-dimensional RGB images presents inherent challenges due to the loss of depth information, projection distortions, and occlusions under field conditions. In this work, we explore multiple approaches for non-destructive volume estimation of wheat spikes using RGB images and structured-light 3D scans as ground truth references. Wheat spike volume is promising for phenotyping as it shows high correlation with spike dry weight, a key component of fruiting efficiency. Accounting for the complex geometry of the spikes, we compare different neural network approaches for volume estimation from 2D images and benchmark them against two conventional baselines: a 2D area-based projection and a geometric reconstruction using axis-aligned cross-sections. Fine-tuned Vision Transformers (DINOv2 and DINOv3) with MLPs achieve the lowest MAPE of 5.08\% and 4.67\% and the highest correlation of 0.96 and 0.97 on six-view indoor images, outperforming fine-tuned CNNs (ResNet18 and ResNet50), wheat-specific backbones, and both baselines. When using frozen DINO backbones, deep-supervised LSTMs outperform MLPs, whereas after fine-tuning, improved high-level representations allow simple MLPs to outperform LSTMs. We demonstrate that object shape significantly impacts volume estimation accuracy, with irregular geometries such as wheat spikes posing greater challenges for geometric methods than for deep learning approaches. Fine-tuning DINOv3 on field-based single side-view images yields a MAPE of 8.39\% and a correlation of 0.90, providing a novel pipeline and a fast, accurate, and non-destructive approach for wheat spike volume phenotyping. 
\end{abstract}




\maketitle

\section{Introduction}
Accurate volume estimation plays a critical role across a range of agricultural applications. In the context of food production and processing, volume serves as a non-destructive proxy for various biological traits, enabling tasks such as monitoring fruit growth, plant phenotyping, post-harvest handling, size-based sorting, and quality grading of horticultural products \citep{monVisionBasedVolume2020, suPotatoFeaturePrediction2017, kocDeterminationWatermelonVolume2007, moredaNondestructiveTechnologiesFruit2009}. Consequently, volume estimation has been addressed using 2D images, 3D sensing technologies, geometric reconstruction methods, and deep learning approaches. 

In 2D computer vision systems, volume is commonly estimated by extracting geometric features, such as area and length, or by fitting shapes (cylinders, spheres, or ellipsoids) to approximate volume \citep{moredaNondestructiveTechnologiesFruit2009}. These approaches have been applied to various agricultural products, including tomatoes \citep{uluisikImageProcessingBased2018}, oranges \citep{khojastehnazhandDeterminationOrangeVolume}, lemons \citep{vivekvenkateshEstimationVolumeMass2015}, and different citrus fruits \citep{omidEstimatingVolumeMass2010} on multiple images, or on apples \citep{iqbalVolumeEstimationApple2011}, and eggs \citep{soltaniEggVolumePrediction2015} on single images. While such approaches are effective for regularly shaped objects, they might perform poorly on irregular or highly non-uniform shapes, where simple geometric approximations introduce significant error.

3D computer vision methods overcome some of these issues by reconstructing depth from multiple views or by generating point clouds. 3D reconstruction methods have been widely used for food volume estimation, including methods based on idealized geometric shapes \citep{xuImagebasedFoodVolume2013}, stereo reconstruction \citep{dehaisTwoView3DReconstruction2017}, or depth maps from single RGB images \citep{graikosSingleImageBasedFood2020}. These methods typically require accurate determination of intrinsic and extrinsic camera parameters. Beyond passive reconstruction, depth information can also be obtained using active sensing technologies. Structured light systems compute depth by projecting patterns onto the scene and applying triangulation using the known geometric relationship between the projector and the camera \citep{rachakondaSourcesErrorsStructured2019}. Similarly, Time-of-Flight (ToF) cameras estimate per-pixel depth by emitting IR (Infra-Red) light and measuring the return time of the reflected signal \citep{hansardTimeofFlightCamerasPrinciples2013}. However, such depth cameras perform poorly under strong illumination conditions, as the IR components of natural light can interfere with the sensors, limiting their applicability in outdoor field environments \citep{haiderWhatCanWe2022}. In contrast, LiDAR (Light Detection and Ranging) systems use laser pulses to generate point clouds of plant surfaces, enabling the estimation of 3D traits such as tree volume \citep{rosellpoloTractormountedScanningLIDAR2009} and above-ground biomass of wheat \citep{jimenez-berniHighThroughputDetermination2018}. However, wind-induced plant motion can introduce noise and reduce the accuracy of the 3D reconstruction \citep{dassotUseTerrestrialLiDAR2011}. 

 Given the complexity, acquisition time, and environmental sensitivity of 3D reconstruction methods, machine learning approaches offer an alternative by learning complex and non-linear relationships directly between input images and output traits \citep{omahonyRealtimeMonitoringPowder2017}.
 In recent years, deep learning has been widely adopted for computer vision tasks such as image colourisation, object detection, semantic segmentation, and classification \citep{omahonyDeepLearningVs2020}. With the availability of GPUs, machine learning libraries and large-scale training datasets, deep neural networks have substantially improved performance across these tasks since their revival in 2012 by \citep{NIPS2012_c399862d}. Hybrid approaches that combine segmentation and deep learning-based view synthesis have been proposed for 3D reconstruction to estimate food volume from RGB and depth images \citep{loFoodVolumeEstimation2018}. However, as discussed above, the reliance on depth sensors limits the applicability of this approach. Consequently, image-based models such as convolutional neural networks (CNNs) and long short-term memory (LSTM) networks have been proposed to estimate fruit or vegetable volume directly from monocular images or video sequences, without requiring camera calibrations or explicit depth information \citep{steinbrenerLearningMetricVolume2023b, liDeepVolDeepFruit2018}. 

Within crop research, wheat spikes represent a particularly relevant plant organ for structural and volumetric analysis. The wheat inflorescence (spike or head) consists of short side branches, called spikelets, that are attached alternately along a central main axis (rachis), and terminate in a single apical spikelet \citep{gaoArchitectureWheatInflorescence2019}. As a result, the wheat spike exhibits a bilateral symmetry, with a dorsiventral view in which most bracts and bract-like structures enclosing the florets are visible, and a lateral view in which the rachis and a subset of bracts are exposed \citep{lerstenMorphologyAnatomyWheat1987}. In recent literature, these dorsiventral and lateral views are commonly referred to as frontal and side views, respectively \citep{niuNovelMethodWheat2024}. 
Thus, the overall volume of the wheat spike is geometrically complex, owing to the irregular arrangement and shape of the spikelets and their associated floral structures. This irregular complexity suggests that volume estimation based solely on geometric feature extraction may be insufficient for accurate volume estimation, motivating the use of more expressive modelling approaches. 

Fruiting efficiency is the final outcome of floret development and is defined as the number of grains set per unit of spike dry weight at flowering. It reflects the overall efficiency with which a genotype allocates resources to grain set \citep{slaferFruitingEfficiencyAlternative2015}, and has been proposed as a promising trait for improving yield potential \citep{pretiniPhysiologyGeneticsFruiting2021}. We hypothesise that spike dry weight and its volume at flowering are closely correlated, allowing for a non-destructive quantification of spike dry weight at flowering. Building on this, we propose to term the total volume of spikes in a canopy at flowering, i.e., the summed volume per area of all individual spikes as "fruiting capacity". Environmental stresses encountered during spike development, such as reduced radiation, have been shown to reduce spike volume and, consequently, the fruiting capacity \citep{andereggThermalImagingCan2024}. Hence, scalable, image-based estimation of spike volume might facilitate the identification of high-performing genotypes with improved resilience to climate-induced stressors, enabling to dissect yield into its components.

Understanding the relationship between spike phenotypes and wheat yield is essential to select wheat genotypes. However, efficient spike phenotyping methods are limited, particularly in-field where light conditions are complex. Individual spikes have been segmented from field-collected LiDAR data, and spike length estimates were highly accurate when compared with LiDAR measurements ($R^2$ = 0.99)  \citep{liuExtractionWheatSpike2023}. However, spike volume constructed from the point cluster of the spike was not evaluated with manual measurements. Structured light scanners were used on a mobile field platform to generate 3D point clouds of wheat plants in the field \citep{wangUnsupervisedAutomaticMeasurement2022}. Spike volume was estimated using small cuboids adapted to the shape of the spikes. However, accurately capturing the detailed structure of the spikes requires a large number of cuboids, which increases computational cost and processing time. In addition, spike length, width and volume have been estimated for high-throughput phenotyping from multi-view images using 3D Gaussian Splatting \citep{zhangWheat3DGSInfield3D2025}. Although high accuracies were achieved for spike length and width, volume estimation resulted in a mean absolute percentage error (MAPE) of over 40\%.

\begin{figure*}[htbp]
    \includegraphics[width=\linewidth]{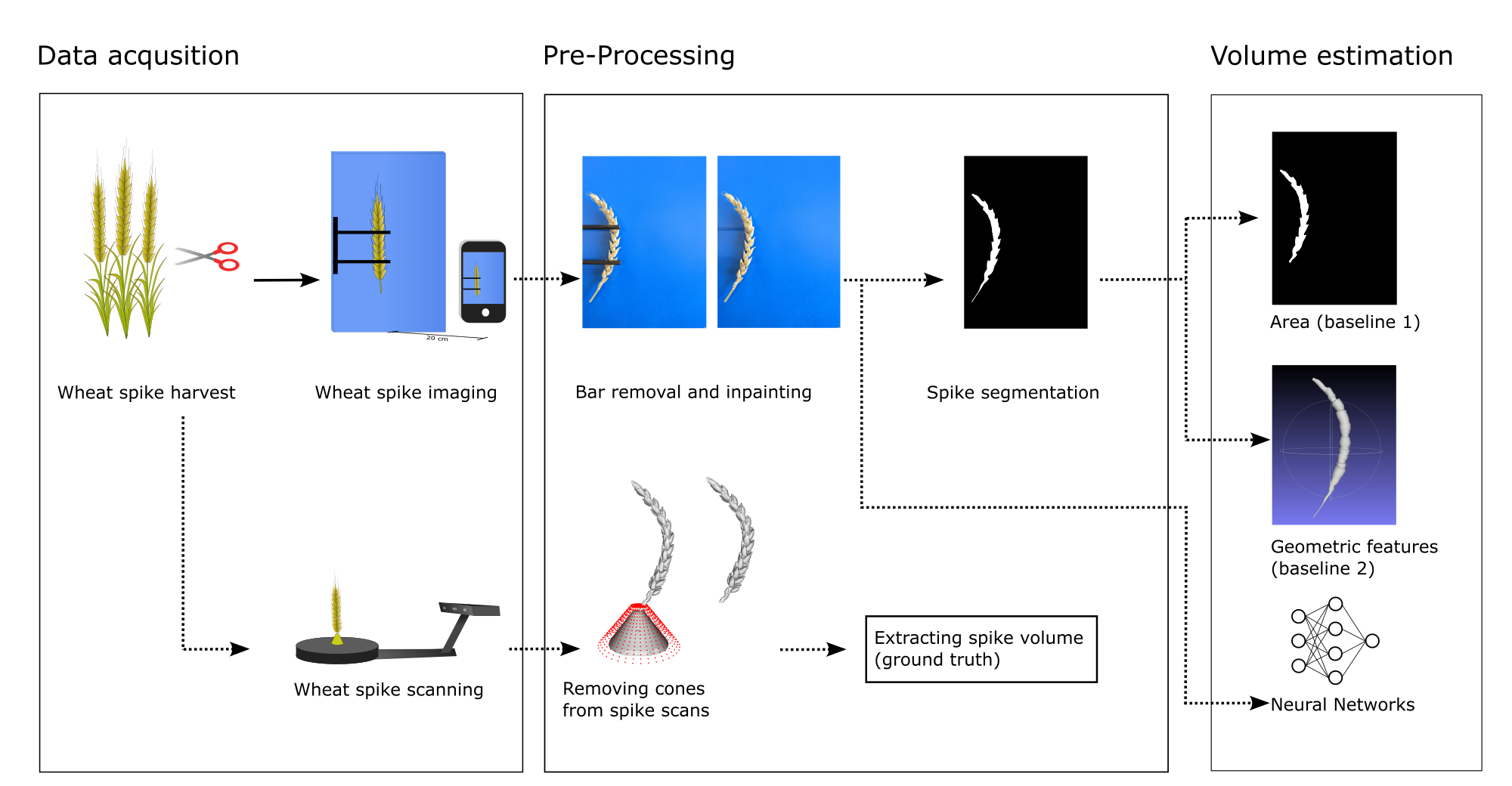}
    \caption{Overview of the study design separated in three sections data acquisition, pre-processing, and volume estimation. In the data acquisition step, wheat spikes were sampled, imaged, and scanned. Images were pre-processed to get the binary segmentations masks. The scans were processed and the ground truth volume was extracted. Two baselines and neural networks were used to estimate wheat spike volume based on images.}
    \label{fig:overview}
\end{figure*}

 Building on these considerations, we constructed a comprehensive, high-quality dataset comprising over 1,700 wheat spikes from more than 90 genotypes, captured across multiple developmental stages and from several view points per spike. This dataset is complemented by 3D ground truth volume measurements obtained via structured-light scanning. To address the challenge of precise measurements, we propose a novel pipeline for accurate spike volume estimation from RGB images acquired under both indoors and field conditions. The pipeline integrates image inpainting \citep{yuFreeFormImageInpainting2019}, state-of-the-art backbone feature extraction, targeted fine-tuning, and downstream regression models. To address the challenge of fast, high-throughput phenotyping, the proposed approach enables image-based volume estimation without the need for calibrated images or depth estimation. The pipeline relies on close-up smartphone images captured at a fixed distance of 20\,cm from the spike against a blue background. This configuration yields clean spike views after inpainting the fixing bars (Figure~\ref{fig:overview}). We evaluate the proposed pipeline by comparing multiple state-of-the-art neural network architectures against conventional algorithms, explicitly analysing the influence of object shape on model performance. In addition, we investigate the effect of varying the number of views on estimation accuracy, assessing both robustness under limited visual information and the ability of different models to exploit additional views. Finally, we investigate model adaptability to field conditions by fine-tuning on field images, enabling accurate and non-destructive spike volume phenotyping.   

The subsequent sections of this paper are organised as follows. Section 2 provides an in-depth description of the proposed spike volume estimation pipeline. Section 3 details the comparative analysis of the algorithms and models. Section 4 discusses the accuracy of the different volume estimation methods and analyses the impact of image count on estimation performance. Section 5 concludes the paper with a summary of the key findings.

\section{Methods}
In the following, we first describe the procedure of acquiring images of the wheat spikes and collecting the corresponding measured volumes. Subsequently, two conventional baseline models are presented to estimated the spike volume from images. The first model is based on the pixel area, while the second utilises geometric features of the spikes. Finally, we introduce our proposed neural networks, which estimate the spike volume directly from images. 

\subsection{Data acquisition and processing}

The experiment was carried out in the field phenotyping platform (FIP, \citep{kirchgessnerETHFieldPhenotyping2016}) at the ETH plant research station Lindau-Eschikon, Switzerland (47.449°N, 8.682°E, 520 m.a.s). Wheat spikes were collected from the trait calibration panel planted in the FIP. This panel consisted of i) historic varieties from Switzerland, France and Germany covering important post-green revolution varieties, ii) varieties included in variety registration in Switzerland, and iii) a diverse set of variety widely tested in the framework of the EU projects INVITE (\citep{INnovationsPlantVarIety, ToolsMethodsExtended}). 
  
    \subsubsection{Wheat spike sampling}
    Wheat spikes are complex-shaped objects that occur in various forms. To create a diverse dataset, spikes of 83 and 82 different genotypes were sampled in 2023 and 2024, respectively, with 72 overlapping genotypes. Furthermore, 120 spikes consisting of two different genotypes from a shading experiment in 2023 were included. Shading tents, mounted before flowering, reduced spike volume by up to -36.5\% \citep{andereggThermalImagingCan2024}, and therefore increased the diversity of the data set. Sampling occurred three times in each season: (i) at flowering (June 9, 2023, June 10, 2024 and June, 2024); (ii) between flowering and maturity (June 29, 2023, July 4, 2023 and July 5, 2024); and (iii) at maturity (July 11, 2023 and July 19, 2024). Furthermore, 100 spikes were randomly sampled at flowering in 2024 to calculate the correlation between spike dry weight and volume at flowering.

    \subsubsection{Wheat spike imaging}
\begin{figure} 
    \centering
    \hspace{0.01\columnwidth}
    \begin{subfigure}[t]{0.22\columnwidth}
        \includegraphics[width=\linewidth]{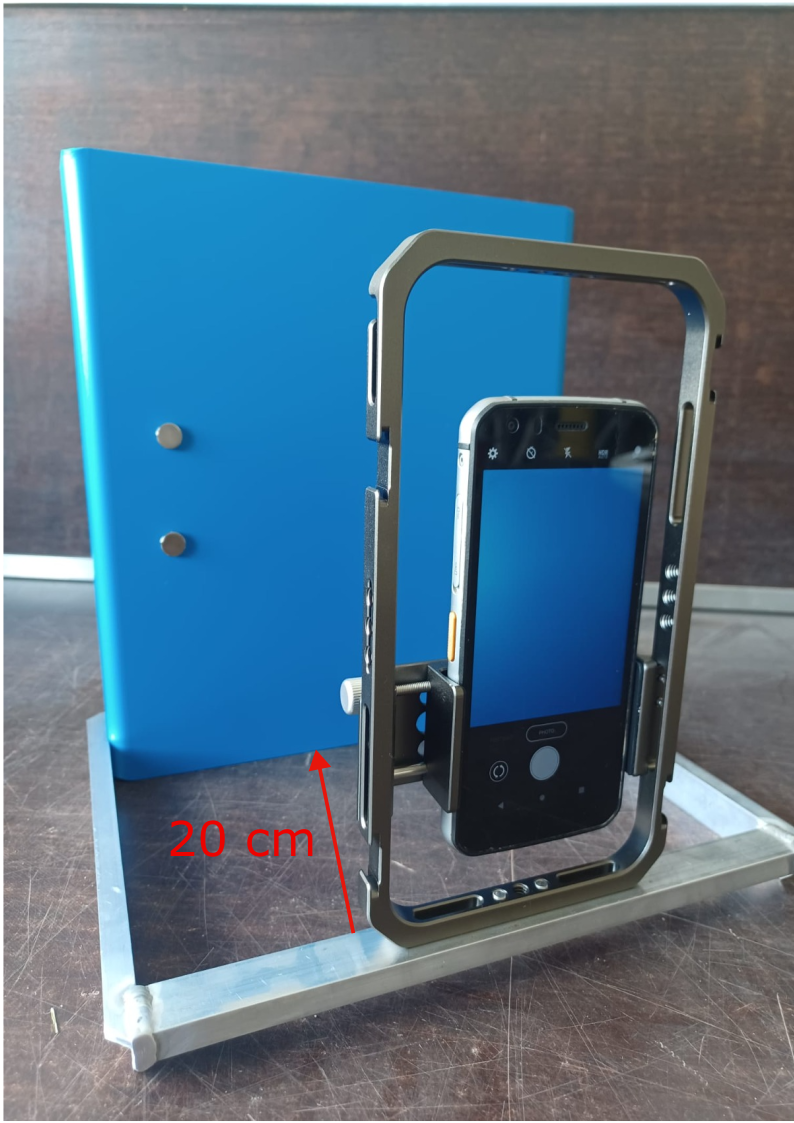}
        \caption{}
    \end{subfigure}
    \hspace{0.01\columnwidth}
    \begin{subfigure}[t]{0.42\columnwidth}
        \includegraphics[width=\linewidth]{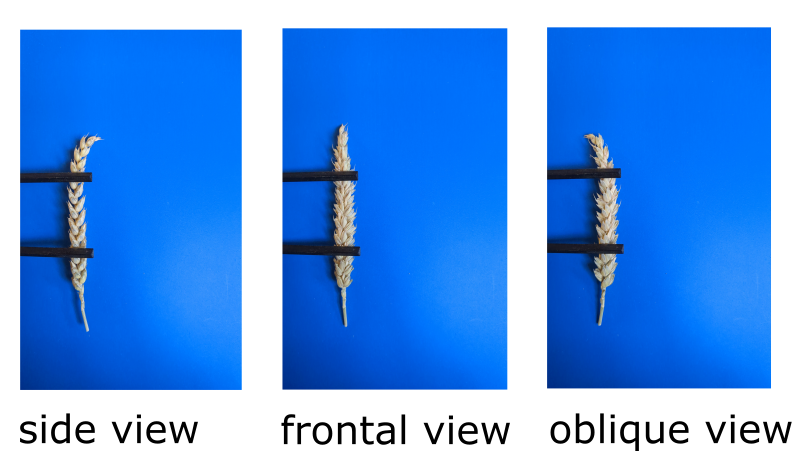}
        \caption{}
    \end{subfigure}
    \hspace{0.01\columnwidth}
    \begin{subfigure}[t]{0.15\columnwidth}
        \includegraphics[width=\linewidth]{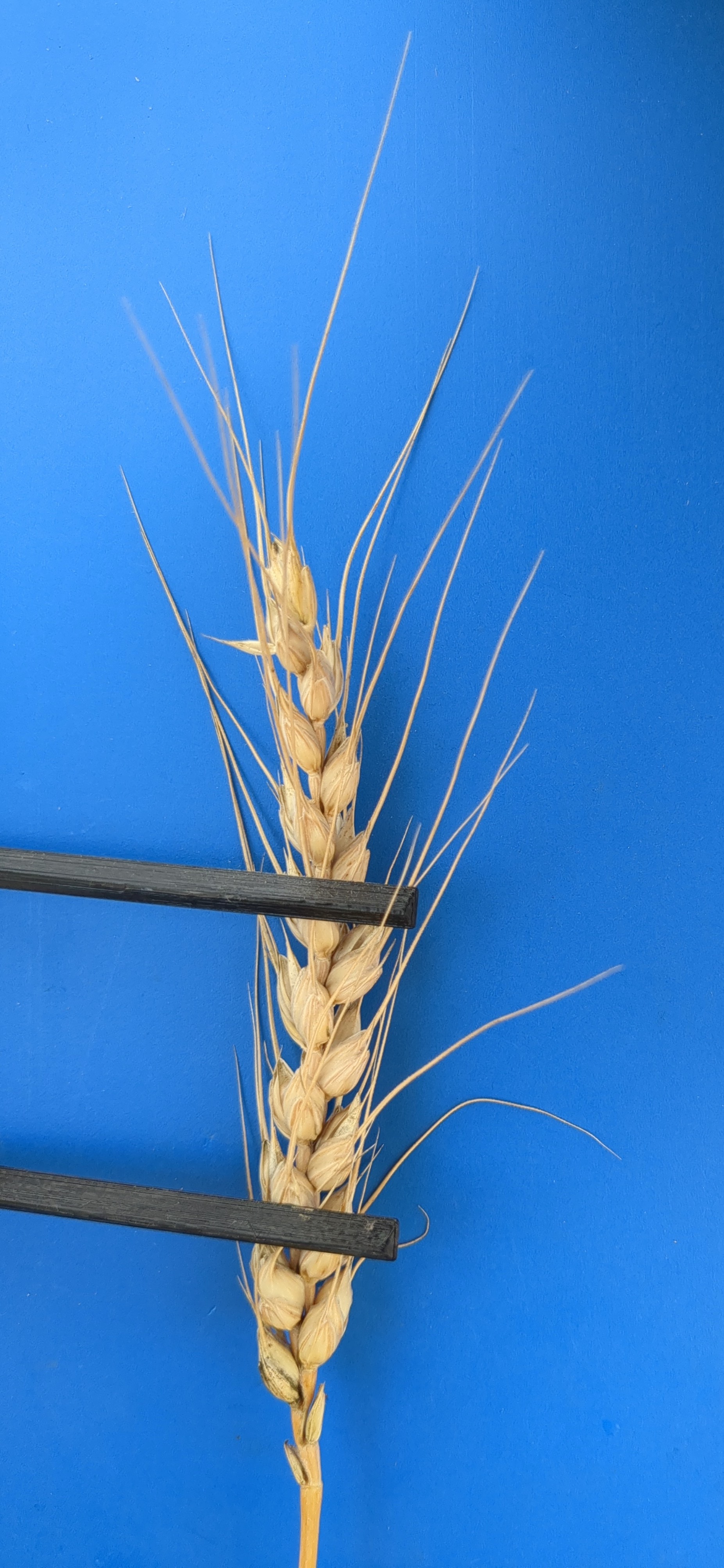}
        \caption{}
    \end{subfigure}
    \caption{Overview of the camera setup and image acquisition process. The imaging setup with a 20 cm distance between the camera lens and a blue background panel (a). Representative images of a side, frontal, and oblique view of the wheat spike captured indoors (b). Representative image of a side view captured in the field (c).}
    \label{fig:combined}
    \end{figure}

    For image acquisition, we built a custom imaging setup consisting of a camera rig (Universal Mobile Phone Cage, SmallRig Technology (HK) Limited, New York, USA) a smartphone with a 12.2 megapixel camera (Google Pixel 6a, Google LLC, USA), with images stored in JPG format, and a blue screen (Kömatex blue 891, Röhm AG, Brüttisellen, Switzerland) with a 3D-printed object holder consisting of two black 3D-printed plastic bars (Figure~\ref{fig:combined}a). The camera rig was fixed at 20 cm distance from the background using aluminium poles, resulting in a ground sampling distance of 0.05 mm/px. Images were collected with the fieldbook app \citep{rifeFieldBookOpenSource2014}.
   
    We created two main image datasets. The indoor dataset consists of six images per spike, including two frontal, two side, and two oblique views (i.e., intermediate between frontal and side view), all acquired under artificial illumination (Figure~\ref{fig:combined}b). Awns were removed prior to imaging and scanning to enable unobstructed rotations of the spike, ensure clear images, and allow for accurate structured-light scans. Images were captured sequentially while rotating around the spike, beginning with a side view and followed by a frontal view, with this pattern repeated before acquiring the oblique views.

    The second imaging dataset was collected outdoors in 2023, where 843 spikes were additionally imaged directly in-field prior to sampling. Due to the presence of awns, each spike was imaged only from the side perspective. The operator pointed the imaging setup in a way to avoid direct sunlight on either the object or lens of the phone to minimise direct shadows and lens flare, respectively (Figure~\ref{fig:combined}c). Awns were removed after imaging and before scanning to ensure accurate structured-light scans.

    \subsubsection{Wheat spike scanning}

    After imaging, objects were scanned using a structured-light 3D scanner (Shining 3D Einscan-SE V2, SHINING3D, Hangzou, China) to approximate the volume and use it as ground-truth volume. The number of turnable steps of the turning table were set to 12. This setting allowed for relatively fast scanning of the objects while maintaining high volume accuracy. The high dynamic range (HDR) was used to help with objects of high contrast. Each object was fixed in a plastic cone to avoid movements during the scanning process. After scanning a spike, the scanner software (EXScan S Version 3.1.4.3 (Windows)) was used to convert the obtained 3D point cloud into a triangular mesh surface in a watertight fashion, which fills all holes directly and with the highest accuracy possible. The meshes were then exported as Polygon File Format (ply). 

    \subsubsection{Pre-processing of images and scans}
    
    The pre-processing phase focused on processing the images and object meshes to generate datasets for establishing baselines and training neural networks. During these processing steps, the black bars were inpainted and the wheat spikes were segmented to produce the baseline image dataset. To approximate the ground-truth volume, the cone was removed and the remaining volume extracted. 

    \subsubsection{Image bar removal and inpainting}

     In a first step, the plastic bars that held the spike in place during imaging were removed. This step aimed to enable generalised volume estimation independent of the specific imaging setup and formed the basis for the area and geometric baselines, and the neural network models. The removal process consisted of two main steps: (i) segmenting the bars; and (ii) inpainting the segmented regions to preserve the visual integrity of the wheat spike (Figure~\ref{fig:preprocessing}, a and b).
     Segmentation was performed using Segment Anything, a robust and adaptable segmentation algorithm \citep{kirillovSegmentAnything2023}, which performs well under varying lighting conditions. We used the official Python API, running the pre-trained Segment Anything Model (SAM) locally. The model weights, available at \url{https://github.com/facebookresearch/segment-anything}, were downloaded and used for inference. For each image, we identified a point on the bar by selecting the darkest pixels within a region where the bar was expected to appear. We determined the darkest pixels by analysing the brightness profile of the leftmost 5\% of the image in the HSV Value channel. Local minima in the vertical brightness profile, indicating dark regions, were extracted using peak detection on the inverted signal. The corresponding y-coordinates of these minima were paired with a fixed x-coordinate of half the width of the analysed strip. The binary bar masks generated by SAM, along with the corresponding RGB images, were then passed to the inpainting algorithm developed by \citep{yuFreeFormImageInpainting2019}, which reconstructs the occluded image regions in a visually coherent manner. The inpainting model is accessible at \url{https://github.com/JiahuiYu/generative_inpainting}.

    \subsubsection{Spike segmentation}
    \label{subsubsec:segmentation}
    
\begin{figure} 
    \centering
    \begin{subfigure}[htbp]{0.27\columnwidth}
        \centering
        \includegraphics[width=0.5\linewidth]{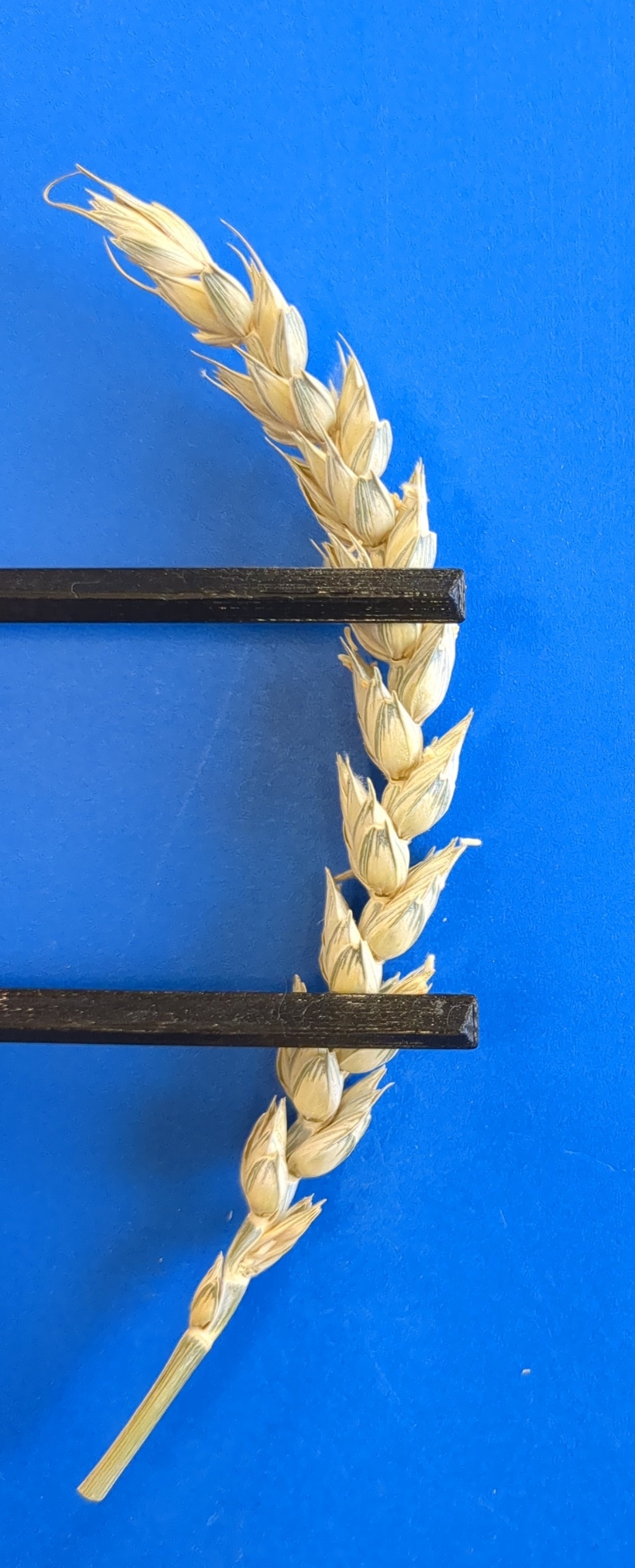}
        \caption{}
    \end{subfigure}
    \hfill
    \begin{subfigure}[htbp]{0.27\columnwidth}
        \centering
        \includegraphics[width=0.5\linewidth]{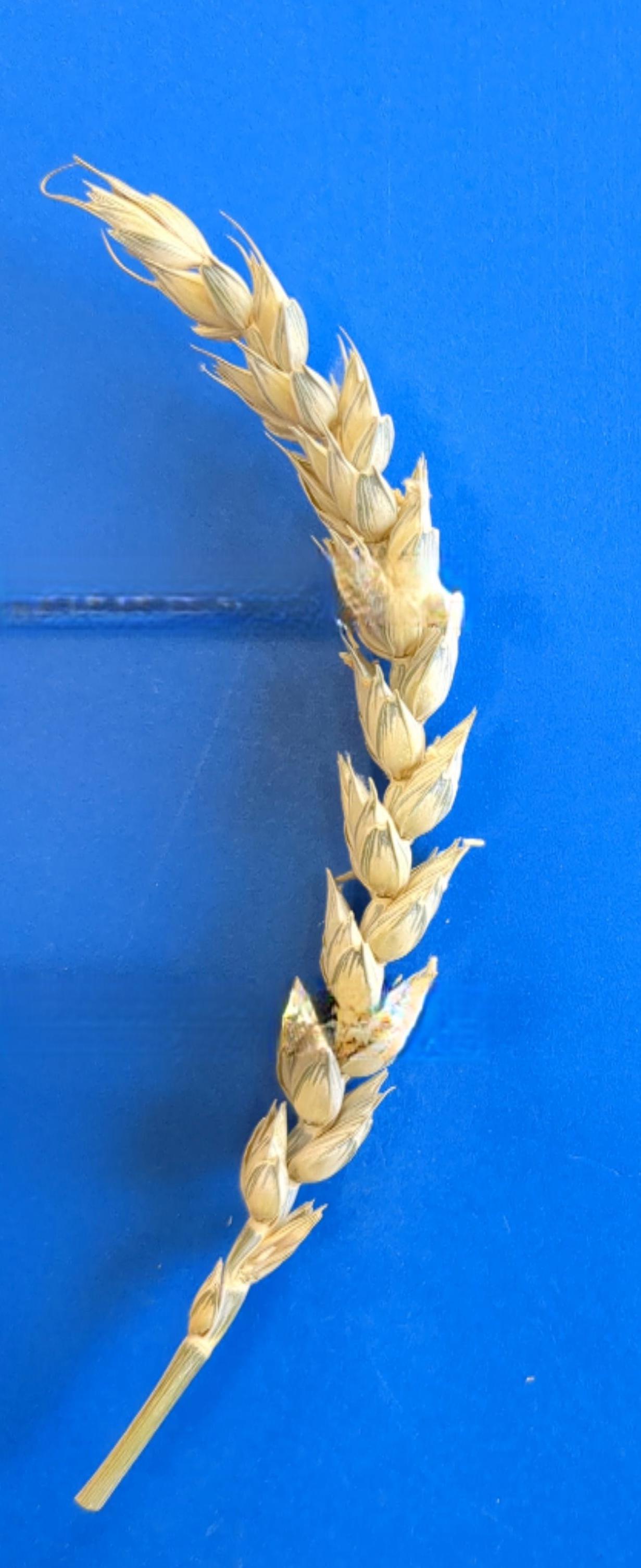}
        \caption{}
    \end{subfigure}
    \hfill
    \begin{subfigure}[htbp]{0.27\columnwidth}
        \centering
        \includegraphics[width=0.5\linewidth]{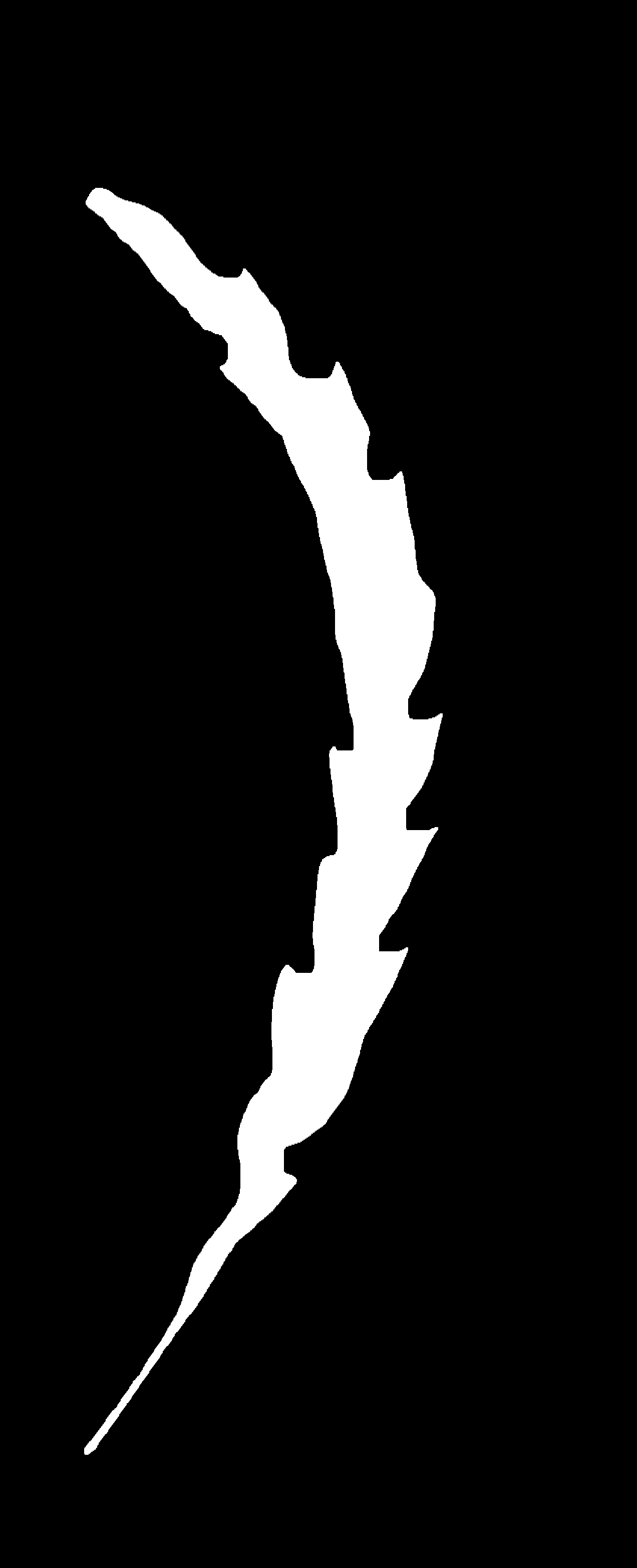}
        \caption{}
    \end{subfigure}
    \caption{Overview of preprocessing steps, including an original image (a), after bar removal and inpainting (b), and after segmentation (c).}
    \label{fig:preprocessing}
    \end{figure}

    To compute the area and geometric baselines, full-spike segmentations were performed after removal of the fixation bars, again using SAM \citep{kirillovSegmentAnything2023}. As in the bar segmentation process, the initial step required identification of the target object (i.e. the spike) within the RGB image. To identify a seed point on the spike, we converted the image to HSV colour space and applied a colour-based mask to isolate regions with hue values corresponding to orange tones $(\text{hue} \in [10, 25])$. This mask was further refined by retaining only pixels with high saturation and brightness values. The coordinates of the identified seed point were supplied for object segmentation. The resulting binary masks delineating the spike regions were used for the geometric and area baseline calculation (Figure~\ref{fig:preprocessing}c). 
    
    \subsubsection{Volume extraction as ground truth}
\begin{figure} 
        \centering
        \begin{subfigure}[htbp]{0.45\columnwidth}
            \centering
            \includegraphics[width=0.5\linewidth]{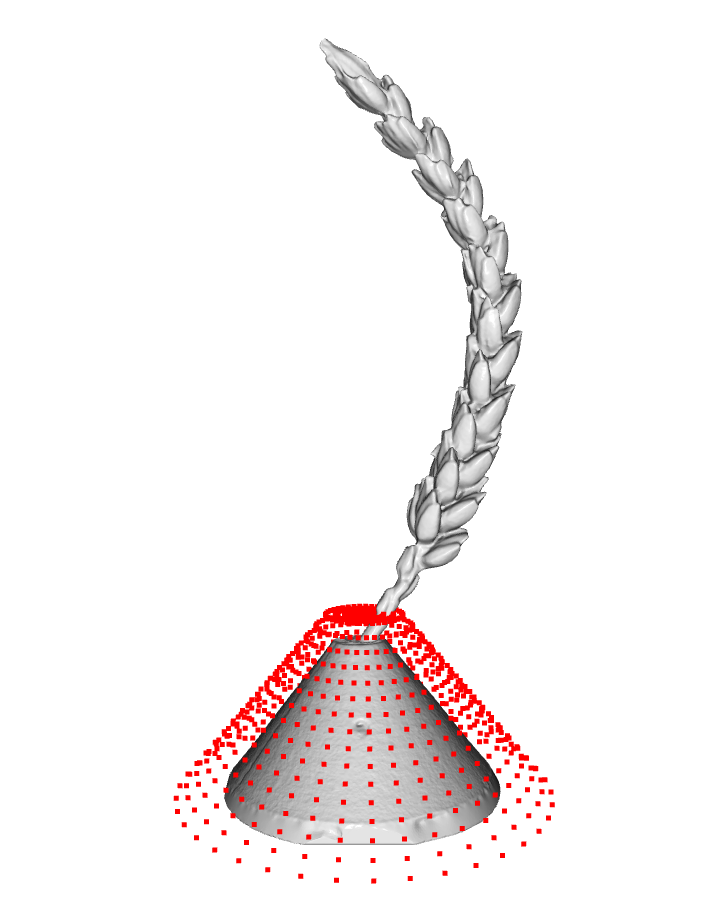}
            \caption{}
        \end{subfigure}
        \hspace{0.02\columnwidth}
        \begin{subfigure}[htbp]{0.45\columnwidth}
            \centering
            \includegraphics[width=0.5\linewidth]{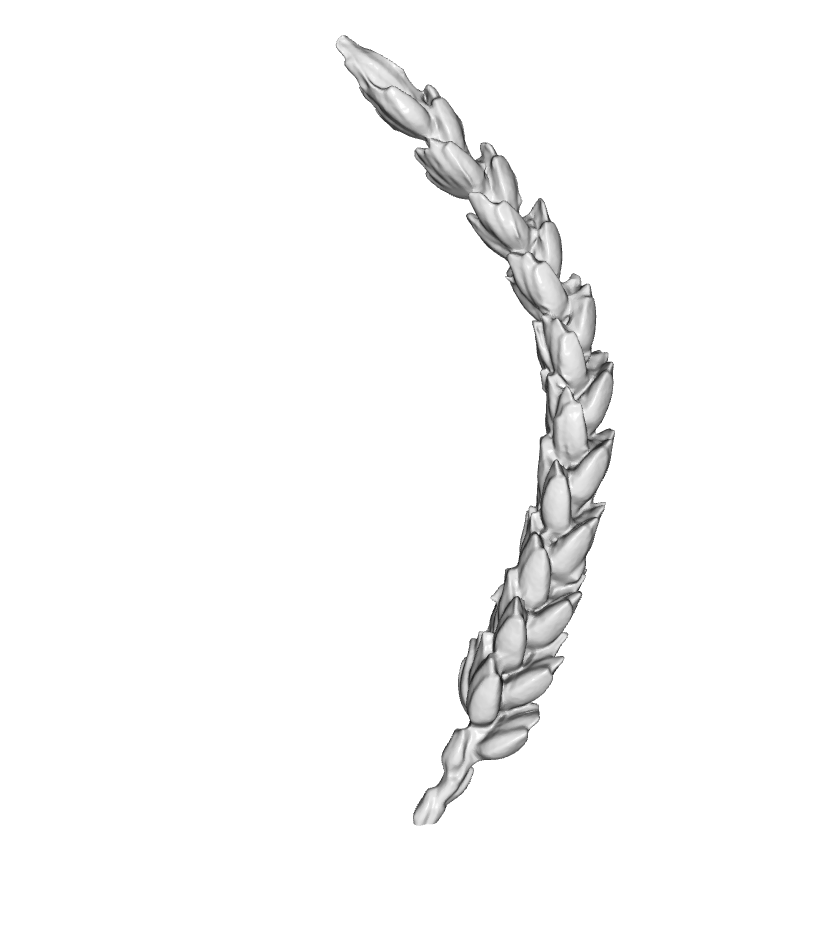}
            \caption{}
        \end{subfigure}
    
        \caption{3D meshes before (a) and after cone removal (b).}
        \label{fig:cone_comparison}
    \end{figure}

     The meshes of reconstructed objects obtained from the scanning process included both the wheat spike and the supporting cone used to stabilise during scanning. To isolate the volume of the spike, the underlying cone was removed, as illustrated in Figure~\ref{fig:cone_comparison}, a and b. The removal of the cones was facilitated by consistent placement of the cones in all scans, where the vertical position ($z$-coordinate) remained fixed and only the horizontal coordinates ($x$ and $y$) required determination. This localisation was achieved by identifying the minimum and maximum extent of the cone along the $x$- and $y$-axes and computing the midpoint to approximate the cone's central axis. Combined with a predefined apex height and a known base radius at cone level $z$, these parameters enabled the reconstruction of a cone-shaped volume, which was later removed from the mesh to isolate the spike. Subsequent volume computation was performed using the Python libraries PyMeshLab \citep{muntoniCnristivclabPyMeshLabPyMeshLab2023} and Open3D \citep{zhouOpen3DModernLibrary2018}, which provide efficient tools for processing and analysing 3D mesh data. 

\subsection{Volume estimation}

    To evaluate the hypothesis that neural networks can improve volume estimation for complex-shaped objects, we compared their performance to two baseline models of varying complexity. The dataset of wheat spikes was split based on genotypes into training, validation and test sets in a 70\%, 10\% and 20\% ratio. Splitting was done based on genotypes to prevent overfitting and ensure that the model does not learn genotype-specific features.
    
\subsection{Area baseline}

    The first, simpler area-baseline was computed by counting the number of pixels belonging to the spike, and analysing their correlation with the ground truth volume. To analyse volume estimation from multiple view points, the average pixel count across views was used as a proxy for the 2D projected area of the spike. This baseline served to assess the extent to which simple 2D information can approximate volumetric measurements. 
    
\subsection{Geometric baseline}

\begin{figure} 
    \centering
    \begin{subfigure}[htbp]{0.27\columnwidth}
        \centering
        \includegraphics[width=\linewidth]{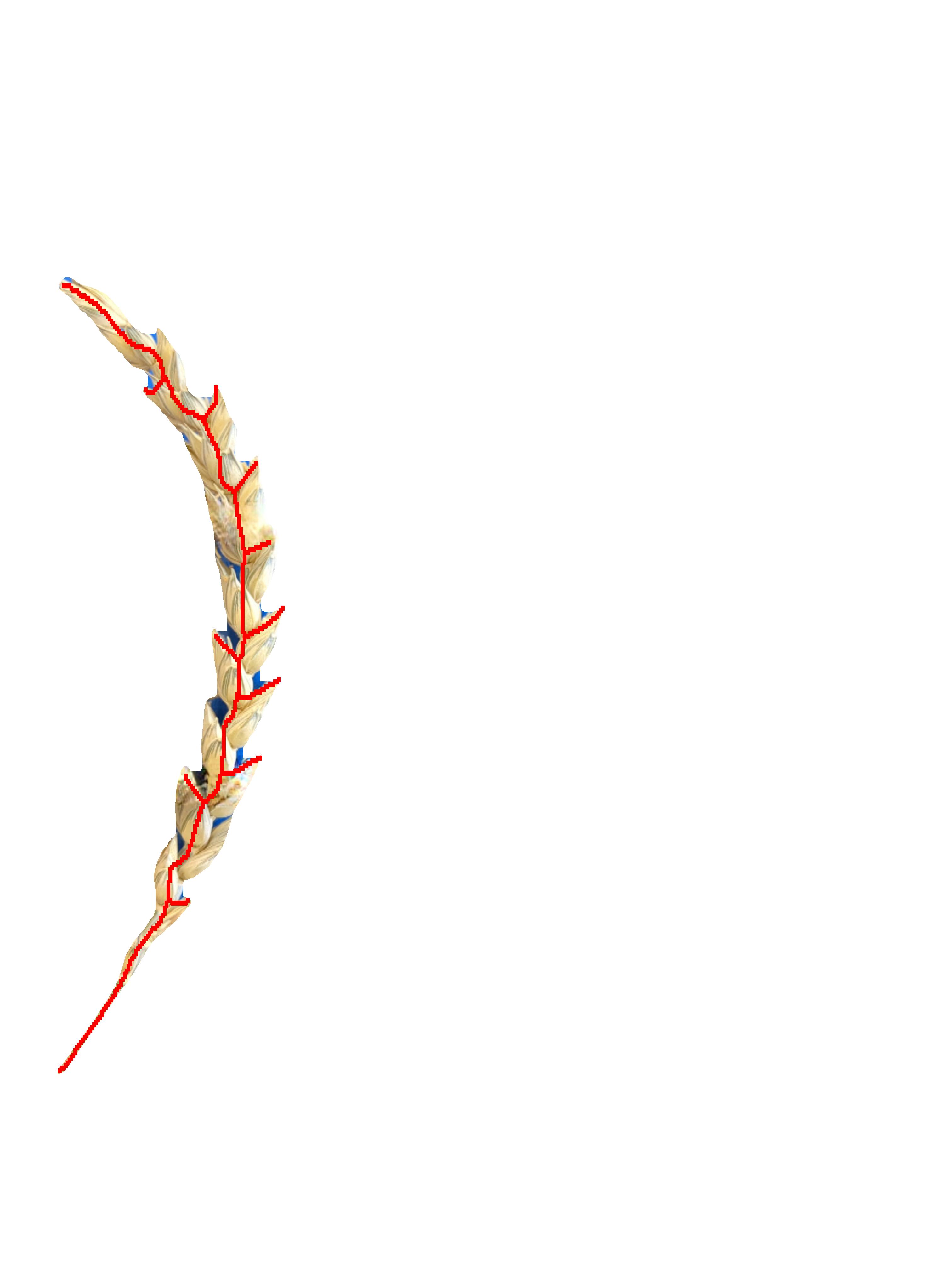}
        \caption{}
    \end{subfigure}
    \hfill
    \begin{subfigure}[htbp]{0.27\columnwidth}
        \centering
        \includegraphics[width=\linewidth]{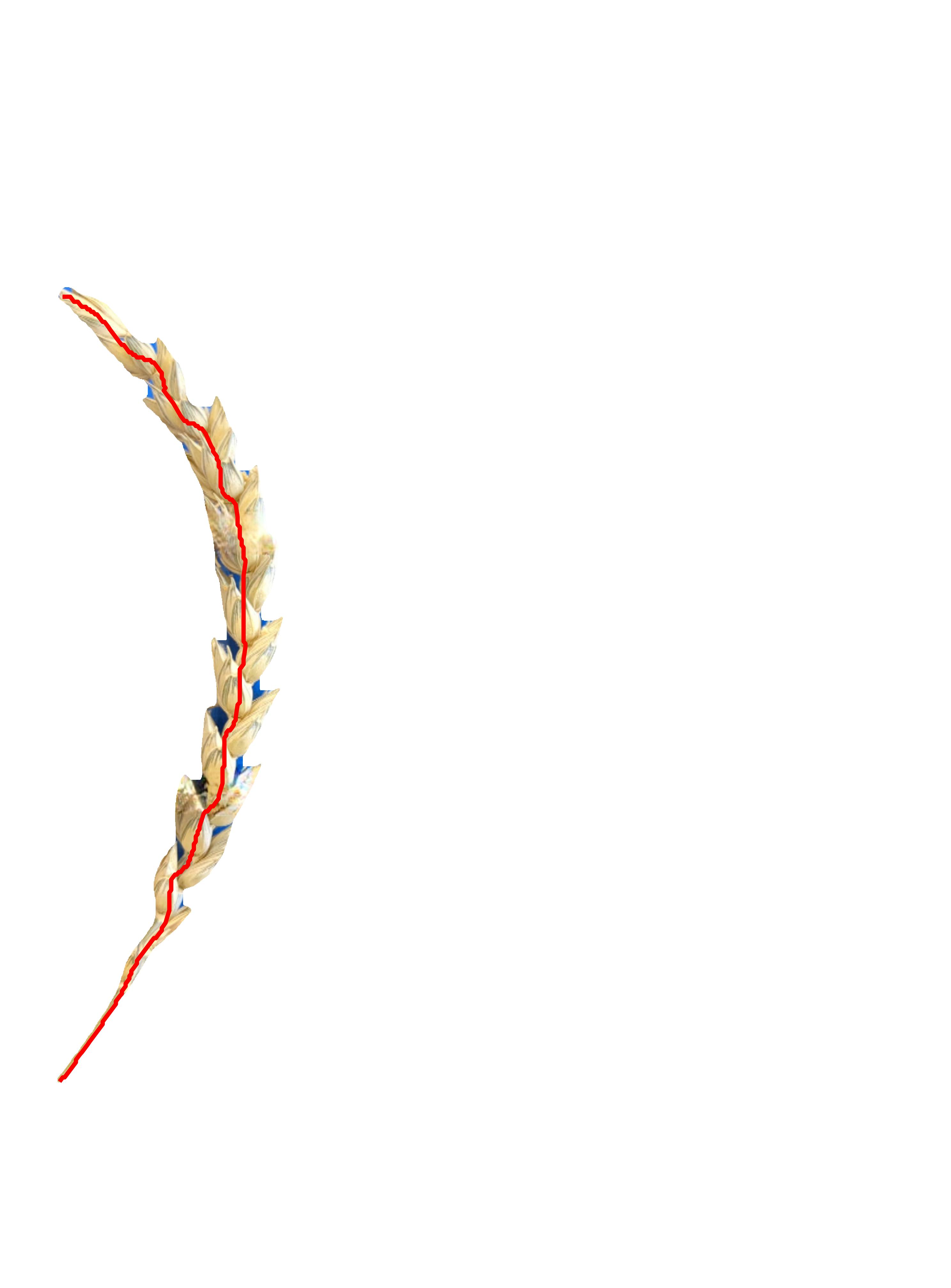}
        \caption{}
    \end{subfigure}
    \hfill
    \begin{subfigure}[htbp]{0.27\columnwidth}
        \centering
        \includegraphics[width=\linewidth]{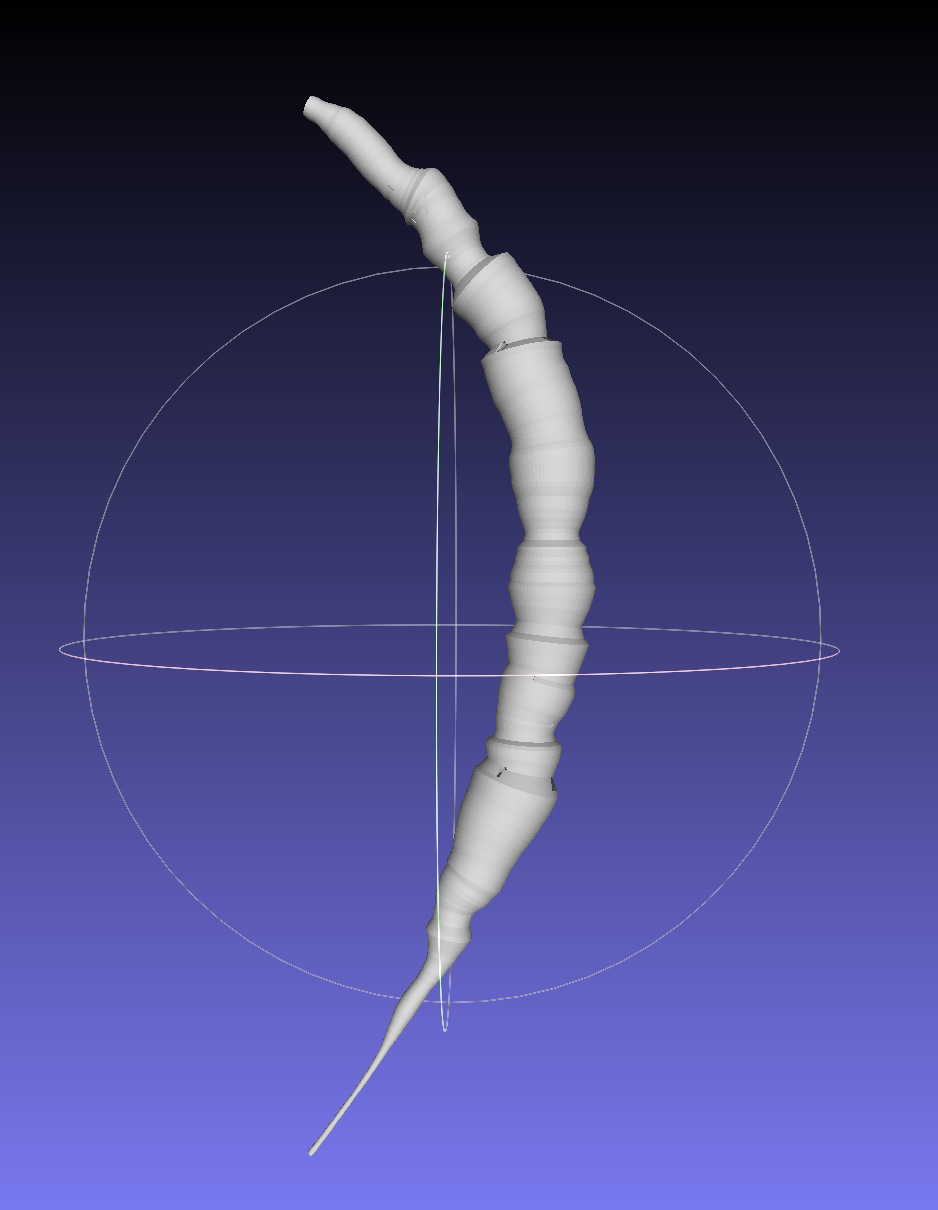}
        \caption{}
    \end{subfigure}

    \caption{Visualization of the geometric baseline, including the skeleton of a spike (a), defined main axis (b), and the final spike volume approximation (c).}
    \label{fig:geometric_baseline}
    \end{figure}

    The geometric baseline is based on the assumption that the spike's volume increases non-linearly with its radius and that its cross-section can be approximated as a circle along its longitudinal axis. To estimate the volume, the spike was treated as a series of stacked discs with varying radii. First, a binary spike mask was generated via segmentation using the Segment Anything Model (SAM), as described above. Then the skeleton of the spike within the binary segmentation mask was computed using the thinning algorithm proposed by \citep{zhangFastParallelAlgorithm1984} as illustrated in Figure~\ref{fig:geometric_baseline}a. To extract the main axis of the spike (Figure~\ref{fig:geometric_baseline}b), we applied a custom recursive algorithm (Appendix Algorithm~\ref{alg:main_axis}) that operates on the skeleton. Starting from an initial point, the algorithm traverses the skeleton by recursively exploring all neighbouring pixels and identifying the longest continuous path, which is then returned as the spike's main axis. We fitted a smoothed central axis of the spike to obtain its length and measured its width orthogonally along this curve. We derive a radius function \(r : [0,1] \rightarrow \mathbb{R}_+\), which gives the spike radius along its length. Assuming circular cross-sections, the total volume is estimated by:
    
    \[
    V = \pi \int_{0}^{L} \left( r\left(\frac{t}{L} \right) \right)^2 dt.
    \]
    
    where $V$ is the estimated spike volume and $L$ the spike length. We approximate this integral using the trapezoidal rule with 600 evenly spaced evaluation points. Higher-order quadrature methods were avoided due to spline smoothness limitations. The resulting volume is visualised in Figure~\ref{fig:geometric_baseline}c.

\subsection{Neural network}

\begin{figure} 
        \centering
        \vskip 0.2in
        \includegraphics[width=\linewidth]{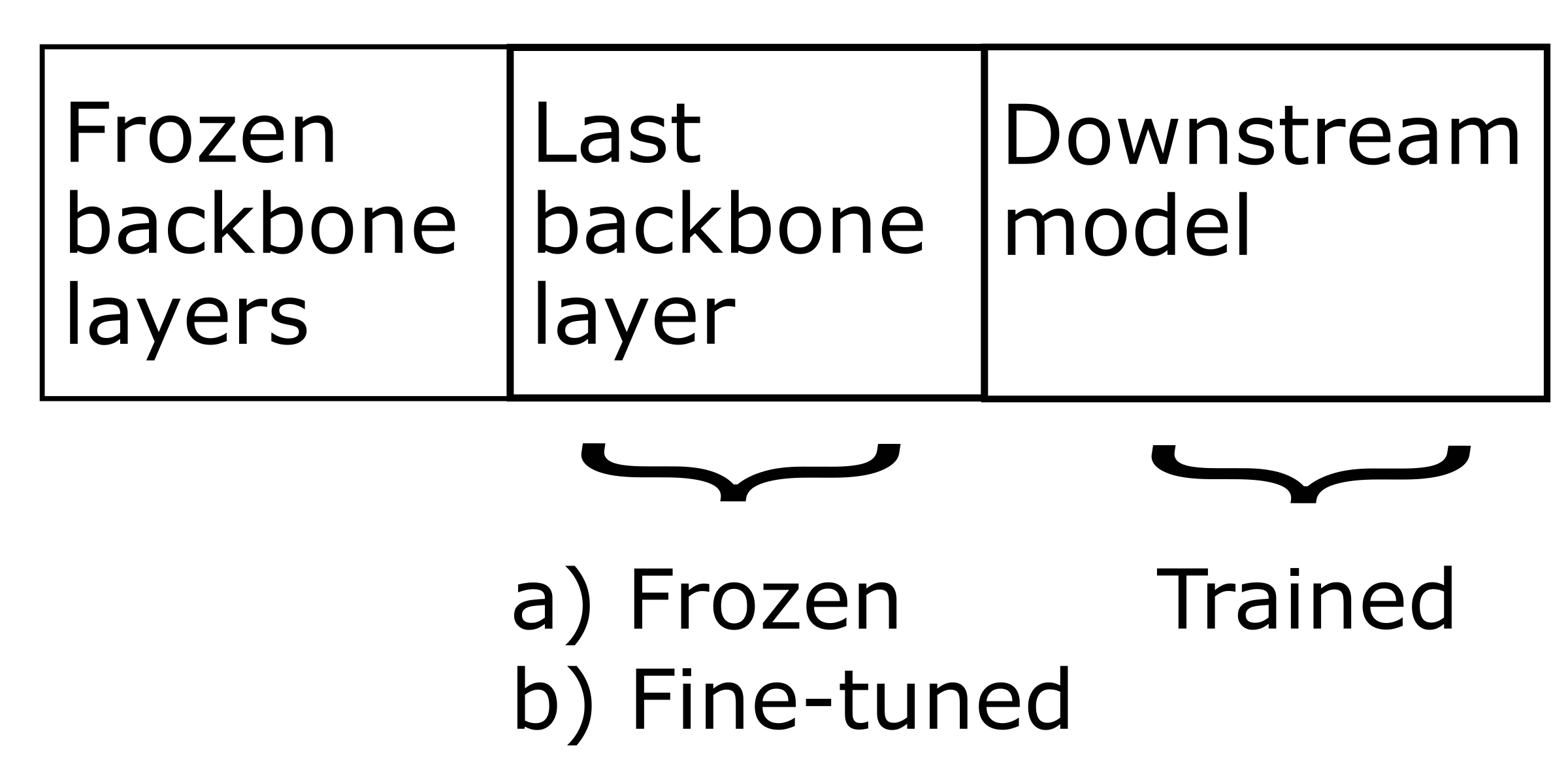}
        \caption{Neural networks were trained using either a fully frozen backbone and a downstream model (a), or a backbone with the final layer fine-tuned and a downstream model (b).}
        \label{fig:backbone_downstream}
        \vskip -0.2in
    \end{figure}

    To directly predict volume from images by training neural networks, input images were downsampled from their original resolution of 3024 x 4032 pixels to 256 x 341 pixels using bilinear interpolation with anti-aliasing enabled, and subsequently center-cropped to 244 x 244 pixels to align with the input size required by pre-trained backbones. The measured spike volume was normalised using the empirical mean and standard deviation of the training set to facilitate stable optimisation. 
    To take advantage of models pre-trained on large datasets, we employed the small versions of DINOv2 and DINOv3, two state-of-the-art self-supervised Vision Transformer models (ViTs), to extract intermediate image embeddings suitable for general visual understanding tasks \citep{oquabDINOv2LearningRobust2024, simeoniDINOv32025}.     
    In the first setup, each image was passed through the fully frozen DINOv2 or DINOv3 encoder producing embeddings of dimension 384. The sequence of embeddings, representing different views of the spike, was then fed into downstream models, either a Multi-Layer Perceptron (MLP), an LSTM, or a Transformer to estimate spike volume (Figure~\ref{fig:backbone_downstream}, setup (a)). MLPs contain fully connected input, output, and hidden layers. The features extracted from multiple images per spike were averaged prior to being passed to the MLP, resulting in a single aggregated representation analogous to the averaging used in the baseline methods. LSTMs are a type of recurrent neural network (RNN) with the ability to learn and predict long-term dependencies of sequential data by capturing information from previous time steps \citep{hochreiterLongShortTermMemory1997}. Transformers further extend these capabilities by modelling contextual relationships through their self-attention mechanism \citep{vaswaniAttentionAllYou2023}. This hybrid approach, using backbone features and a downstream model, may recognise complex patterns by combining the two networks to make use of each model's advantages. While the backbone models may learn global features such as edges, textures, and other low-level traits, the downstream models (MLP, LSTM and Transformer) may learn higher level features and relationships based on the sequential nature of the input (LSTM and Transformer).

    We then compared this first setup to a second setup where we additionally fine-tuned the last layer of the pre-trained backbone models in combination with the best-performing downstream models (Figure~\ref{fig:backbone_downstream}, setup (b)). In this approach, all previous layers were frozen to preserve their generic representations while the last layer and the same task-specific downstream models were trained and optimised for our dataset. This approach retains the pre-trained representations, but allows the highest-level representations of the backbone model to adjust to the target domain. We compared the best-performing ViTs-downstream model combinations with CNNs, specifically ResNet18 and ResNet50, as backbone models, which were originally designed for image tasks such as classifications \citep{heDeepResidualLearning2015}. In addition, we evaluated FoMo4wheat, a foundation model designed specifically for wheat image tasks \citep{hanFoMo4WheatReliableCrop2025}.

    \subsubsection{Training procedure}

    Due to the imbalance in volume distribution, where more samples are clustered around the mean, we used a scaled MSE loss similar to \citep{renBalancedMSEImbalanced2022}. In this loss, the sample is scaled inversely proportional to its frequency in the dataset:
    
    \begin{equation}
    \mathcal{L}_{\text{Scaled}} = \frac{1}{N} \sum_{i=1}^{N} w_i (y_i - \hat{y}_i)^2 \,,
    \end{equation}

    where \( N \) denotes the number of samples, \( y_i \) is the measured volume for the \( i \)-th sample, \( \hat{y}_i \) is the corresponding estimated volume, and \( w_i \) is a weighting factor used to scale the squared error of the \( i \)-th sample.
    
    The weights $w_i$ were calculated using an inverse frequency approach based on histogram binning of the denormalised volume values. Samples in under-represented bins (extremely small or large volumes) received higher weights,
    \begin{align}
    w_i &= \frac{1}{\max \{w_k' : 1 <= k <= N\}} w_i', &w_i' &=   \frac{1}{\text{freq}(b_i)}\,,
    \end{align}
    where $b_i$ denotes the bin index of sample $i$. The weights are then normalised to have a maximum value of 1.

\begin{figure*} 
        \makebox[\textwidth][l]{
            \includegraphics[width=\textwidth]{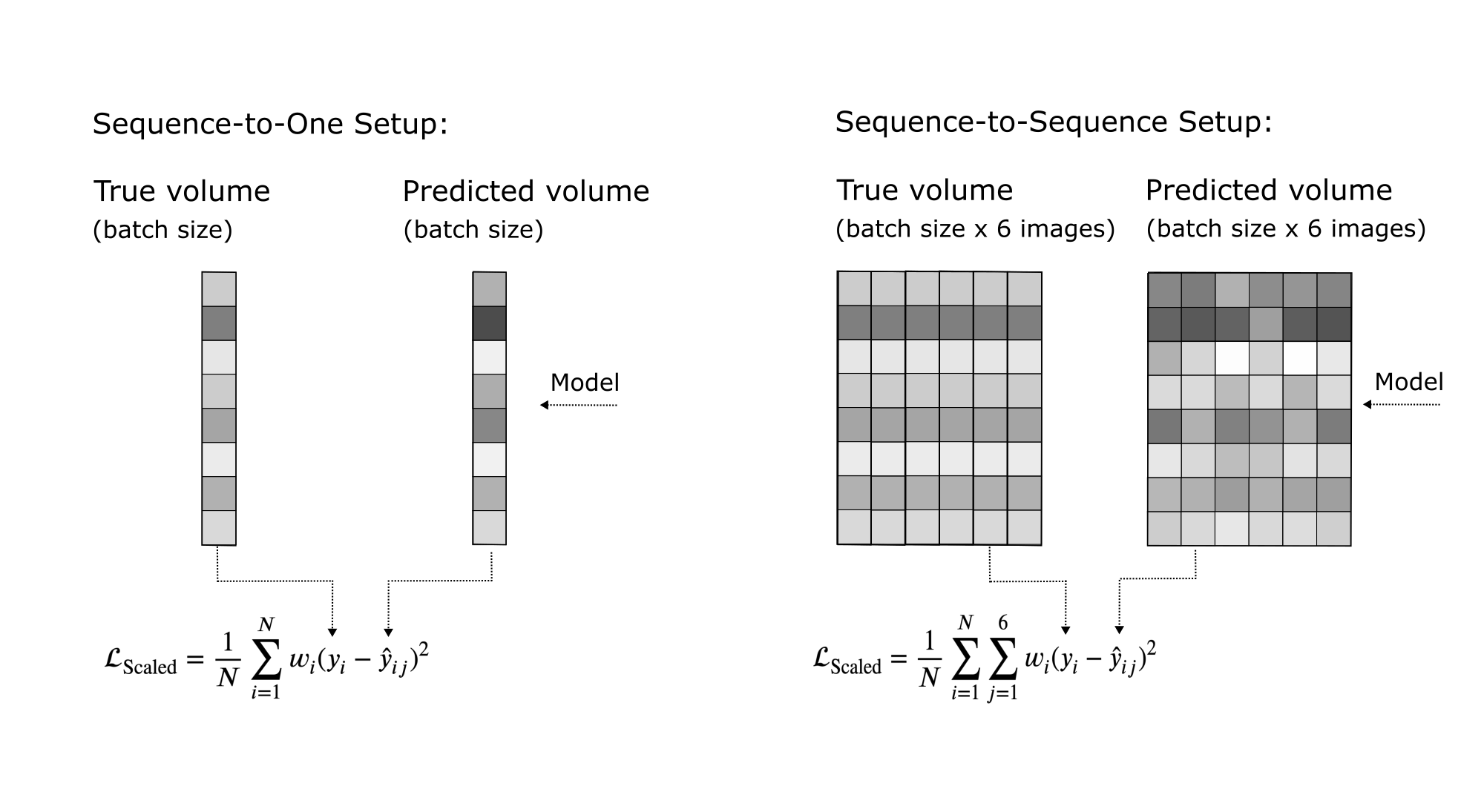}}
        \caption{Sequence-to-one setup (a), and sequence-to-sequence setup (b) used to train volume estimation on 6 input images per spike.}
        \label{fig:sequences}
    \end{figure*}

    We implemented two different estimation schemes for mapping sequential model outputs to spike volume estimates in the LSTM and Transformer architectures. In the sequence-to-one setup, only the final hidden state was used and passed through a linear layer to predict a single scalar volume per spike (Figure~\ref{fig:sequences}a). In the sequence-to-sequence setup, the linear layer was applied to each LSTM or Transformer output, resulting in one scalar estimation per input image (Figure~\ref{fig:sequences}b). This corresponds to a deep supervision approach, where supervision is applied to all sequence outputs rather than only to the final representation \citep{li2022comprehensivereviewdeepsupervisionlu}.

     Therefore, the input to the scaled loss was adapted to 

    \begin{equation}
        \mathcal{L}_{\text{Scaled}} = \frac{1}{N}  \sum_{i=1}^{N} \sum_{j=1}^{6} w_i (y_i - \hat{y}_{ij})^2\,,
    \end{equation}

    where $j$ denotes the six images per spike. 
    
    The LSTMs and the Transformers were trained on all six images to leverage their ability to integrate information across multiple views, while the MLPs were trained using the same number of images as used during evaluation. The models were trained for 5000 epochs with a batch size of 32, using the Adam optimiser, CeLU activations, and dropout regularisation ($p = 0.5$). The learning rate started at $1 \times 10^{-4}$ decreasing linearly to $1 \times 10^{-6}$ over the course of the training. The model corresponding to the lowest validation loss was selected as the final model. During training, the images were randomly sampled such that models trained on fewer than six images still encountered all available views across epochs, ensuring a fair comparison between models. The models were then evaluated on six, four, two, and one image(s), respectively. The best-performing models evaluated on single side-view indoor images were evaluated on the single side-view field images. Additionally, the same models were fine-tuned using the single side-view field images from the training dataset. For fine-tuning on field images, the volumes were normalised during training using the mean and standard deviation of the field image training set.

\section{Results}

\begin{figure} 
        \centering
        \includegraphics[width=\columnwidth]{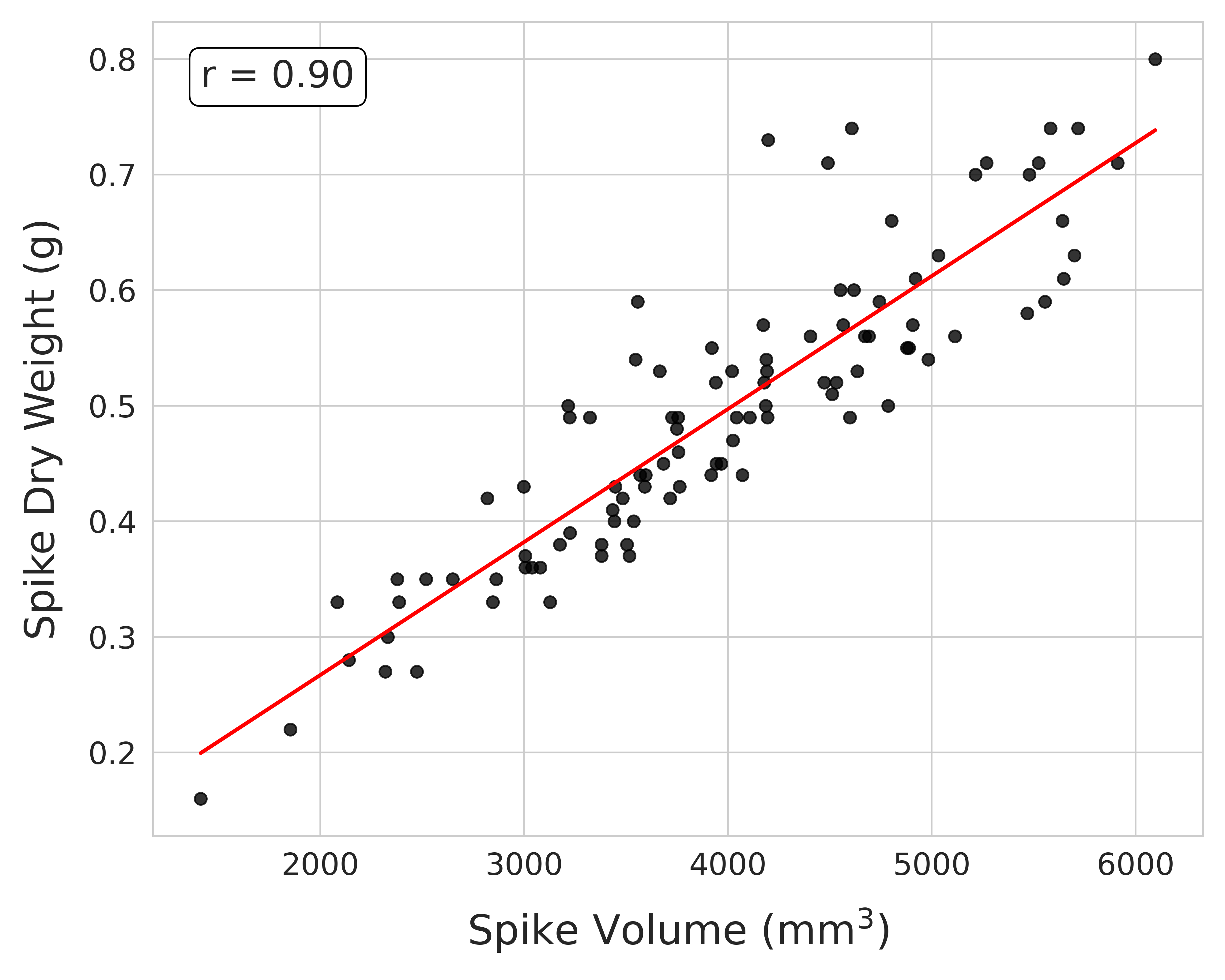}
        \caption{Scatter plot with linear regression between spike volume measured by structured-light 3D scanning and dry weight at flowering. Measurements were taken in 2024. Each point represents one spike.}
        \label{fig:correlation_plot}
    \end{figure}

    We found a high correlation between spike dry weight and the measured spike volume at flowering of 100 spikes (Figure~\ref{fig:correlation_plot}). Hence, the spike volume can serve as a proxy for the spike dry weight at anthesis representing the fruiting capacity as a novel digital trait indicating an upper bound of the grain-filling potential.

\begin{figure*} 
        \centering
        \includegraphics[width=1\textwidth]{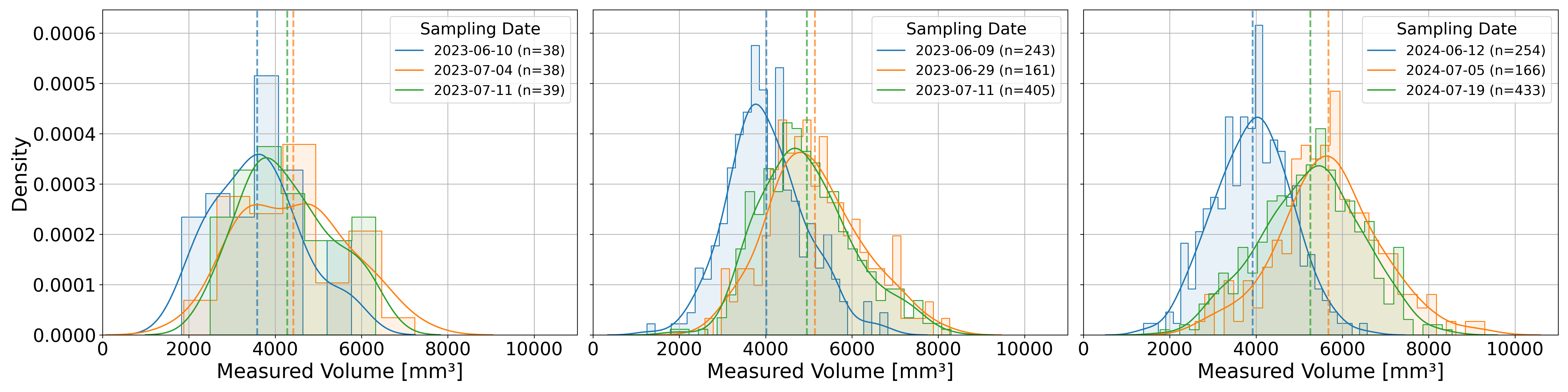}
        \caption{Measured spike volume distribution of the smaller shaded panel in 2023 (a), the regular panel in 2023 (b), and the regular panel in 2024 (c), as well as the number of sampled spikes. The mean of the distribution is represented as a dashed line.}
        \label{fig:distribution_volume}
    \end{figure*}

    Figure~\ref{fig:distribution_volume} shows the volume distribution of the small shaded panel sampled in 2023 (a), the regular panel sampled in 2023 (b), and the regular panel sampled in 2024 (c), as well as the number of sampled spikes. The mean volume of the first sampling group was smaller, compared to the second and third sampling. Furthermore, the means of the small panel, which contains shaded spikes, were lower compared to the regular panels, increasing the diversity of the wheat spikes in the data set (Table~\ref{tab:mean_volume_wide}). 

    \begin{table*}[htbp]
        \centering
        \caption{Mean volume (in mm\textsuperscript{3}) for each sampling date across groups.}
        \label{tab:mean_volume_wide}
        \begin{tabularx}{\textwidth}{@{}p{8cm}XXX@{}}
            \toprule
            \textbf{Sampling Date} & \textbf{2023 shaded} & \textbf{2023} & \textbf{2024} \\
            \midrule
            2023-06-10 / 2023-06-09 / 2024-06-12 & 3583.24 & 4019.84 & 3921.87 \\
            2023-07-04 / 2023-06-29 / 2024-07-05 & 4421.58 & 5143.01 & 5678.33 \\
            2023-07-11 / 2023-07-11 / 2024-07-19 & 4280.17 & 4956.93 & 5258.91 \\
            \bottomrule
        \end{tabularx}
    \end{table*}
    
    We evaluated our deep learning models against two conventional baselines, the 2D area-based method and a geometric model assuming cylindrical cross-sections. The correlation, the coefficient of determination ($R^2$), the mean absolute percentage error (MAPE), and the mean absolute error (MAE) were used as performance metrics in different numbers of evaluation images (six, four, two and one). Among the baselines, the area baseline consistently outperformed the geometric baseline across all image counts (Table~\ref{tab:baselines}).

    \begin{table*}[htbp] 
        \centering
        \caption{Correlation, $R^2$, MAPE, and MAE of measured and estimated spike volumes using the area and geometric baselines. Models were evaluated on six, four, two, and one image(s), respectively.}
        \label{tab:baselines} 
        \small
        \begin{tabularx}{\textwidth}{@{}p{0.19\textwidth}*{8}{>{\centering\arraybackslash}X}@{}}
        \toprule
        Algorithm & \multicolumn{4}{c}{Six images} & \multicolumn{4}{c}{Four images} \\
        \cmidrule(r){2-5} \cmidrule(l){6-9} & Corr. & $R^2$ & MAPE (\%) & MAE (mm$^3$) & Corr. & $R^2$ & MAPE (\%) & MAE (mm$^3$) \\
        \midrule
            Area baseline     & \textbf{0.88} & \textbf{0.73}  & \textbf{9.55}  & \textbf{439.02} & \textbf{0.88} & \textbf{0.73} & \textbf{9.66} & \textbf{445.07} \\
            Geometric baseline  & 0.86 & 0.70 & 9.86 & 445.43 & 0.85 & 0.68 & 10.11 & 456.81 \\
            \addlinespace[1ex]
            \multicolumn{9}{c}{\textbf{}}\\[-1.5ex]
            \midrule
            Algorithm 
            & \multicolumn{4}{c}{Two images}  & \multicolumn{4}{c}{One image} \\
            \cmidrule(r){2-5} \cmidrule(l){6-9}
            & Corr. & $R^2$ & MAPE (\%) & MAE (mm$^3$) & Corr. & $R^2$ & MAPE (\%) & MAE (mm$^3$) \\
            \midrule
            Area baseline     & \textbf{0.88} & \textbf{0.73} & \textbf{9.80} & \textbf{448.35} & \textbf{0.81} & \textbf{0.62} & \textbf{11.71} & \textbf{535.11} \\
            Geometric baseline  & 0.85 & 0.69 & 10.23 & 462.40 & 0.76 & 0.54 & 13.52 &  599.30 \\
            \bottomrule
        \end{tabularx}
    \end{table*}

    Among the neural network models with fully frozen backbones, DINOv2 outperformed the newer version DINOv3 in most cases. For LSTMs and Transformers, a deep supervised strategy (sequence-to-sequence) consistently improved volume estimation across different number of views. Overall, the best-performing LSTM (DINOv2, sequence-to-sequence) and MLP (DINOv2) clearly outperformed the Transformers, with the LSTM showing slightly higher accuracy (Table~\ref{tab:merged-models-frozen}). 

    \begin{table*}[htbp]
        \centering
        \caption{Correlation, $R^2$, MAPE, and MAE of measured and estimated spike volumes for different image counts. Backbone models were fully frozen. Models were evaluated on six, four, two, and one image(s), respectively.}
        \label{tab:merged-models-frozen}
        \small
        \setlength{\tabcolsep}{3pt}
        \makebox[\linewidth][c]{%
        \begin{tabularx}{1.0\textwidth}{@{}p{0.23\textwidth}*{8}{>{\centering\arraybackslash}p{0.085\textwidth}}@{}}
        \toprule
        Algorithm & \multicolumn{4}{c}{Six images} & \multicolumn{4}{c}{Four images} \\
        \cmidrule(r){2-5} \cmidrule(l){6-9} & Corr. & $R^2$ & MAPE (\%) & MAE (mm$^3$) & Corr. & $R^2$ & MAPE (\%) & MAE (mm$^3$) \\
            \midrule
            MLP (DINOv2)  & \textbf{0.94} & \textbf{0.88} & \textbf{6.69} & \textbf{301.57}&\textbf{0.93}&\textbf{0.86}  & \textbf{7.22} &  \textbf{320.24}  \\ 
            MLP (DINOv3)   & 0.94 & 0.87 & 7.06   & 316.98 & 0.93 & 0.86  & 7.53 & 337.05  \\
            \midrule 
            LSTM (DINOv2, seq-to-one) & 0.94 & 0.86 & 6.64 & 305.61  & 0.94  & 0.82  & 7.74  &  355.73  \\ 
            LSTM (DINOv3, seq-to-one) & 0.95 & 0.89 & 6.38  & 288.59  & 0.93  & 0.87  & 7.16  & 321.59 \\
            LSTM (DINOv2, seq-to-seq)&\textbf{0.95}& \textbf{0.89}&\textbf{6.31} &\textbf{289.70}  & \textbf{0.94} & \textbf{0.88}& \textbf{6.59}&\textbf{300.66}   \\ 
            LSTM (DINOv3, seq-to-seq) & 0.95& 0.89 & 6.41  & 292.14  &0.95& 0.88 & 6.75  & 309.05 \\ 
            \midrule 
            Transf. (DINOv2, seq-to-one) & 0.90 & 0.79 & 9.05  & 412.69  & 0.87 & 0.74 &  10.81 & 462.51 \\  
            Transf. (DINOv3, seq-to-one) & 0.87 & 0.70 &  10.47 & 486.16  & 0.86 & 0.62 & 14.15  & 578.94 \\ 
            Transf. (DINOv2, seq-to-seq)&\textbf{0.91}&\textbf{0.80}&\textbf{8.71}&\textbf{397.81}&\textbf{0.89}&\textbf{0.78}&\textbf{9.33}& \textbf{418.12}\\  
            Transf. (DINOv3, seq-to-seq) & 0.90 & 0.78 &  9.03 &  414.42 & 0.89 & 0.77 &  9.40 & 430.02 \\ 
            \addlinespace[1ex]
            \multicolumn{9}{c}{\textbf{}}\\[-1.5ex]
            \midrule
            Algorithm 
            & \multicolumn{4}{c}{Two images}  & \multicolumn{4}{c}{One image} \\
            \cmidrule(r){2-5} \cmidrule(l){6-9}
            & Corr. & $R^2$ & MAPE (\%) & MAE (mm$^3$) & Corr. & $R^2$ & MAPE (\%) & MAE (mm$^3$) \\
            \midrule
            MLP (DINOv2)   & \textbf{0.92} & \textbf{0.85} & \textbf{7.92}   & \textbf{350.87} &   \textbf{0.89}    &  \textbf{0.78}  &  \textbf{9.71} & \textbf{427.22}   \\ 
            MLP (DINOv3) &   0.91 & 0.83 &  8.46   &  375.70   &  0.88   & 0.76    & 9.97  & 441.85 \\
            \midrule 
            LSTM (DINOv2, seq-to-one) & 0.92  & 0.23 & 19.79 & 841.10  &  0.84 & -0.90  & 31.66 & 1322.61    \\ 
            LSTM (DINOv3, seq-to-one) & 0.90 & 0.47 & 15.46  & 686.98  & 0.79 & -0.93  & 31.10  & 1367.58 \\
            LSTM (DINOv2, seq-to-seq) & \textbf{0.93}&\textbf{0.86} & \textbf{7.13} & \textbf{323.79}&\textbf{0.90}&\textbf{0.81} & \textbf{8.94}&\textbf{399.94}   \\ 
            LSTM (DINOv3, seq-to-seq) & 0.92 & 0.84 & 7.79  &  354.52 & 0.87 & 0.76 & 9.85  &  447.39\\ 
            \midrule 
            Transf. (DINOv2, seq-to-one)  & 0.86 & 0.73 &  10.57 &  473.85 & 0.80 & 0.36 &  14.93 & 725.26 \\ 
            Transf. (DINOv3, seq-to-one)  & 0.81 &  0.62&  12.17 &  552.29 & 0.71 & 0.35 & 18.31  &755.47  \\ 
            Transf. (DINOv2, seq-to-seq) & 0.88 &  0.76 &  10.03 & 445.56  & \textbf{0.85} & \textbf{0.72} & \textbf{10.90}  & \textbf{482.79} \\  
            Transf. (DINOv3, seq-to-seq) & \textbf{0.88} & \textbf{0.75} &  \textbf{9.72} & \textbf{444.63}  & 0.80 & 0.60 &  11.85 & 554.89 \\ 
            \bottomrule
        \end{tabularx}
        }
    \end{table*}

    Given the lower performance of the Transformer-based models, subsequent analyses focused on MLPs and LSTMs using the sequence-to-sequence setup (Table~\ref{tab:merged-models-fine-tuned}). Fine-tuning the DINOv2 and DINOv3 backbones in combination with these downstream models led to substantial performance gains compared to fully frozen backbones. Notably, after fine-tuning, DINOv3 outperformed DINOv2 in most cases. We further compared the best-performing fine-tuned DINO-based models with fine-tuned CNN backbones (ResNet18 and ResNet50), as well as the wheat-specific FoMo4Wheat backbone, using the same MLP and LSTM downstream architectures. While ResNet50 generally outperformed ResNet18, both the CNN and the FoMo4Wheat backbones remained inferior to the fine-tuned DINOv2 and DINOv3 models. The fine-tuned DINOv3-MLP model evaluated on six images achieved the best performance with a correlation of 0.97, $R^2$ of 0.94, MAPE of 4.67\%, and MAE of $214.82~\mathrm{mm}^3$. As expected, the estimation error decreased with an increasing number of evaluation images. Remarkably, even when evaluated on a single side-view image per spike, the fine-tuned DINOv3-MLP model surpassed the performance of both baseline methods evaluated on all six images.

    \begin{table*}[htbp]
        \centering
        \caption{Correlation, $R^2$, MAPE, and MAE of measured and estimated spike volumes for different image counts. The final backbone layer was fine-tuned. The LSTMs were trained using sequence-to-sequence setup. Models were evaluated on six, four, two, and one image(s), respectively.}
        \label{tab:merged-models-fine-tuned}
        \small
        \setlength{\tabcolsep}{3pt}
        \makebox[\linewidth][c]{%
        \begin{tabularx}{1\textwidth}{@{}p{0.23\textwidth}*{8}{>{\centering\arraybackslash}p{0.085\textwidth}}@{}}
        \toprule
        Algorithm & \multicolumn{4}{c}{Six images} & \multicolumn{4}{c}{Four images} \\
        \cmidrule(r){2-5} \cmidrule(l){6-9} & Corr. & $R^2$ & MAPE (\%) & MAE (mm$^3$) & Corr. & $R^2$ & MAPE (\%) & MAE (mm$^3$) \\
        \midrule
            MLP (DINOv2) & 0.96 & 0.93 & 5.08 &230.49 & 0.96 & 0.91 & 5.68 & 257.56 \\ 
            MLP (DINOv3) & \textbf{0.97} & \textbf{0.94} & \textbf{4.67} & \textbf{214.82} & 0.96 &0.92 &5.71 & 261.25\\
            LSTM (DINOv2) & 0.97 & 0.93 & 5.16 & 233.67 & 0.96 & 0.90 & 5.93 & 273.63 \\
            LSTM (DINOv3) & 0.97 & 0.93 & 5.31 & 243.82 & \textbf{0.96} & \textbf{0.92} & \textbf{5.55} & \textbf{255.72} \\
            \midrule 
            MLP (ResNet18) &  0.96 & 0.91 & 5.69 & 258.37 & 0.95 & 0.88 & 6.99 & 311.73 \\ 
            MLP (ResNet50) &  \textbf{0.96} & \textbf{0.91} & \textbf{5.69} & \textbf{257.01} & \textbf{0.95} & \textbf{0.90} & \textbf{6.39} & \textbf{286.79} \\ 
            LSTM (ResNet18) & 0.95 & 0.90 & 6.05 & 277.72 & 0.94 & 0.89 & 6.36 & 293.44 \\
            LSTM (ResNet50) &0.94 & 0.88 & 6.70 & 300.83 & 0.94 & 0.88 &  6.81 & 304.68\\
            \midrule
            MLP (FoMo) & \textbf{0.94} & \textbf{0.88} & \textbf{6.62} & \textbf{303.25} & \textbf{0.94} & \textbf{0.87} & \textbf{7.18} & \textbf{324.26} \\ 
            LSTM (FoMo) & 0.95 & 0.88 & 6.80 & 312.90 & 0.94 & 0.86 & 7.18 & 337.85 \\ 
            \addlinespace[1ex]
            \multicolumn{9}{c}{\textbf{}}\\[-1.5ex]
            \midrule
            Algorithm 
            & \multicolumn{4}{c}{Two images}  & \multicolumn{4}{c}{One image} \\
            \cmidrule(r){2-5} \cmidrule(l){6-9}
            & Corr. & $R^2$ & MAPE (\%) & MAE (mm$^3$) & Corr. & $R^2$ & MAPE (\%) & MAE (mm$^3$) \\
            \midrule
            MLP (DINOv2) &\textbf{0.95} &\textbf{0.91} &\textbf{5.97} &\textbf{265.43} & 0.94& 0.87& 7.20& 318.43\\ 
            MLP (DINOv3) &0.95 &0.90 &6.16 &279.83 & \textbf{0.93} & \textbf{0.87} & \textbf{6.92} & \textbf{316.22} \\
            LSTM (DINOv2) &0.95 & 0.91 & 6.13 & 279.92 & 0.93 & 0.85 & 7.73 & 347.95 \\
            LSTM (DINOv3) & 0.95 & 0.90 & 6.20 & 285.28 & 0.92 & 0.84 & 7.82 & 354.17  \\
            \midrule 
            MLP (ResNet18) & 0.94 & 0.88 & 7.05 & 314.05 & \textbf{0.93} & \textbf{0.85} & \textbf{7.67} &  \textbf{342.29}  \\  
            MLP (ResNet50) & \textbf{0.94} & \textbf{0.89} & \textbf{6.76} & \textbf{302.75} & 0.92 & 0.83 & 8.10& 355.35  \\ 
            LSTM (ResNet18) & 0.93 & 0.86& 7.00 & 323.12 & 0.88 & 0.76 & 9.63 & 442.42 \\
            LSTM (ResNet50) & 0.93 & 0.86 & 7.58 & 339.55 & 0.89 & 0.76 &  9.59 &  432.68 \\
            \midrule 
            MLP (FoMo) & \textbf{0.91} & \textbf{0.83} & \textbf{8.16} & \textbf{371.75} & \textbf{0.86} & \textbf{0.73} & \textbf{10.20} & \textbf{466.52} \\ 
            LSTM (FoMo) &  0.91& 0.81 & 8.55 & 398.90 & 0.82 & 0.61 & 11.97 & 558.21 \\ 
            \bottomrule
        \end{tabularx}
        }
    \end{table*}

\begin{figure} 
        \centering
        \begin{subfigure}[t]{0.48\textwidth}
            \centering
            \includegraphics[width=1\linewidth]{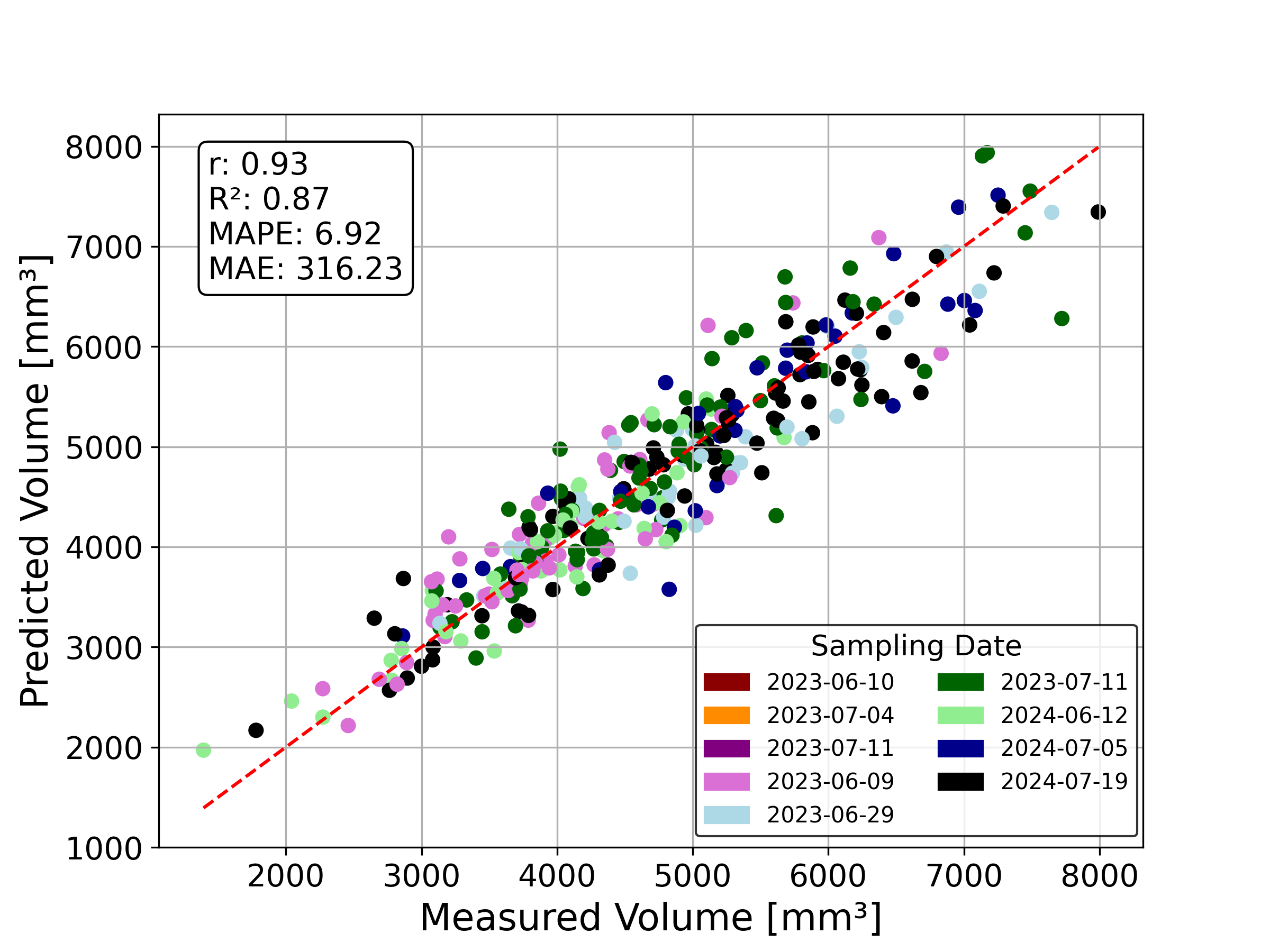}
            \caption{}
        \end{subfigure}
        \vspace{0.02\columnwidth}
        \begin{subfigure}[t]{0.48\textwidth}
            \centering
            \includegraphics[width=1\linewidth]{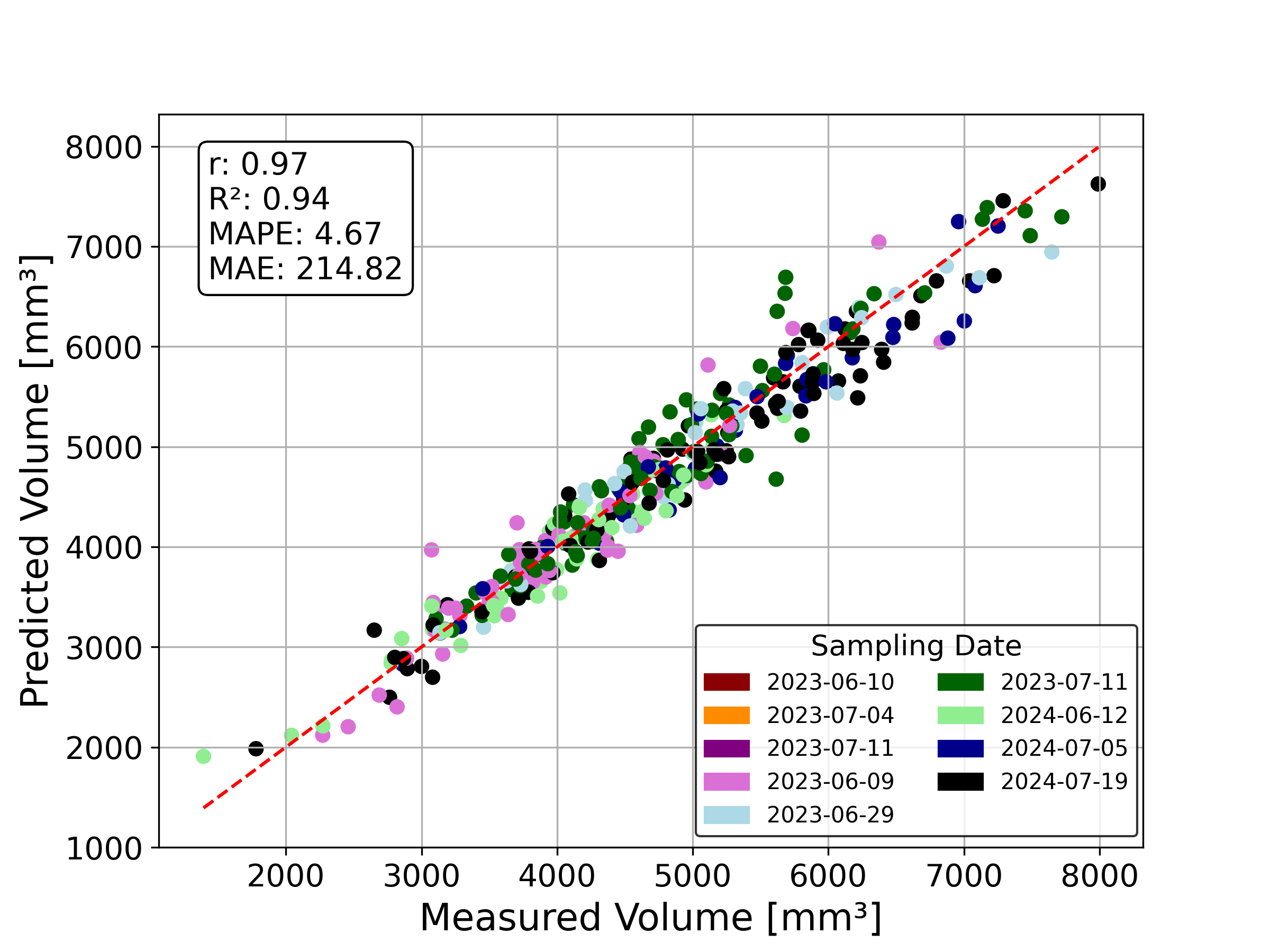}
            \caption{}
        \end{subfigure}
        \caption{Correlation, $R^2$, MAPE, and MAE of measured and estimated volumes of the fine-tuned DINOv3-MLP evaluated on one (a) and on six images (b). Each point represents a single spike of the test set, with color coding indicating the sampling date across the two years. The red line represents a perfect correlation.}
        \label{fig:correlation}
    \end{figure}

    To better understand how the estimation accuracy varies across developmental stages and growing seasons, we visualized the correlation between the estimated and measured volumes evaluated on one and six images (Figure~\ref{fig:correlation}). No discernible pattern or clustering tendency was visible among the sampling groups, indicating the absence of clear outlier groups or group-specific trends. Minor deviations could be observed in some late season samples, likely reflecting increased morphological variability or a smaller number of genotypes exhibiting particularly large volumes in the training dataset.

\begin{figure}
        \vskip 0.2in \centering{\includegraphics[width=\columnwidth]{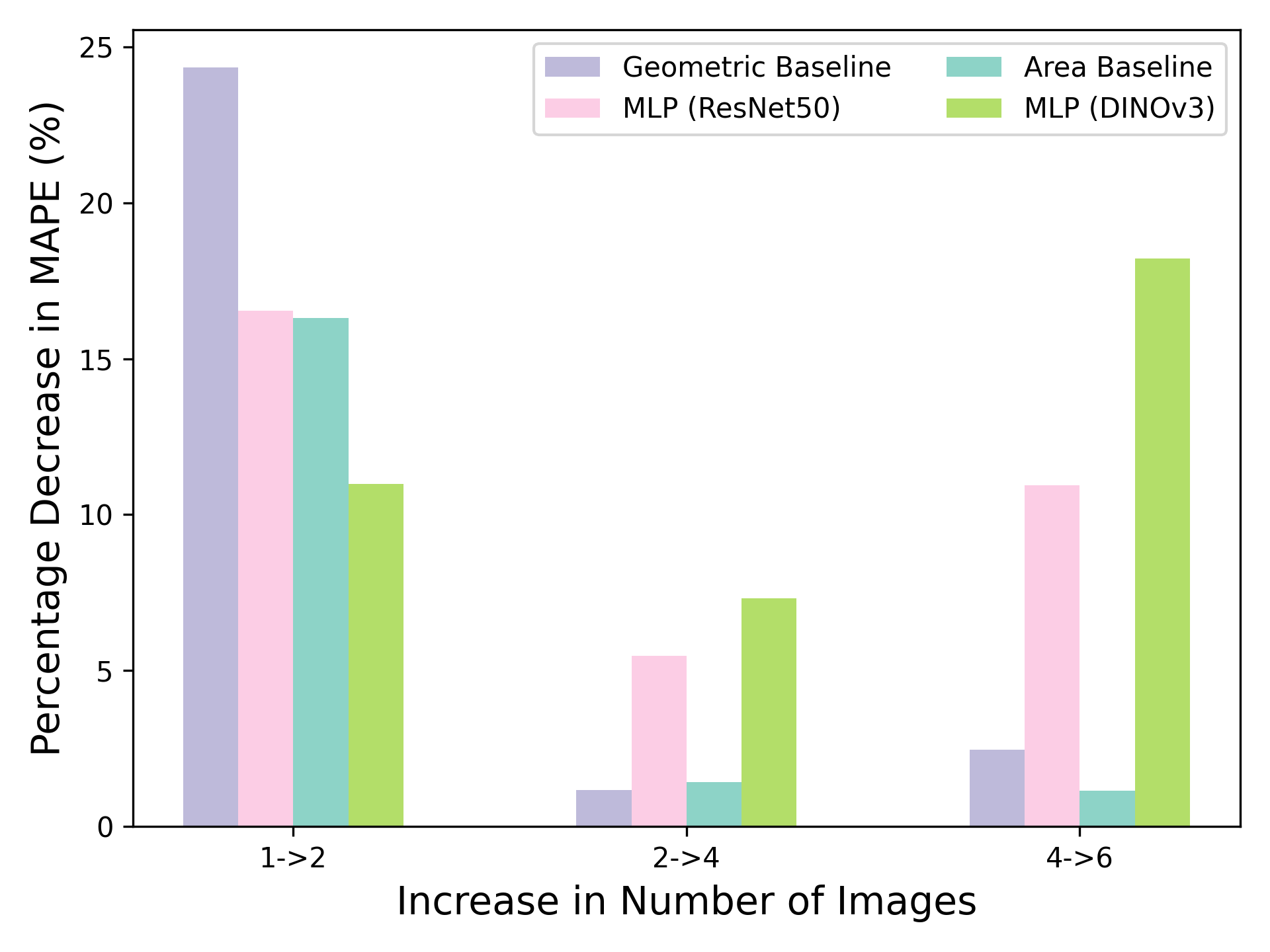}} 
        \caption{Decrease of Mean Absolute Percentage Error of baselines and the neural networks based on different numbers of images.}
        \label{fig:results_images}
        \vskip -0.2in
    \end{figure}

    We investigated how increasing the number of evaluation images affected estimation accuracy, measured by MAPE. We compared the performance of the two baseline models with the best-performing neural networks evaluated on six images, namely MLPs with fine-tuned ResNet50 and DINOv3 (Figure~\ref{fig:results_images}). Although all methods showed improvements with more evaluation images, the performance gains of the baseline methods began to level-off beyond two images (side and front view). Specifically, the MAPE of the area baseline decreased by 16.31\% when increasing from one to two images, but only by 1.43\% from two to four images, and 1.14\% from four to six images. The geometric baseline followed a similar pattern, with reductions of 24.33\%, 1.17\%, and 2.47\%, respectively. In contrast, the two neural networks, particularly the DINOv3-MLP, showed stronger improvements with additional images. The MAPE of the ResNet50-MLP and DINOv3-MLP decreased by 16.54\% and 10.98\% from one to two images, 5.47\% and 7.31\% from two to four images, and 10.95\% and 18.21\% from four to six images, showing a steady reduction in estimation error as the number of input images increases.

    \subsection{Volume Estimation from in-Field Images}
\begin{figure}
        \centering
        \begin{subfigure}[t]{0.48\textwidth}
            \centering
            \includegraphics[width=1\linewidth]{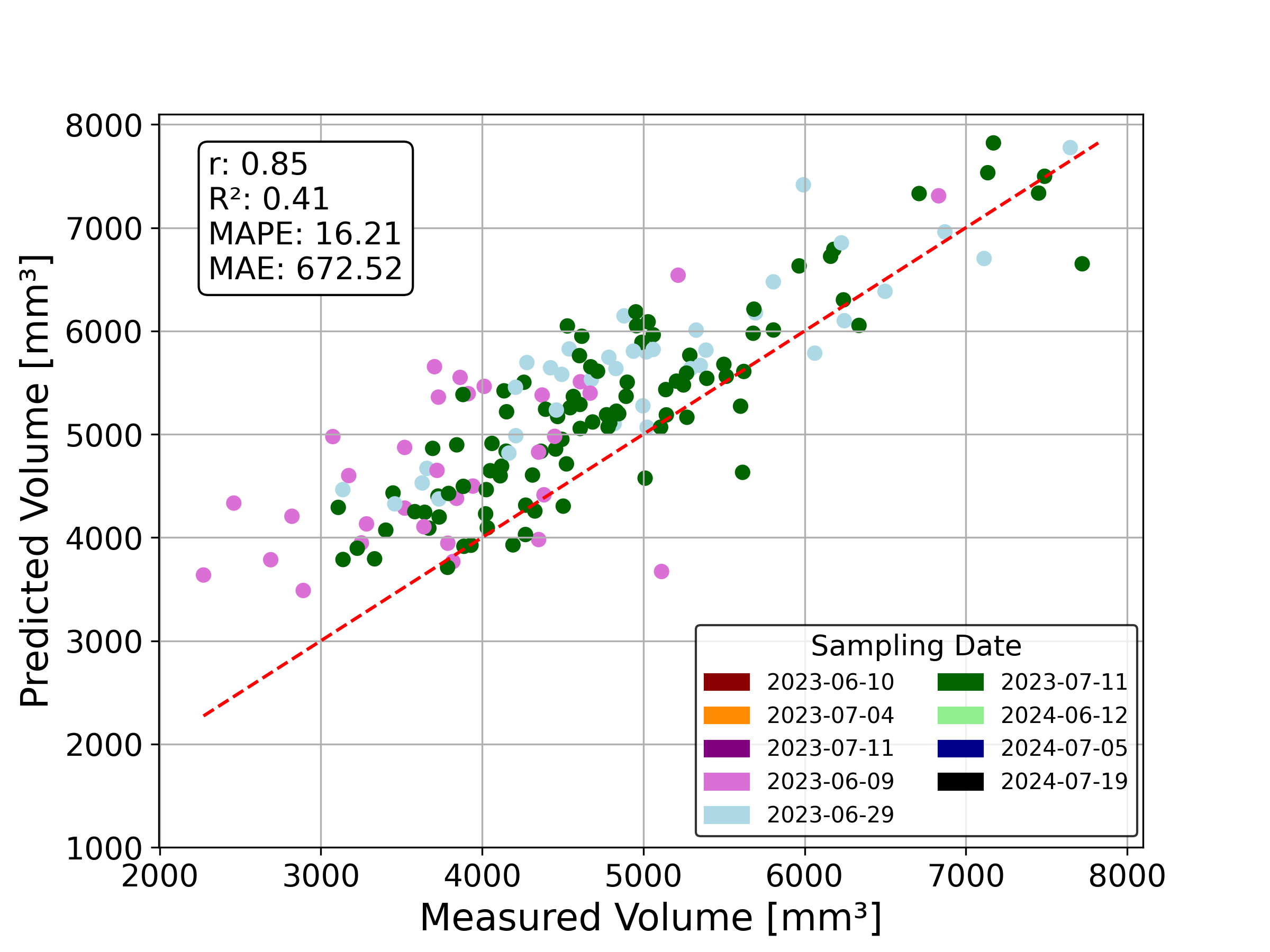}
            \caption{}
        \end{subfigure}
        \vspace{0.02\columnwidth}
        \begin{subfigure}[t]{0.48\textwidth}
            \centering
            \includegraphics[width=1\linewidth]{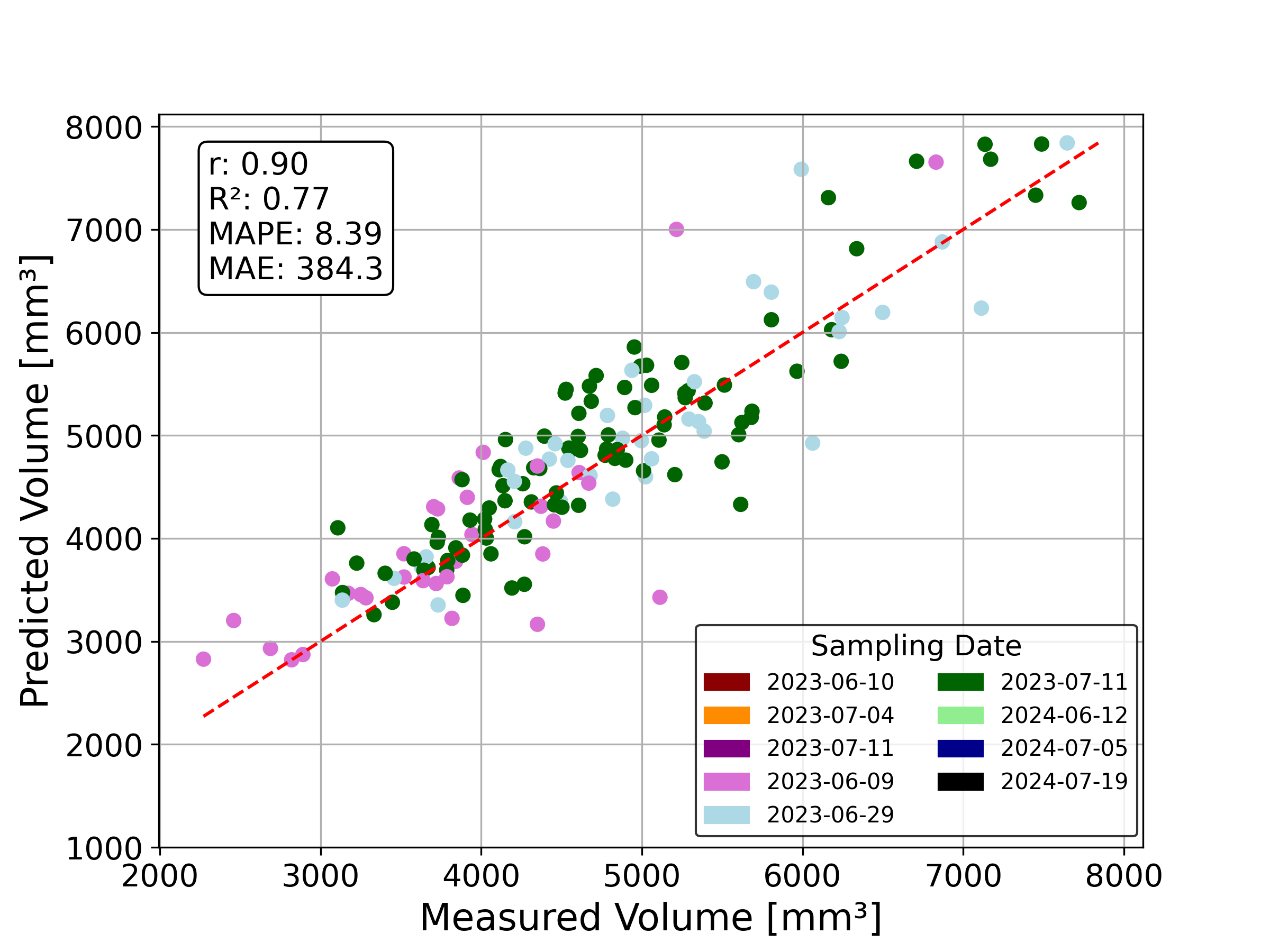}
            \caption{}
        \end{subfigure}
        \caption{Correlation, $R^2$, MAPE, and MAE of measured and estimated volumes of the DINOv3-MLP evaluated on single side-view field images (a). DINOv3-MLP model fine-tuned on single side-view field images (b). Each point represents a single spike of the test set, with color coding indicating the sampling date across the two years. The red line represents a perfect correlation}
        \label{fig:field_image}
    \end{figure}

    Effective plant phenotyping requires methods that are both accurate and time-efficient. When using the imaging setup in the field, manually rotating plants to acquire multiple views is not feasible, especially in the presence of awns. Consequently, we further focused on single side-view volume estimation where awns, if present, were not removed before imaging, and on the best-performing models with fine-tuned backbones. Additional fine-tuning these best-performing single side-view models on field-based images (DINO- and ResNet-based MLPs and LSTMs) resulted in higher estimation accuracies compared to the same single-view models without fine-tuning on field images. Compared to the model trained only on indoor images, MAPE of the fine-tuned DINOv3-MLP decreased from 16.21\% to 8.39\% and MAE from 672.52 $\text{mm}^3$ to 384.30 $\text{mm}^3$. Similarly, for the fine-tuned ResNet50-MLP, MAPE decreased from 16.24\% to 8.35\%, and MAE from 679.65 $\text{mm}^3$ to 393.95 $\text{mm}^3$ (Table~\ref{tab:field_table}). This suggests that the fine-tuned models learnt to account for features such as awns and variable lighting, reducing volume over-estimation (Figure~\ref{fig:field_image}).

    \begin{table*}[htbp]
    \centering
    \caption{Correlation, $R^2$, MAPE, and MAE of the best-performing single side-view models trained on indoor images, and the same models fine-tuned on single side-view field images.}
    \label{tab:field_table}
    \small
        \setlength{\tabcolsep}{3pt}
        \makebox[\linewidth][c]{%
        \begin{tabularx}{1\textwidth}{@{}p{0.23\textwidth}*{8}{>{\centering\arraybackslash}p{0.085\textwidth}}@{}}
    \toprule
        Model    
        & \multicolumn{4}{c}{Without Fine-Tuning on Field Images}
        & \multicolumn{4}{c}{Fine-tuning on Field Images} \\
        \cmidrule(r){2-5} \cmidrule(l){6-9}
        & Corr. & $R^2$ & MAPE (\%) & MAE (mm$^3$) 
        & Corr. & $R^2$ & MAPE (\%) & MAE (mm$^3$) \\
        \midrule
        MLP (ResNet18) & 0.78 & 0.27 & 17.13 & 725.29 & 0.88 & 0.76 & 9.17 & 424.33 \\
        MLP (ResNet50)    & 0.82 & 0.39 & 16.24 & 679.65 & \textbf{0.88} & \textbf{0.77} & \textbf{8.35} & \textbf{393.95} \\
        LSTM (ResNet18)   & 0.76 & 0.40 & 15.09 & 657.22 & 0.88 & 0.74 & 9.22 & 426.78 \\
        LSTM (ResNet50) & 0.73 & 0.34 & 16.48 & 692.81 & 0.87 & 0.74 & 9.51 & 434.94 \\
        \midrule 
        MLP (DINOv2) & 0.80 & 0.51 & 14.29 & 602.52 & 0.89 & 0.77 & 8.87 & 396.63 \\
        MLP (DINOv3)    & 0.85 &  0.41 & 16.21 & 672.52 & \textbf{0.90} &  \textbf{0.77} & \textbf{8.39} & \textbf{384.30} \\
        LSTM (DINOv2)   & 0.75 & 0.38 & 16.31  & 683.93 & 0.85 & 0.66 & 11.04 & 500.23 \\
        LSTM (DINOv3) & 0.85 & 0.43 & 15.46 & 655.96 & 0.89 & 0.71 & 9.55 & 432.50 \\
        \bottomrule                 
    \end{tabularx}
    }
    \end{table*}

    \subsection{Runtime performances}

    We evaluated both the training time (seconds per epoch) and the evaluation time (seconds per spike) of manual volume measurement and the proposed neural networks (Table~\ref{tab:time_comparison}). All models were executed on an NVIDIA RTX 5000 GPU. Manual volume measurements using the 3D scanner required approximately 300 seconds per spike. Training the neural networks required between 0.9 and 2.7 seconds per epoch, using a batch size of 32 and one image per spike, while inference was highly efficient, with evaluation times ranging from 0.0006-0.0013 seconds per spike and per image. CNN-based models trained faster than DINO-based models and showed comparable or slightly faster inference times. Among the ViTs, DINOv3 models trained faster than DINOv2 models, but exhibited slightly slower evaluation times. 

    \begin{table}[htbp]
        \centering
        \caption{Comparison of manual measurement and neural networks in terms of approximated time for training and evaluation. Training and evaluation time was measured based on one image per spike.}
        \label{tab:time_comparison}
            \begin{tabularx}{\columnwidth}{@{}
            >{\raggedright\arraybackslash}p{0.45\columnwidth}
            >{\raggedright\arraybackslash}p{0.25\columnwidth}
            >{\raggedright\arraybackslash}p{0.30\columnwidth}
            @{}
            }
            \toprule
            \textbf{Method} & \textbf{Training Time (s/epoch)} & \textbf{Evaluation Time (s/spike)}
            \\
            \midrule
            Structured light scanner & - & 300 \\
            MLP (DINOv2) & 2.6 & 0.0011\\  
            MLP (DINOv3)& 2.3 & 0.0013 \\  
            LSTM (DINOv2)& 2.7 & 0.0011 \\ 
            LSTM (DINOv3)& 2.4 & 0.0013 \\ 
            MLP (ResNet18)  & 0.9 & 0.0006 \\ 
            MLP (ResNet50)  & 1.8 & 0.0011 \\ 
            LSTM (ResNet18)  & 1.0 & 0.0006 \\ 
            LSTM (ResNet50)  & 1.9 & 0.0011 \\ 
            MLP (ResNet50, field-images)  & 1.0 & 0.0011 \\
            MLP (DINOv3, field-images)  & 1.2 & 0.0013 \\
            \bottomrule
        \end{tabularx}
    \end{table}

\section{Discussion}

Fruiting capacity and the derived fruiting efficiency are two highly relevant traits towards a better understanding of genotype or environment-specific yield formation of wheat. The trait will be relevant for a wide range of applications, including breeding, variety testing, crop physiology, or precision agriculture. However, accurate spike volume estimation within wheat canopies remains challenging. Spikes are often partially occluded, even when images are captures from multiple viewing angles, and varietal differences in stem length lead to spikes being distributed across different canopy layers. Moreover, awns will introduce a strong bias in apparent estimation, particularly in dense canopies, resulting in noisy and biased estimations. The aim of this work was therefore to provide a novel, fast and accurate high-throughput volume estimation pipeline from RGB images acquired indoors or outdoors,  without requiring camera calibration or depth information. To this end, we benchmark models with differing assumptions and analyse their performance, limitations, and robustness with respect to object shape, and view count.

\subsection{Effect of Object Shape on Model Performance} 
\begin{figure}
        \centering
        \begin{subfigure}[t]{0.48\textwidth}
            \centering
            \includegraphics[width=1\linewidth]{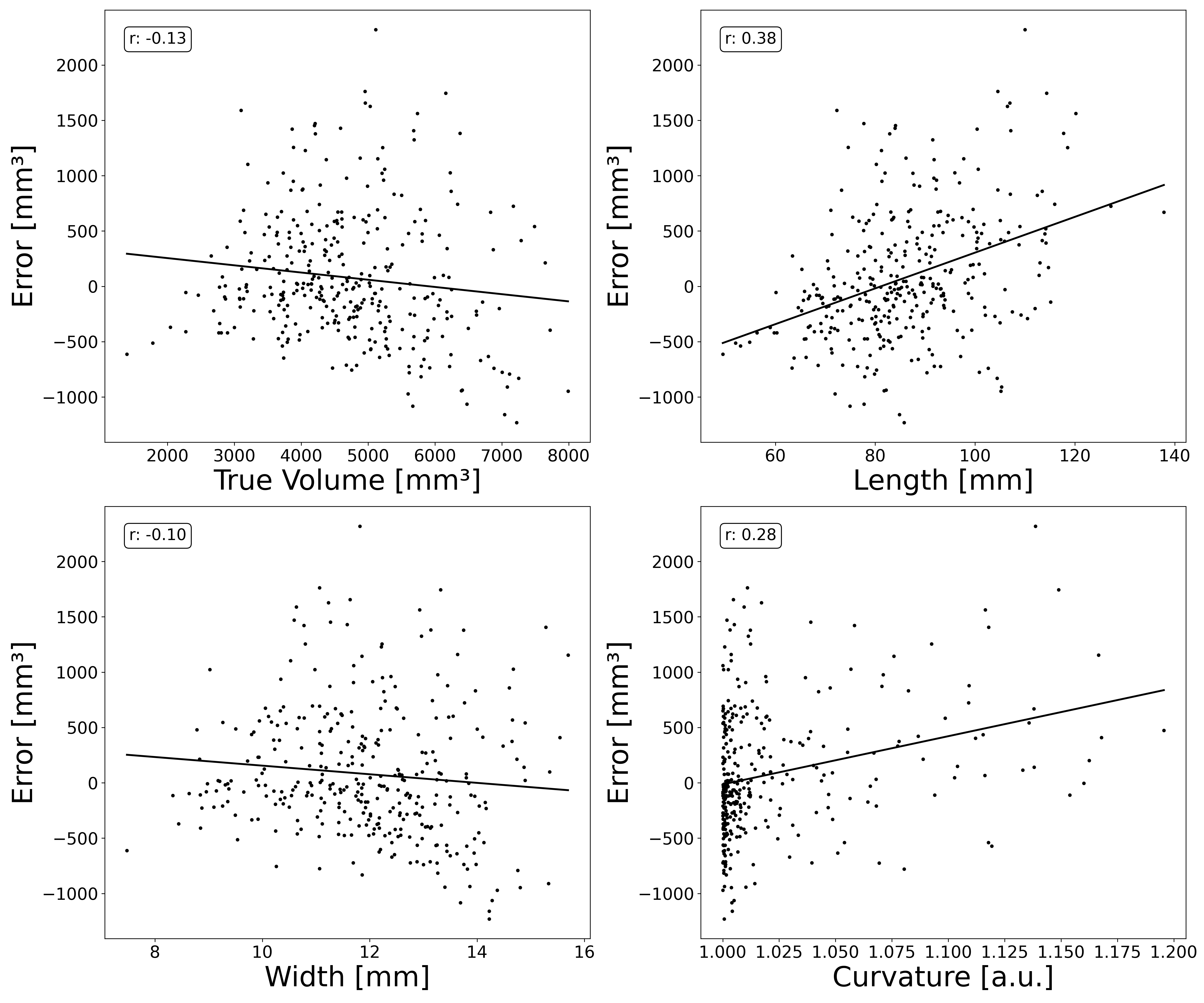}
            \caption{}
        \end{subfigure}
        \hfill
        \begin{subfigure}[t]{0.48\textwidth}
            \centering
            \includegraphics[width=1\linewidth]{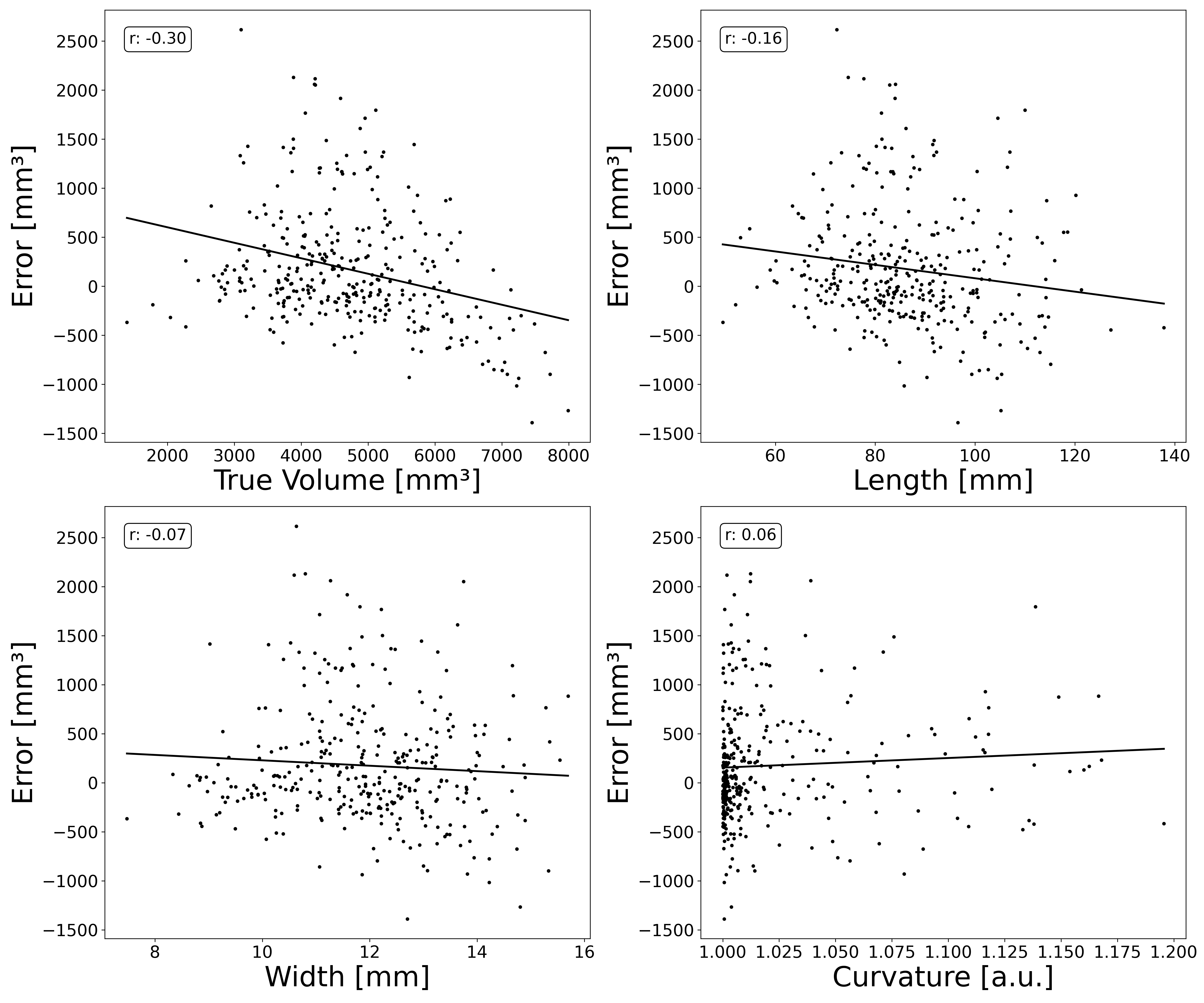}
            \caption{}
        \end{subfigure}
        \caption{Scatterplots of individual testset spikes illustrating the relationship between the signed area baseline error (a) and the geometric baseline error (b) and measured spike morphology (volume, length, width, and curvature).}
        \label{fig:error_area_geo}
    \end{figure}

Across all image counts, neural networks consistently outperformed both the area-based and geometric baselines in terms of correlation, $R^2$, MAPE, and MAE. This performance gap arised from fundamental differences in how these methods approximated the underlying 3D geometry of spikes. Both baselines methods assumed an approximately quadratic, yet in practice nearly linear, relationship between either the number of foreground pixels (area baseline) or the integrated radius function (geometric baseline) and the measured volume (Appendix Table~\ref{tab:baseline_fitted}, Appendix Figure~\ref{fig:train_curves_area}, Appendix Figure~\ref{fig:train_curves_geo}). To better understand the resulting errors, we extracted spike-level traits (length, width, curvature) from the 3D scans following \citep{zhangWheat3DGSInfield3D2025} and examined how these traits, along with measured volume, related to the signed error derived from volume estimation using all six images. A key limitation of the area baseline is that it treated all pixels equally, ignoring that spike volume scales linearly with length but quadratically with width. Consequently, two spikes with the same segmented 2D area, a long thin spike and a short thick spike, received identical volume estimates, even though the thick spike had a larger true volume. This led to systematic errors: long spikes tended to be overestimated, while short thick spikes were underestimated (Figure~\ref{fig:error_area_geo}a). 

\begin{figure}
        \vskip 0.2in \centering{\includegraphics[width=\columnwidth]{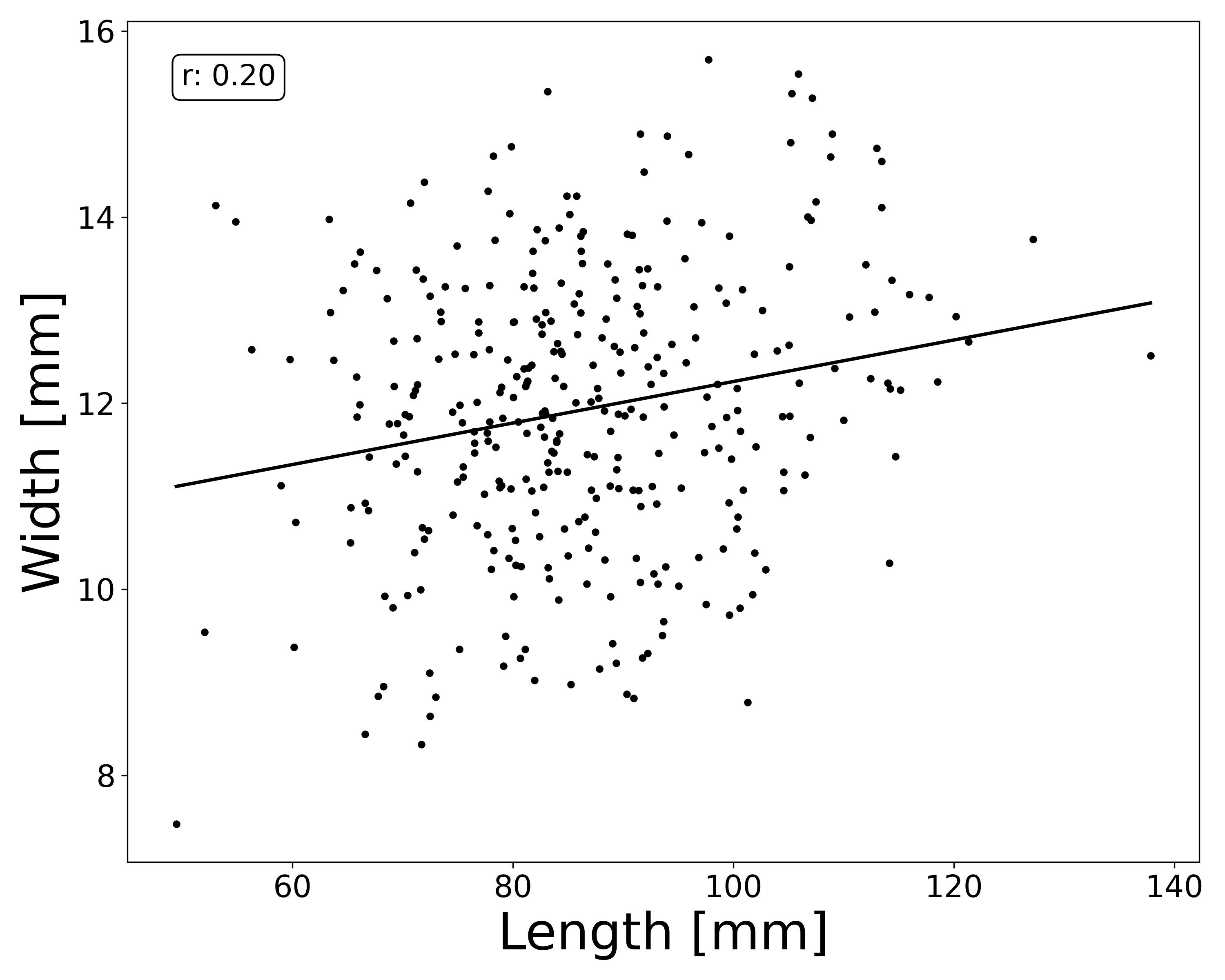}} 
        \caption{Scatterplot of individual testset spikes illustrating the relationship between the width and the length of the spikes.}
        \label{fig:width_length}
        \vskip -0.2in
    \end{figure}

This reasoning assumes that shorter spikes tend to be thicker, and longer spikes thinner. The length–width correlation in our dataset is positive but weak (r=0.21; Figure~\ref{fig:width_length}), which indicates that there are long, thin and short, thick spikes present in the dataset. Short, thin and long, thick spikes do not result in over- or underestimation. As a result, the overall correlation between spike length and signed error remained moderate (r=0.38). Similarly, width was weakly correlated with MAE: volume of thinner spikes was slightly overestimated, while the volume of thicker spikes was slightly underestimated. This might reflect the near-linear scaling of pixel area, despite the true quadratic dependence of volume on width. Finally, spike curvature further amplified volume overestimation, as curved spikes might have exposed a larger projected area to the camera  without increasing physical volume.

The geometric baseline mitigated some of these issues by explicitly modelling the spike as a sequence of discs with circular cross-sections. Thicker spikes therefore received higher volume estimates than long, thin spikes with the same area, reducing the underestimation of short, thick spikes and, in some cases, even slightly overcompensating (Figure~\ref{fig:error_area_geo}b). However, its absolute errors remained larger than those of the area baseline. Large spikes tended to be overestimated and small spikes underestimated, a pattern that originated from the training set: the geometric baseline overestimated large spikes during calibration (Appendix~\ref{fig:train_curves_geo}), resulting in a lower fitted scaling factor and consequently to systematic underestimation in the test set. 

These error patterns suggest that the limitations of the baseline methods are closely tied to the underlying geometry of the spikes. Similar observations have been made in other domains, where object shape influences the success of simple volume estimation techniques. Volume estimation of irregularly shaped tomatoes using geometric features has resulted in errors that were comparable to our baseline errors \citep{uluisikImageProcessingBased2018}. In contrast, more geometrically regular objects such as oranges tended to yield higher estimation accuracy \citep{khojastehnazhandDeterminationOrangeVolume}. For highly axis-symmetric shapes such as eggs, 2D area-based methods have even out-performed neural networks in terms of $R^2$ score \citep{soltaniEggVolumePrediction2015}. A similar issue arises with wheat spikes. While they may appear roughly axis-symmetric from a side view, the spikelets are positioned in an alternating way along the main axis, and vary in shape and size. This results in an asymmetrical radius on either side of the main axis, violating the geometric assumptions of the geometric baseline (e.g., axial symmetry) and increased estimation difficulty. An example of a spike with particularly poor volume estimation under the geometric baseline method is shown in Appendix Figure~\ref{fig:high_error_spike}. The spike appears compressed, artificially increasing width and leading to volume overestimation. While both baselines struggle to accurately estimate volume for such spikes, the fine-tuned DINOv3-MLP model can overcome this issue (Appendix Figure~\ref{fig:genotype_error}). Integrating viewing-angle information by fitting separate calibration curves for each number of images, as demonstrated in apple volume estimation with shape-specific models \citep{iqbalVolumeEstimationApple2011}, improved the estimation accuracy of the baselines but could not match the performance of neural networks. In contrast, the neural networks made no prior assumptions about the object's geometry and did not depend on a predefined scaling relationship between estimated and measured volume. Instead, they learnt complex spatial relationships directly from raw pixel data. During training, random view sampling exposed the models to all six images per spike, allowing them to exploit information from multiple views even when evaluated on fewer images at test time. By contrast, the baseline methods were limited to a single scaling factor estimated from the training data. However, important differences were also observed among the neural network architectures themselves.

\subsection{Volume estimation based on neural networks}
LSTMs and Transformers have shown effective results in domains such as speech recognition \cite{orosooTransformingEnglishLanguage2025, dongSpeechTransformerNoRecurrenceSequencetoSequence2018} or language translation \citep{sutskeverSequenceSequenceLearning2014, wangLearningDeepTransformer2019}. Furthermore, they have also shown good performances in image or video classification tasks \citep{islamCombinedDeepCNNLSTM2020, chenCNNLSTMModelRecognizing2023, al-jabery4SelectedApproaches2020, salehinRealTimeMedicalImage2023, saikiaHybridCNNLSTMModel2022, djoumessiHybridFullyConvolutional2025}, achieving state-of-the-art performance. In these approaches, similar to ours, features were first extracted from images and then fed into an LSTM or Transformer to model sequential dependencies across the image sequence. Code examples show that the final state of the LSTM is usually taken for classification or regression tasks \citep{chrisvanderwethMLToolkitRNNText}, while in tasks such as sequence labelling, it is recommended to use all hidden outputs for estimation \citep{alexandruSequenceModelsTutorial}. We adopted this approach to our LSTMs and Transformers and could show that training the model on sequences of six images while estimating the spike volume based on fewer images leads to high errors when relying solely on the final time step. By adapting the model's output to incorporate all time steps during training, the network learns to improve volume estimation at each step. Since the measured volume is the same for all the six images per plant, the model's ability to estimate the volume based on each single image is increased, ultimately allowing it to make accurate estimations from a single image during evaluation, resulting in both improved performance and faster evaluation. When the backbone models DINOv2 and DINOv3 were fully frozen, our best-performing LSTM outperformed both the MLPs and the Transformers. While the LSTMs and the MLPs achieved comparable performance, the Transformer-based models significantly showed lower estimation accuracies across the training, validation, and test set (Appendix Table~\ref{tab:metrics_mlp_lstm_transformer}). A likely reason is that Transformers are inherently more data-hungry, making volume estimation particularly challenging when only limited training data is available, or that they require structurally complex datasets to outperform simpler architectures. 

Although conceptually simple, fully frozen feature-extraction approaches depend entirely on the pre-computed DINO features of the backbone models. These backbone models were trained for general visual repesentation learning, not specifically for estimating spike volume. If the extracted representation does not encode the exact cues needed for estimating volume, such as thickness or curvature, the downstream models must work with suboptimal embeddings. In contrast, fine-tuning DINOv2, DINOv3  led to consistently superior performance. The early layers of these models extract low-level features such as edges, and color gradients, whereas deeper layers generate high-level features tuned for the original pre-training objective (e.g., semantic clustering). Fine-tuning the last transformer block presumably allowed these high-level features to better capture geometric cues relevant for spike volume estimation, explaining why the fine-tuned models outperformed the fully frozen backbone models, despite sharing the same frozen early backbone layers. 

When focusing on the fine-tuned backbone models, clear differences emerged between the CNNs and the ViTs, DINOv2 and DINOv3. CNNs have long dominated computer vision tasks due to their strong inductive biases, such as locality, and translation equivariance, which make them highly effective at extracting spatial patterns. In contrast, ViTs, originally introduced from natural language processing, lack these built-in inductive biases but compensate through global self-attention mechanisms, enabling them to model long-range dependencies and capture richer contextual relationships. When trained on large-scale datasets, ViTs have been shown to outperformed CNNs across a wide range of vision tasks, including image classification \citep{dosovitskiyImageWorth16x162021}, object detection \citep{carionEndtoEndObjectDetection2020a}, or semantic segmentation \citep{ranftlVisionTransformersDense2021}. Consistent with these findings, our fine-tuned ViT models, especially DINOv3, outperformed fine-tuned ResNet18 and ResNet50 across all evaluated image counts. While the deeper ResNet50 achieved higher estimation accuracy than ResNet18, both CNN backbones remained inferior to the fine-tuned ViTs, DINOv2 and DINOv3. Notably, DINOv3 was inferior to DINOv2 when the backbones were kept fully frozen, but outperformed DINOv2 once the backbones were fine-tuned. One possible explanation for this pattern is that DINOv3 introduces architectural and training changes designed to excel at high-level semantic representation learning \citep{simeoniDINOv32025}. These improvements do not necessarily translate into better performance in fine-grained geometric estimation tasks, such as estimating spike volume when the backbone is fully frozen. In contrast, the representations learnt by DINOv2 appeared to align more closely with the geometric and structural cues relevant to this dataset. However, once fine-tuned, the greater representational capacity of DINOv3 enabled more effective adaptation to task-specific requirements, resulting in superior performance. In related work, a crop-specific foundation model, FoMo4Wheat, has been shown to outperform general-domain pre-trained models on certain wheat-related tasks \citep{hanFoMo4WheatReliableCrop2025}. While the authors report that DINOv2 features are suboptimal for wheat image tasks, as evidenced by degraded feature visualisations, these conclusions are drawn from tasks that differ from volume estimation. In contrast, our results demonstrated that both DINOv2 and DINOv3 outperformed FoMo4Wheat for wheat spike volume estimation, indicating that general-domain foundation models can be highly effective when fine-tuned for geometry-driven regression tasks. 

\begin{figure}
        \centering
        \begin{subfigure}[t]{0.48\textwidth}
            \centering
            \includegraphics[width=1\linewidth]{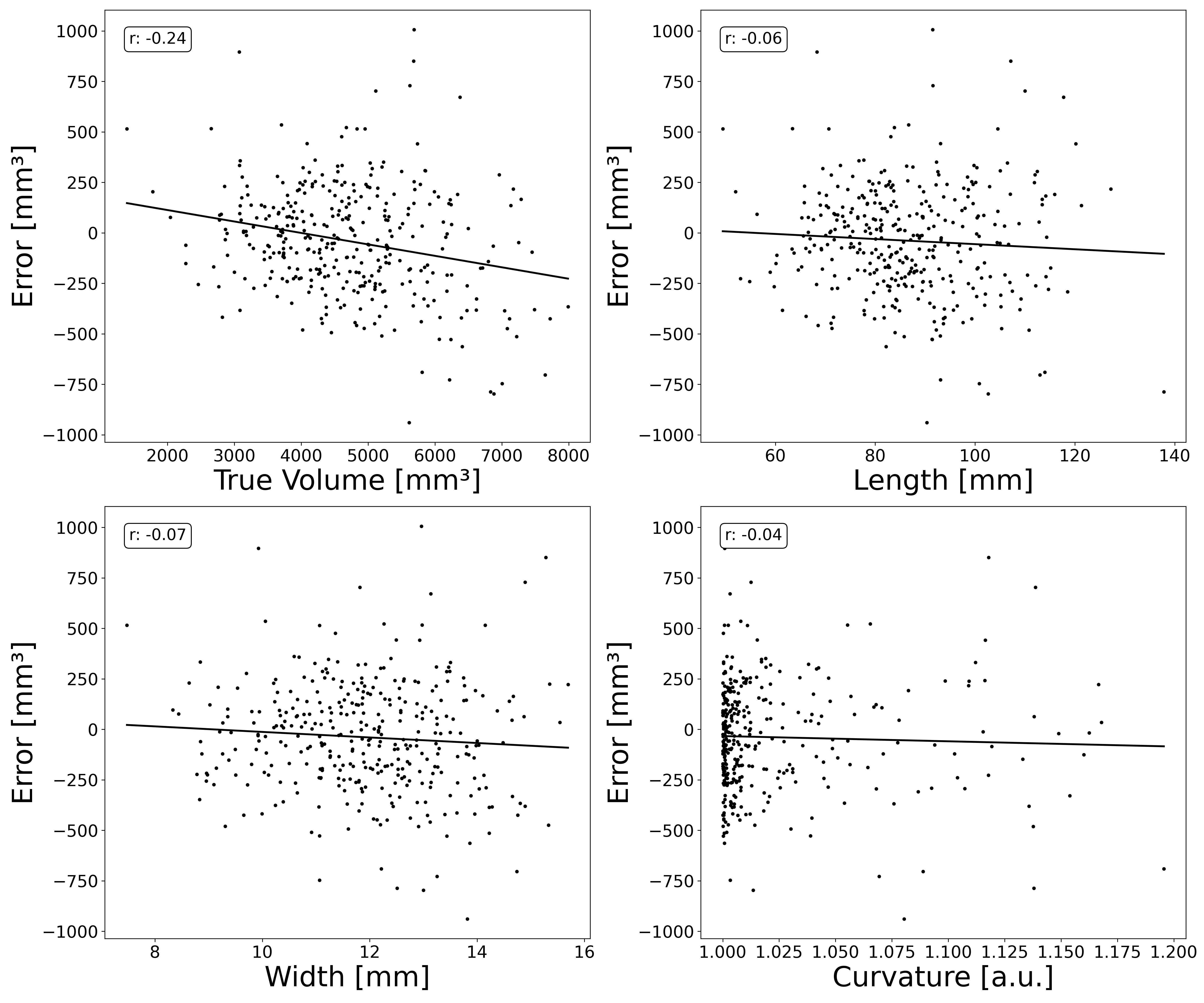}
        \end{subfigure}
        \caption{Scatterplots of individual testset spikes illustrating the relationship between the fine-tuned DINOv2 error and measured spike morphology. Panel (a) shows the absolute error, while panel (b) shows the signed error, each plotted against measured volume, length, width, and curvature.}
        \label{fig:error_NN}
    \end{figure}

Notably, after backbone fine-tuning, MLPs outperformed the deep-supervised LSTMs. While the LSTMs achieved higher performance when using suboptimal features from fully frozen backbones, i.e., benefiting from deep supervision and their ability to aggregate information across multiple views, the simpler MLPs were sufficient once the backbone models were fine-tuned. The signed error of the best-performing fine-tuned DINOv3-MLP model revealed a slight volume overestimation of small spikes and underestimation of large spikes, while showing almost no correlation with spike length, width, or curvature (Figure~\ref{fig:error_NN}). While these results demonstrate the effectiveness of deep learning for spike volume estimation, they also raise an important question: how do different models behave when the number of available views changes?

\subsection{Effect of image count in evaluation} 
While all methods benefit from increasing the number of evaluation images, the magnitude of improvement differs substantially between the baselines and the neural networks. The baseline models gain most when moving from one side-view image to two (side and front view), after which performance improvements levelled off. In contrast, the neural networks, especially the best-performing ResNet50-MLP and DINOv3-MLP, continued to benefit from additional views, showing a more consistent reduction in estimation error as the number of images increased. This suggests that the neural networks were capable of integrating complementary geometric cues from multiple view points to build a more complete internal representation of object volume. However, the improved performance with additional views also underscores the importance of large and diverse training datasets. The models' ability to generalise across a broad range of spike geometries depends on being exposed to sufficient variability during training. Thus, the model can learn the inherent geometric relationship between the different views in the present set of genotypes. However, these neural networks might not generalise to genotypes that deviate strongly in their appearance from those represented in the training dataset, whereas the two baselines rely on explicit pixel-based and geometry-based measurements and may therefore be more robust to previously unseen genotypes. 

\subsection{Fine-tuning on Field Images} 
Although the neural networks achieved a low estimation error evaluated on a single side-view indoor image, the same models did not generalise well to single side-view field images. Fine-tuning the model using field images from the training dataset reduced volume over-estimation. This over-estimation was likely due to the presence of awns and varying lighting conditions in the field setting. In contrast, the baseline methods are not directly applicable to field images, as the presence or absence of awns varies across genotypes and would require very precise artificial removal during pre-processing to ensure consistent geometric and pixel-based measurements. In practice, reliably removing thin structures such as awns without simultaneously removing other fine spike structures is challenging. In addition, further work could explore including other crops, such as barley, to support generalisation across different spike morphologies.

\section{Conclusion}

Fruiting efficiency is a promising trait to improve wheat yield, but requires to destructively measure the spike dry weight at flowering. We showed that the spike volume and dry weight at flowering are closely related. To calibrate and validate this trait, rapid in-field measurements are required for which reason we tested different approaches. Using neural networks, we achieve improvements in volume estimation over our baselines that rely on area projection or geometric approximations. Our results reveal that object shape plays a critical role in estimation performance, with irregular and asymmetric geometries, such as those of wheat spikes, posing challenges to the geometric model assuming axis symmetry for spikes. In contrast, neural networks do not depend on pre-defined geometric assumptions and can learn complex visual cues directly from images. Fully frozen backbone approaches were limited by its reliance on fixed representations that were not specifically tailored to the estimation of the spike volume. As a result, the downstream models had to learn from suboptimal features. In contrast, fine-tuned models performed substantially better, allowing to refine their high-level features leading to markedly improved volume estimation. Furthermore, we show that fine-tuning on single-view field images enables the model to retain predictive accuracy in field environments, providing a method to measure newly defined fruiting capacity as the summed volume per area of individual spikes.

\section{Data availability}
The full dataset, the source code and the trained models will be made publicly available through the ETH research collection after publication. 

\section{CRediT authorship contribution statement}

\textbf{Olivia Zumsteg}: Writing – review \& editing, Writing - original draft, Methodology, Software, Validation, Visualization.
\textbf{Nico Graf}: Writing – review \& editing, Software, Visualization, Methodology.
\textbf{Aaron Häusler}: Writing – review \& editing, Software, Visualization, Methodology.
\textbf{Norbert Kirchgessner}: Writing – review \& editing, Methodology, Software.
\textbf{Nicola Storni}: Writing – review \& editing, Methodology, Software.
\textbf{Lukas Roth}: Writing – review \& editing, Methodology, Supervision.
\textbf{Andreas Hund}: Writing – review \& editing, Methodology, Conceptualization, Supervision, Funding acquisition.

\section{Ethical statement}
This study does not involve any experiments on humans or animals, and no ethical approval was required.

\section{Declaration of competing interest}
The authors declare that they have no known competing financial interests or personal relationships that could have appeared to influence the work reported in this paper.

\section{Acknowledgement}
We thank Konstantina Gkouramani, Mohammad Hosein Fakhrosae, and  Patrick Fitzi, for their invaluable support with data collection. We are grateful to Prof. Achim Walter for providing funding for this work.

\bibliographystyle{elsarticle-num}
\bibliography{volume_estimation}

@incollection{al-jabery4SelectedApproaches2020,
  title = {4 - {{Selected}} Approaches to Supervised Learning},
  booktitle = {Computational {{Learning Approaches}} to {{Data Analytics}} in {{Biomedical Applications}}},
  author = {{Al-jabery}, Khalid K. and {Obafemi-Ajayi}, Tayo and Olbricht, Gayla R. and Wunsch II, Donald C.},
  editor = {{Al-jabery}, Khalid K. and {Obafemi-Ajayi}, Tayo and Olbricht, Gayla R. and Wunsch II, Donald C.},
  year = {2020},
  month = jan,
  pages = {101--123},
  publisher = {Academic Press},
  doi = {10.1016/B978-0-12-814482-4.00004-8},
  urldate = {2025-06-08},
  isbn = {978-0-12-814482-4}
}

@misc{alexandruSequenceModelsTutorial,
  title        = {Sequence Models Tutorial},
  author       = {Alexandru, Andrei},
  year         = {2017},
  url          = {https://github.com/pytorch/tutorials/blob/main/beginner_source/nlp/sequence_models_tutorial.py},
  note         = {Accessed on 2025-06-19}
}

@article{andereggThermalImagingCan2024,
  title = {Thermal Imaging Can Reveal Variation in Stay-Green Functionality of Wheat Canopies under Temperate Conditions},
  author = {Anderegg, Jonas and Kirchgessner, Norbert and Aasen, Helge and Zumsteg, Olivia and Keller, Beat and Zenkl, Radek and Walter, Achim and Hund, Andreas},
  year = {2024},
  month = jun,
  journal = {Frontiers in Plant Science},
  volume = {15},
  publisher = {Frontiers},
  issn = {1664-462X},
  doi = {10.3389/fpls.2024.1335037},
  urldate = {2025-04-09},
  langid = {english}
}

@article{chenCNNLSTMModelRecognizing2023,
  title = {{{CNN-LSTM Model}} for {{Recognizing Video-Recorded Actions Performed}} in a {{Traditional Chinese Exercise}}},
  author = {Chen, Jing and Wang, Jiping and Yuan, Qun and Yang, Zhao},
  year = {2023},
  journal = {IEEE Journal of Translational Engineering in Health and Medicine},
  volume = {11},
  pages = {351--359},
  issn = {2168-2372},
  doi = {10.1109/JTEHM.2023.3282245},
  urldate = {2025-06-08}
}

@misc{chrisvanderwethMLToolkitRNNText,
  author       = {Chris van der Weth},
  title        = {{ML-Toolkit: RNN Text Classifier in PyTorch}},
  year         = {2019},
  howpublished = {\url{https://github.com/chrisvdweth/ml-toolkit/blob/master/pytorch/models/text/classifier/rnn.py}},
  note         = {Accessed on June 19, 2025}
}

@article{dassotUseTerrestrialLiDAR2011,
  title = {The Use of Terrestrial {{LiDAR}} Technology in Forest Science: Application Fields, Benefits and Challenges},
  shorttitle = {The Use of Terrestrial {{LiDAR}} Technology in Forest Science},
  author = {Dassot, Mathieu and Constant, Thi{\'e}ry and Fournier, Meriem},
  year = {2011},
  month = aug,
  journal = {Annals of Forest Science},
  volume = {68},
  number = {5},
  pages = {959--974},
  issn = {1297-966X},
  doi = {10.1007/s13595-011-0102-2},
  urldate = {2025-06-19},
  langid = {english}
}

@article{dehaisTwoView3DReconstruction2017,
  title = {Two-{{View 3D Reconstruction}} for {{Food Volume Estimation}}},
  author = {Dehais, Joachim and Anthimopoulos, Marios and Shevchik, Sergey and Mougiakakou, Stavroula},
  year = {2017},
  month = may,
  journal = {IEEE Transactions on Multimedia},
  volume = {19},
  number = {5},
  pages = {1090--1099},
  issn = {1941-0077},
  doi = {10.1109/TMM.2016.2642792},
  urldate = {2024-01-04}
}

@article{gaoArchitectureWheatInflorescence2019,
  title = {Architecture of {{Wheat Inflorescence}}: {{Insights}} from {{Rice}}},
  shorttitle = {Architecture of {{Wheat Inflorescence}}},
  author = {Gao, Xin-Qi and Wang, Ning and Wang, Xiu-Ling and Zhang, Xian Sheng},
  year = {2019},
  month = sep,
  journal = {Trends in Plant Science},
  volume = {24},
  number = {9},
  pages = {802--809},
  issn = {1360-1385},
  doi = {10.1016/j.tplants.2019.06.002},
  urldate = {2025-06-19}
}

@article{graikosSingleImageBasedFood2020,
  title = {Single {{Image-Based Food Volume Estimation Using Monocular Depth-Prediction Networks}}},
  author = {Graikos, Alexandros and Charisis, Vasileios and Iakovakis, Dimitrios and Hadjidimitriou, Stelios and Hadjileontiadis, Leontios},
  year = {2020},
  journal = {Universal Access in Human-Computer Interaction. Applications and Practice},
  volume = {12189},
  pages = {532--543},
  doi = {10.1007/978-3-030-49108-6_38},
  urldate = {2023-10-27},
  langid = {english}
}

@article{haiderWhatCanWe2022,
  title = {What {{Can We Learn}} from {{Depth Camera Sensor Noise}}?},
  author = {Haider, Azmi and {Hel-Or}, Hagit},
  year = {2022},
  month = jan,
  journal = {Sensors},
  volume = {22},
  number = {14},
  pages = {5448},
  publisher = {Multidisciplinary Digital Publishing Institute},
  issn = {1424-8220},
  doi = {10.3390/s22145448},
  urldate = {2025-06-16},
  copyright = {http://creativecommons.org/licenses/by/3.0/},
  langid = {english}
}

@book{hansardTimeofFlightCamerasPrinciples2013,
  title = {Time-of-{{Flight Cameras}}: {{Principles}}, {{Methods}} and {{Applications}}},
  shorttitle = {Time-of-{{Flight Cameras}}},
  author = {Hansard, Miles and Lee, Seungkyu and Choi, Ouk and Horaud, Radu},
  year = {2013},
  series = {{{SpringerBriefs}} in {{Computer Science}}},
  publisher = {Springer},
  address = {London},
  doi = {10.1007/978-1-4471-4658-2},
  urldate = {2025-06-02},
  copyright = {https://www.springernature.com/gp/researchers/text-and-data-mining},
  isbn = {978-1-4471-4657-5 978-1-4471-4658-2},
  langid = {english}
}

@inproceedings{iqbalVolumeEstimationApple2011,
  title = {Volume Estimation of Apple Fruits Using Image Processing},
  booktitle = {2011 {{International Conference}} on {{Image Information Processing}}},
  author = {Iqbal, S. {\relax Md}. and Gopal, A. and Sarma, A. S. V.},
  year = {2011},
  month = nov,
  pages = {1--6},
  doi = {10.1109/ICIIP.2011.6108909},
  urldate = {2025-03-21}
}

@article{islamCombinedDeepCNNLSTM2020,
  title = {A Combined Deep {{CNN-LSTM}} Network for the Detection of Novel Coronavirus ({{COVID-19}}) Using {{X-ray}} Images},
  author = {Islam, Md. Zabirul and Islam, Md. Milon and Asraf, Amanullah},
  year = {2020},
  month = jan,
  journal = {Informatics in Medicine Unlocked},
  volume = {20},
  pages = {100412},
  issn = {2352-9148},
  doi = {10.1016/j.imu.2020.100412},
  urldate = {2025-06-08}
}

@article{jimenez-berniHighThroughputDetermination2018,
  title = {High {{Throughput Determination}} of {{Plant Height}}, {{Ground Cover}}, and {{Above-Ground Biomass}} in {{Wheat}} with {{LiDAR}}},
  author = {{Jimenez-Berni}, Jose A. and Deery, David M. and {Rozas-Larraondo}, Pablo and Condon, Anthony (Tony) G. and Rebetzke, Greg J. and James, Richard A. and Bovill, William D. and Furbank, Robert T. and Sirault, Xavier R. R.},
  year = {2018},
  month = feb,
  journal = {Frontiers in Plant Science},
  volume = {9},
  publisher = {Frontiers},
  issn = {1664-462X},
  doi = {10.3389/fpls.2018.00237},
  urldate = {2025-06-19},
  langid = {english}
}

@article{khojastehnazhandDeterminationOrangeVolume,
  title = {Determination of Orange Volume and Surface Area Using Image Processing Technique},
  author = {Khojastehnazhand, M and Omid, M and Tabatabaeefar, A},
  year = {2009},
  journal = {Int. Agrophys.},
  langid = {english}
}

@article{kirchgessnerETHFieldPhenotyping2016,
  title = {The {{ETH}} Field Phenotyping Platform {{FIP}}: A Cable-Suspended Multi-Sensor System},
  shorttitle = {The {{ETH}} Field Phenotyping Platform {{FIP}}},
  author = {Kirchgessner, Norbert and Liebisch, Frank and Yu, Kang and Pfeifer, Johannes and Friedli, Michael and Hund, Andreas and Walter, Achim},
  year = {2016},
  month = feb,
  journal = {Functional plant biology: FPB},
  volume = {44},
  number = {1},
  pages = {154--168},
  issn = {1445-4416},
  doi = {10.1071/FP16165},
  langid = {english},
  pmid = {32480554}
}

@misc{kirillovSegmentAnything2023,
  title = {Segment {{Anything}}},
  author = {Kirillov, Alexander and Mintun, Eric and Ravi, Nikhila and Mao, Hanzi and Rolland, Chloe and Gustafson, Laura and Xiao, Tete and Whitehead, Spencer and Berg, Alexander C. and Lo, Wan-Yen and Doll{\'a}r, Piotr and Girshick, Ross},
  year = {2023},
  month = apr,
  number = {arXiv:2304.02643},
  eprint = {2304.02643},
  primaryclass = {cs},
  publisher = {arXiv},
  doi = {10.48550/arXiv.2304.02643},
  urldate = {2023-12-18},
  archiveprefix = {arXiv}
}

@article{kocDeterminationWatermelonVolume2007,
  title = {Determination of Watermelon Volume Using Ellipsoid Approximation and Image Processing},
  author = {Koc, Ali Bulent},
  year = {2007},
  month = sep,
  journal = {Postharvest Biology and Technology},
  volume = {45},
  number = {3},
  pages = {366--371},
  issn = {0925-5214},
  doi = {10.1016/j.postharvbio.2007.03.010},
  urldate = {2025-04-09}
}

@inproceedings{NIPS2012_c399862d,
  title = {{{ImageNet}} Classification with Deep Convolutional Neural Networks},
  booktitle = {Advances in Neural Information Processing Systems},
  author = {Krizhevsky, Alex and Sutskever, Ilya and Hinton, Geoffrey E},
  editor = {Pereira, F. and Burges, C.J. and Bottou, L. and Weinberger, K.Q.},
  year = 2012,
  volume = {25},
  publisher = {Curran Associates, Inc.}
}

@incollection{lerstenMorphologyAnatomyWheat1987,
  title = {Morphology and {{Anatomy}} of the {{Wheat Plant}}},
  booktitle = {Wheat and {{Wheat Improvement}}},
  author = {Lersten, Nels R.},
  year = {1987},
  pages = {33--75},
  publisher = {John Wiley \& Sons, Ltd},
  doi = {10.2134/agronmonogr13.2ed.c2},
  urldate = {2025-06-19},
  chapter = {2},
  isbn = {978-0-89118-208-5},
  langid = {english}
}

@misc{li2022comprehensivereviewdeepsupervisionlu,
      title={A Comprehensive Review on Deep Supervision: Theories and Applications}, 
      author={Renjie Li and Xinyi Wang and Guan Huang and Wenli Yang and Kaining Zhang and Xiaotong Gu and Son N. Tran and Saurabh Garg and Jane Alty and Quan Bai},
      year={2022},
      eprint={2207.02376},
      archivePrefix={arXiv},
      primaryClass={cs.CV},
      url={https://arxiv.org/abs/2207.02376}, 
}

@article{liuExtractionWheatSpike2023,
  title = {Extraction of {{Wheat Spike Phenotypes From Field-Collected Lidar Data}} and {{Exploration}} of {{Their Relationships With Wheat Yield}}},
  author = {Liu, Zhonghua and Jin, Shichao and Liu, Xiaoqiang and Yang, Qiuli and Li, Qing and Zang, Jingrong and Li, Zhaofeng and Hu, Tianyu and Guo, Zifeng and Wu, Jin and Jiang, Dong and Su, Yanjun},
  year = {2023},
  journal = {IEEE Transactions on Geoscience and Remote Sensing},
  volume = {61},
  pages = {1--13},
  issn = {1558-0644},
  doi = {10.1109/TGRS.2023.3333344},
  urldate = {2025-06-16}
}

@article{loFoodVolumeEstimation2018,
  title = {Food {{Volume Estimation Based}} on {{Deep Learning View Synthesis}} from a {{Single Depth Map}}},
  author = {Lo, Frank P.-W. and Sun, Yingnan and Qiu, Jianing and Lo, Benny},
  year = {2018},
  month = dec,
  journal = {Nutrients},
  volume = {10},
  number = {12},
  pages = {2005},
  publisher = {Multidisciplinary Digital Publishing Institute},
  issn = {2072-6643},
  doi = {10.3390/nu10122005},
  urldate = {2023-10-27},
  copyright = {http://creativecommons.org/licenses/by/3.0/},
  langid = {english}
}

@article{monVisionBasedVolume2020,
  title = {Vision Based Volume Estimation Method for Automatic Mango Grading System},
  author = {Mon, TheOo and ZarAung, Nay},
  year = {2020},
  month = oct,
  journal = {Biosystems Engineering},
  volume = {198},
  pages = {338--349},
  issn = {1537-5110},
  doi = {10.1016/j.biosystemseng.2020.08.021},
  urldate = {2025-04-09}
}

@article{moredaNondestructiveTechnologiesFruit2009,
  title = {Non-Destructive Technologies for Fruit and Vegetable Size Determination -- {{A}} Review},
  author = {Moreda, G. P. and {Ortiz-Ca{\~n}avate}, J. and {Garc{\'i}a-Ramos}, F. J. and {Ruiz-Altisent}, M.},
  year = {2009},
  month = {may},
  journal = {Journal of Food Engineering},
  volume = {92},
  number = {2},
  pages = {119--136},
  issn = {0260-8774},
  doi = {10.1016/j.jfoodeng.2008.11.004},
  urldate = {2025-04-09}
}

@misc{muntoniCnristivclabPyMeshLabPyMeshLab2023,
  title = {Cnr-Isti-Vclab/{{PyMeshLab}}: {{PyMeshLab}} V2023.12},
  shorttitle = {Cnr-Isti-Vclab/{{PyMeshLab}}},
  author = {Muntoni, Alessandro and {jmespadero} and Cignoni, Paolo and Luaces, Alberto and RichardScottOZ and {luzpaz} and Zhang, Fubin},
  year = {2023},
  month = dec,
  doi = {10.5281/zenodo.10363967},
  urldate = {2024-01-04},
  howpublished = {Zenodo}
}

@article{niuNovelMethodWheat2024,
  title = {A {{Novel Method}} for {{Wheat Spike Phenotyping Based}} on {{Instance Segmentation}} and {{Classification}}},
  author = {Niu, Ziang and Liang, Ning and He, Yiyin and Xu, Chengjia and Sun, Sashuang and Zhou, Zhenjiang and Qiu, Zhengjun},
  year = {2024},
  month = jan,
  journal = {Applied Sciences},
  volume = {14},
  number = {14},
  pages = {6031},
  publisher = {Multidisciplinary Digital Publishing Institute},
  issn = {2076-3417},
  doi = {10.3390/app14146031},
  urldate = {2025-06-19},
  copyright = {http://creativecommons.org/licenses/by/3.0/},
  langid = {english}
}

@incollection{omahonyDeepLearningVs2020,
  title = {Deep {{Learning}} vs. {{Traditional Computer Vision}}},
  booktitle = {Advances in {{Computer Vision}}},
  author = {O'Mahony, Niall and Campbell, Sean and Carvalho, Anderson and Harapanahalli, Suman and Hernandez, Gustavo Velasco and Krpalkova, Lenka and Riordan, Daniel and Walsh, Joseph},
  editor = {Arai, Kohei and Kapoor, Supriya},
  year = {2020},
  volume = {943},
  pages = {128--144},
  publisher = {Springer International Publishing},
  address = {Cham},
  doi = {10.1007/978-3-030-17795-9_10},
  urldate = {2025-04-17},
  isbn = {978-3-030-17794-2 978-3-030-17795-9},
  langid = {english}
}

@inproceedings{omahonyRealtimeMonitoringPowder2017,
  title = {Real-Time Monitoring of Powder Blend Composition Using near Infrared Spectroscopy},
  author = {O' Mahony, Niall and Murphy, Trevor and Panduru, Krishna and Riordan, Daniel and Walsh, Joseph},
  year = {2017},
  month = {dec},
  booktitle = {2017 Eleventh International Conference on Sensing Technology (ICST)},
  pages = {1--6},
  doi = {10.1109/ICSensT.2017.8304431}
}

@article{omidEstimatingVolumeMass2010,
  title = {Estimating Volume and Mass of Citrus Fruits by Image Processing Technique},
  author = {Omid, M. and Khojastehnazhand, M. and Tabatabaeefar, A.},
  year = {2010},
  month = sep,
  journal = {Journal of Food Engineering},
  volume = {100},
  number = {2},
  pages = {315--321},
  issn = {0260-8774},
  doi = {10.1016/j.jfoodeng.2010.04.015},
  urldate = {2025-03-21}
}

@misc{oquabDINOv2LearningRobust2024,
  title = {{{DINOv2}}: {{Learning Robust Visual Features}} without {{Supervision}}},
  shorttitle = {{{DINOv2}}},
  author = {Oquab, Maxime and Darcet, Timoth{\'e}e and Moutakanni, Th{\'e}o and Vo, Huy and Szafraniec, Marc and Khalidov, Vasil and Fernandez, Pierre and Haziza, Daniel and Massa, Francisco and {El-Nouby}, Alaaeldin and Assran, Mahmoud and Ballas, Nicolas and Galuba, Wojciech and Howes, Russell and Huang, Po-Yao and Li, Shang-Wen and Misra, Ishan and Rabbat, Michael and Sharma, Vasu and Synnaeve, Gabriel and Xu, Hu and Jegou, Herv{\'e} and Mairal, Julien and Labatut, Patrick and Joulin, Armand and Bojanowski, Piotr},
  year = {2024},
  month = {feb},
  number = {arXiv:2304.07193},
  eprint = {2304.07193},
  primaryclass = {cs},
  publisher = {arXiv},
  doi = {10.48550/arXiv.2304.07193},
  urldate = {2024-12-19},
  archiveprefix = {arXiv}
}

@article{pretiniPhysiologyGeneticsFruiting2021,
  title = {The Physiology and Genetics behind Fruiting Efficiency: A Promising Spike Trait to Improve Wheat Yield Potential},
  shorttitle = {The Physiology and Genetics behind Fruiting Efficiency},
  author = {Pretini, Nicole and Alonso, Mar{\'i}a P and Vanzetti, Leonardo S and Pontaroli, Ana C and Gonz{\'a}lez, Fernanda G},
  year = {2021},
  month = may,
  journal = {Journal of Experimental Botany},
  volume = {72},
  number = {11},
  pages = {3987--4004},
  issn = {0022-0957},
  doi = {10.1093/jxb/erab080},
  urldate = {2025-06-20}
}

@article{rachakondaSourcesErrorsStructured2019,
  title = {Sources of {{Errors}} in {{Structured Light 3D Scanners}}},
  author = {Rachakonda, Prem K. and Muralikrishnan, Bala and Sawyer, Daniel S.},
  year = {2019},
  month = apr,
  journal = {NIST},
  publisher = {Prem K. Rachakonda, Bala Muralikrishnan, Daniel S. Sawyer},
  urldate = {2025-04-09},
  langid = {english}
}

@inproceedings{renBalancedMSEImbalanced2022,
  title = {Balanced {{MSE}} for {{Imbalanced Visual Regression}}},
  booktitle = {2022 {{IEEE}}/{{CVF Conference}} on {{Computer Vision}} and {{Pattern Recognition}} ({{CVPR}})},
  author = {Ren, Jiawei and Zhang, Mingyuan and Yu, Cunjun and Liu, Ziwei},
  year = {2022},
  month = jun,
  pages = {7916--7925},
  publisher = {IEEE},
  address = {New Orleans, LA, USA},
  doi = {10.1109/CVPR52688.2022.00777},
  urldate = {2025-06-17},
  copyright = {https://doi.org/10.15223/policy-029},
  isbn = {978-1-6654-6946-3},
  langid = {english}
}

@article{rosellpoloTractormountedScanningLIDAR2009,
  title = {A Tractor-Mounted Scanning {{LIDAR}} for the Non-Destructive Measurement of Vegetative Volume and Surface Area of Tree-Row Plantations: {{A}} Comparison with Conventional Destructive Measurements},
  shorttitle = {A Tractor-Mounted Scanning {{LIDAR}} for the Non-Destructive Measurement of Vegetative Volume and Surface Area of Tree-Row Plantations},
  author = {Rosell Polo, Joan Ramon and Sanz, Ricardo and Llorens, Jordi and Arn{\'o}, Jaume and Escol{\`a}, Alexandre and {Ribes-Dasi}, Manel and Masip, Joan and Camp, Ferran and Gr{\`a}cia, Felip and Solanelles, Francesc and Pallej{\`a}, Tom{\`a}s and Val, Luis and Planas, Santiago and Gil, Emilio and Palac{\'i}n, Jordi},
  year = {2009},
  month = {feb},
  journal = {Biosystems Engineering},
  volume = {102},
  number = {2},
  pages = {128--134},
  issn = {1537-5110},
  doi = {10.1016/j.biosystemseng.2008.10.009},
  urldate = {2025-06-16}
}

@article{slaferFruitingEfficiencyAlternative2015,
  title = {Fruiting Efficiency: An Alternative Trait to Further Rise Wheat Yield},
  shorttitle = {Fruiting Efficiency},
  author = {Slafer, Gustavo and El{\'i}a, M{\'o}nica and Savin, Roxana and Garc{\'i}a, Guillermo and Terrile, Ignacio and Ferrante, Ariel and Miralles, Daniel and Gonz{\'a}lez, Fernanda},
  year = {2015},
  month = may,
  journal = {Food and Energy Security},
  volume = {4},
  doi = {10.1002/fes3.59}
}

@article{soltaniEggVolumePrediction2015,
  title = {Egg Volume Prediction Using Machine Vision Technique Based on Pappus Theorem and Artificial Neural Network},
  author = {Soltani, Mahmoud and Omid, Mahmoud and Alimardani, Reza},
  year = {2015},
  month = may,
  journal = {Journal of Food Science and Technology},
  volume = {52},
  number = {5},
  pages = {3065--3071},
  issn = {0975-8402},
  doi = {10.1007/s13197-014-1350-6},
  urldate = {2025-04-08},
  langid = {english}
}

@article{steinbrenerLearningMetricVolume2023b,
  title = {Learning Metric Volume Estimation of Fruits and Vegetables from Short Monocular Video Sequences},
  author = {Steinbrener, Jan and Dimitrievska, Vesna and Pittino, Federico and Starmans, Frans and Waldner, Roland and Holzbauer, J{\"u}rgen and Arnold, Thomas},
  year = {2023},
  month = apr,
  journal = {Heliyon},
  volume = {9},
  number = {4},
  pages = {e14722},
  issn = {2405-8440},
  doi = {10.1016/j.heliyon.2023.e14722},
  urldate = {2025-04-23}
}

@article{suPotatoFeaturePrediction2017,
  title = {Potato Feature Prediction Based on Machine Vision and {{3D}} Model Rebuilding},
  author = {Su, Qinghua and Kondo, Naoshi and Li, Minzan and Sun, Hong and Al Riza, Dimas Firmanda},
  year = {2017},
  month = may,
  journal = {Computers and Electronics in Agriculture},
  volume = {137},
  pages = {41--51},
  issn = {0168-1699},
  doi = {10.1016/j.compag.2017.03.020},
  urldate = {2025-04-09}
}

@inproceedings{uluisikImageProcessingBased2018,
  title = {Image Processing Based Machine Vision System for Tomato Volume Estimation},
  booktitle = {2018 {{Electric Electronics}}, {{Computer Science}}, {{Biomedical Engineerings}}' {{Meeting}} ({{EBBT}})},
  author = {Ului{\c S}ik, Selman and Yildiz, Fikret and {\"O}zdem{\.I}r, Ahmet Turan},
  year = {2018},
  month = apr,
  pages = {1--4},
  doi = {10.1109/EBBT.2018.8391460},
  urldate = {2025-03-21}
}

@article{vivekvenkateshEstimationVolumeMass2015,
  title = {Estimation of {{Volume}} and {{Mass}} of {{Axi-Symmetric Fruits Using Image Processing Technique}}},
  author = {Vivek Venkatesh, G. and , S. Md., Iqbal and , A., Gopal and {and Ganesan}, D.},
  year = {2015},
  month = mar,
  journal = {International Journal of Food Properties},
  volume = {18},
  number = {3},
  pages = {608--626},
  publisher = {Taylor \& Francis},
  issn = {1094-2912},
  doi = {10.1080/10942912.2013.831444},
  urldate = {2025-04-08}
}

@article{wangUnsupervisedAutomaticMeasurement2022,
  title = {An Unsupervised Automatic Measurement of Wheat Spike Dimensions in Dense {{3D}} Point Clouds for Field Application},
  author = {Wang, Fuli and Li, Fengping and Mohan, Vishwanathan and Dudley, Richard and Gu, Dongbing and Bryant, Ruth},
  year = {2022},
  month = nov,
  journal = {Biosystems Engineering},
  series = {New Advances in Measurement and Data Processing Techniques for {{Agriculture}}, {{Food}} and {{Environment}}.},
  volume = {223},
  pages = {103--114},
  issn = {1537-5110},
  doi = {10.1016/j.biosystemseng.2021.11.022},
  urldate = {2025-06-16}
}

@inproceedings{xuImagebasedFoodVolume2013,
  title = {Image-Based Food Volume Estimation},
  booktitle = {Proceedings of the 5th International Workshop on {{Multimedia}} for Cooking \& Eating Activities},
  author = {Xu, Chang and He, Ye and Khannan, Nitin and Parra, Albert and Boushey, Carol and Delp, Edward},
  year = {2013},
  month = oct,
  series = {{{CEA}} '13},
  pages = {75--80},
  publisher = {Association for Computing Machinery},
  address = {New York, NY, USA},
  doi = {10.1145/2506023.2506037},
  urldate = {2024-01-04},
  isbn = {978-1-4503-2392-5}
}

@misc{yuFreeFormImageInpainting2019,
  title = {Free-{{Form Image Inpainting}} with {{Gated Convolution}}},
  author = {Yu, Jiahui and Lin, Zhe and Yang, Jimei and Shen, Xiaohui and Lu, Xin and Huang, Thomas},
  year = {2019},
  month = oct,
  number = {arXiv:1806.03589},
  eprint = {1806.03589},
  primaryclass = {cs},
  publisher = {arXiv},
  doi = {10.48550/arXiv.1806.03589},
  urldate = {2023-12-18},
  archiveprefix = {arXiv}
}

@article{zhangFastParallelAlgorithm1984,
  title = {A Fast Parallel Algorithm for Thinning Digital Patterns},
  author = {Zhang, T. Y. and Suen, C. Y.},
  year = {1984},
  month = mar,
  journal = {Communications of the ACM},
  volume = {27},
  number = {3},
  pages = {236--239},
  issn = {0001-0782},
  doi = {10.1145/357994.358023},
  urldate = {2023-12-22}
}

@InProceedings{zhangWheat3DGSInfield3D2025,
    author    = {Zhang, Daiwei and Gajardo, Joaquin and Medic, Tomislav and Katircioglu, Isinsu and Boss, Mike and Kirchgessner, Norbert and Walter, Achim and Roth, Lukas},
    title     = {Wheat3DGS: In-field 3D Reconstruction, Instance Segmentation and Phenotyping of Wheat Heads with Gaussian Splatting},
    booktitle = {Proceedings of the Computer Vision and Pattern Recognition Conference (CVPR) Workshops},
    month     = {June},
    year      = {2025},
    pages     = {5360-5370}
}

@misc{zhouOpen3DModernLibrary2018,
  title = {{{Open3D}}: {{A Modern Library}} for {{3D Data Processing}}},
  shorttitle = {{{Open3D}}},
  author = {Zhou, Qian-Yi and Park, Jaesik and Koltun, Vladlen},
  year = {2018},
  month = jan,
  number = {arXiv:1801.09847},
  eprint = {1801.09847},
  primaryclass = {cs},
  publisher = {arXiv},
  doi = {10.48550/arXiv.1801.09847},
  urldate = {2024-01-04},
  archiveprefix = {arXiv}
}

@misc{simeoniDINOv32025,
  title = {{{DINOv3}}},
  author = {Sim{\'e}oni, Oriane and Vo, Huy V. and Seitzer, Maximilian and Baldassarre, Federico and Oquab, Maxime and Jose, Cijo and Khalidov, Vasil and Szafraniec, Marc and Yi, Seungeun and Ramamonjisoa, Micha{\"e}l and Massa, Francisco and Haziza, Daniel and Wehrstedt, Luca and Wang, Jianyuan and Darcet, Timoth{\'e}e and Moutakanni, Th{\'e}o and Sentana, Leonel and Roberts, Claire and Vedaldi, Andrea and Tolan, Jamie and Brandt, John and Couprie, Camille and Mairal, Julien and J{\'e}gou, Herv{\'e} and Labatut, Patrick and Bojanowski, Piotr},
  year = 2025,
  month = aug,
  number = {arXiv:2508.10104},
  eprint = {2508.10104},
  primaryclass = {cs},
  publisher = {arXiv},
  doi = {10.48550/arXiv.2508.10104},
  urldate = {2025-12-10},
  archiveprefix = {arXiv},
  langid = {english}
}

@article{orosooTransformingEnglishLanguage2025,
  title = {Transforming {{English}} Language Learning: {{Advanced}} Speech Recognition with {{MLP-LSTM}} for Personalized Education},
  shorttitle = {Transforming {{English}} Language Learning},
  author = {Orosoo, Myagmarsuren and Raash, Namjildagva and Treve, Mark and M. Lahza, Hassan Fareed and Alshammry, Nizal and Ramesh, Janjhyam Venkata Naga and Rengarajan, Manikandan},
  year = 2025,
  month = jan,
  journal = {Alexandria Engineering Journal},
  volume = {111},
  pages = {21--32},
  issn = {1110-0168},
  doi = {10.1016/j.aej.2024.10.065},
  urldate = {2025-12-10}
}

@misc{sutskeverSequenceSequenceLearning2014,
  title = {Sequence to {{Sequence Learning}} with {{Neural Networks}}},
  author = {Sutskever, Ilya and Vinyals, Oriol and Le, Quoc V.},
  year = 2014,
  month = dec,
  number = {arXiv:1409.3215},
  eprint = {1409.3215},
  primaryclass = {cs},
  publisher = {arXiv},
  doi = {10.48550/arXiv.1409.3215},
  urldate = {2025-12-10},
  archiveprefix = {arXiv},
  langid = {english}
}

@article{salehinRealTimeMedicalImage2023,
  title = {Real-{{Time Medical Image Classification}} with {{ML Framework}} and {{Dedicated CNN}}--{{LSTM Architecture}}},
  author = {Salehin, Imrus and Islam, Md. Shamiul and Amin, Nazrul and Baten, Md. Abu and Noman, S. M. and Saifuzzaman, Mohd and Yazmyradov, Serdar},
  year = 2023,
  journal = {Journal of Sensors},
  volume = {2023},
  number = {1},
  pages = {3717035},
  issn = {1687-7268},
  doi = {10.1155/2023/3717035},
  urldate = {2025-12-10},
  langid = {english}
}

@misc{saikiaHybridCNNLSTMModel2022,
  title = {A {{Hybrid CNN-LSTM}} Model for {{Video Deepfake Detection}} by {{Leveraging Optical Flow Features}}},
  author = {Saikia, Pallabi and Dholaria, Dhwani and Yadav, Priyanka and Patel, Vaidehi and Roy, Mohendra},
  year = 2022,
  month = jul,
  number = {arXiv:2208.00788},
  eprint = {2208.00788},
  primaryclass = {cs},
  publisher = {arXiv},
  doi = {10.48550/arXiv.2208.00788},
  urldate = {2025-12-10},
  archiveprefix = {arXiv},
  langid = {english}
}

@misc{vaswaniAttentionAllYou2023,
  title = {Attention {{Is All You Need}}},
  author = {Vaswani, Ashish and Shazeer, Noam and Parmar, Niki and Uszkoreit, Jakob and Jones, Llion and Gomez, Aidan N. and Kaiser, Lukasz and Polosukhin, Illia},
  year = {2023},
  month = {aug},
  number = {arXiv:1706.03762},
  eprint = {1706.03762},
  primaryclass = {cs},
  publisher = {arXiv},
  urldate = {2024-01-27},
  archiveprefix = {arXiv},
  langid = {english}
}

@article{hochreiterLongShortTermMemory1997,
  title = {Long {{Short-Term Memory}}},
  author = {Hochreiter, Sepp and Schmidhuber, J{\"u}rgen},
  year = 1997,
  month = nov,
  journal = {Neural Computation},
  volume = {9},
  number = {8},
  pages = {1735--1780},
  issn = {0899-7667},
  doi = {10.1162/neco.1997.9.8.1735},
  urldate = {2025-12-10}
}

@misc{djoumessiHybridFullyConvolutional2025,
  title = {A {{Hybrid Fully Convolutional CNN-Transformer Model}} for {{Inherently Interpretable Disease Detection}} from {{Retinal Fundus Images}}},
  author = {Djoumessi, Kerol and Mensah, Samuel Ofosu and Berens, Philipp},
  year = 2025,
  month = apr,
  journal = {arXiv.org},
  urldate = {2025-12-10},
  howpublished = {https://arxiv.org/abs/2504.08481v4},
  langid = {english}
}

@misc{heDeepResidualLearning2015,
  title = {Deep {{Residual Learning}} for {{Image Recognition}}},
  author = {He, Kaiming and Zhang, Xiangyu and Ren, Shaoqing and Sun, Jian},
  year = 2015,
  month = dec,
  number = {arXiv:1512.03385},
  eprint = {1512.03385},
  primaryclass = {cs},
  publisher = {arXiv},
  doi = {10.48550/arXiv.1512.03385},
  urldate = {2024-01-04},
  archiveprefix = {arXiv}
}

@misc{ranftlVisionTransformersDense2021,
  title = {Vision {{Transformers}} for {{Dense Prediction}}},
  author = {Ranftl, Ren{\'e} and Bochkovskiy, Alexey and Koltun, Vladlen},
  year = 2021,
  month = mar,
  number = {arXiv:2103.13413},
  eprint = {2103.13413},
  primaryclass = {cs},
  publisher = {arXiv},
  doi = {10.48550/arXiv.2103.13413},
  urldate = {2025-12-11},
  archiveprefix = {arXiv},
  langid = {english}
}

@inproceedings{dongSpeechTransformerNoRecurrenceSequencetoSequence2018,
  title = {Speech-{{Transformer}}: {{A No-Recurrence Sequence-to-Sequence Model}} for {{Speech Recognition}}},
  shorttitle = {Speech-{{Transformer}}},
  booktitle = {2018 {{IEEE International Conference}} on {{Acoustics}}, {{Speech}} and {{Signal Processing}} ({{ICASSP}})},
  author = {Dong, Linhao and Xu, Shuang and Xu, Bo},
  year = {2018},
  month = {apr},
  pages = {5884--5888},
  issn = {2379-190X},
  doi = {10.1109/ICASSP.2018.8462506},
  urldate = {2025-12-13}
}

@misc{wangLearningDeepTransformer2019,
  title = {Learning {{Deep Transformer Models}} for {{Machine Translation}}},
  author = {Wang, Qiang and Li, Bei and Xiao, Tong and Zhu, Jingbo and Li, Changliang and Wong, Derek F. and Chao, Lidia S.},
  year = {2019},
  month = {jun},
  number = {arXiv:1906.01787},
  eprint = {1906.01787},
  primaryclass = {cs},
  publisher = {arXiv},
  doi = {10.48550/arXiv.1906.01787},
  urldate = {2025-12-13},
  archiveprefix = {arXiv},
  langid = {english}
}

@misc{dosovitskiyImageWorth16x162021,
  title = {An {{Image}} Is {{Worth}} 16x16 {{Words}}: {{Transformers}} for {{Image Recognition}} at {{Scale}}},
  shorttitle = {An {{Image}} Is {{Worth}} 16x16 {{Words}}},
  author = {Dosovitskiy, Alexey and Beyer, Lucas and Kolesnikov, Alexander and Weissenborn, Dirk and Zhai, Xiaohua and Unterthiner, Thomas and Dehghani, Mostafa and Minderer, Matthias and Heigold, Georg and Gelly, Sylvain and Uszkoreit, Jakob and Houlsby, Neil},
  year = 2021,
  month = jun,
  number = {arXiv:2010.11929},
  eprint = {2010.11929},
  primaryclass = {cs},
  publisher = {arXiv},
  doi = {10.48550/arXiv.2010.11929},
  urldate = {2025-12-14},
  archiveprefix = {arXiv},
  langid = {english}
}

@misc{carionEndtoEndObjectDetection2020a,
  title = {End-to-{{End Object Detection}} with {{Transformers}}},
  author = {Carion, Nicolas and Massa, Francisco and Synnaeve, Gabriel and Usunier, Nicolas and Kirillov, Alexander and Zagoruyko, Sergey},
  year = {2020},
  month = {may},
  number = {arXiv:2005.12872},
  eprint = {2005.12872},
  primaryclass = {cs},
  publisher = {arXiv},
  doi = {10.48550/arXiv.2005.12872},
  urldate = {2025-12-14},
  archiveprefix = {arXiv}
}

@inproceedings{liDeepVolDeepFruit2018,
  title = {{{DeepVol}}: {{Deep Fruit Volume Estimation}}},
  shorttitle = {{{DeepVol}}},
  booktitle = {Artificial {{Neural Networks}} and {{Machine Learning}} -- {{ICANN}} 2018},
  author = {Li, Hongyu and Han, Tianqi},
  editor = {K{\r u}rkov{\'a}, V{\v e}ra and Manolopoulos, Yannis and Hammer, Barbara and Iliadis, Lazaros and Maglogiannis, Ilias},
  year = 2018,
  pages = {331--341},
  publisher = {Springer International Publishing},
  address = {Cham},
  doi = {10.1007/978-3-030-01424-7_33},
  isbn = {978-3-030-01424-7},
  langid = {english}
}

@article{rifeFieldBookOpenSource2014,
  title = {Field {{Book}}: {{An Open-Source Application}} for {{Field Data Collection}} on {{Android}}},
  shorttitle = {Field {{Book}}},
  author = {Rife, Trevor W. and Poland, Jesse A.},
  year = 2014,
  journal = {Crop Science},
  volume = {54},
  number = {4},
  pages = {1624--1627},
  issn = {1435-0653},
  doi = {10.2135/cropsci2013.08.0579},
  urldate = {2025-12-19},
  copyright = {Copyright \copyright{} 2014 by the Crop Science Society of America, Inc.},
  langid = {english}
}

@misc{hanFoMo4WheatReliableCrop2025,
  title = {{{FoMo4Wheat}}: {{Toward}} Reliable Crop Vision Foundation Models with Globally Curated Data},
  shorttitle = {{{FoMo4Wheat}}},
  author = {Han, Bing and Zhu, Chen and Han, Dong and Yu, Rui and Cao, Songliang and Wu, Jianhui and Chapman, Scott and Wang, Zijian and Zheng, Bangyou and Guo, Wei and Weiss, Marie and de Solan, Benoit and Hund, Andreas and Roth, Lukas and Norbert, Kirchgessner and Visioni, Andrea and Ge, Yufeng and Li, Wenjuan and Comar, Alexis and Jiang, Dong and Han, Dejun and Baret, Fred and Ding, Yanfeng and Lu, Hao and Liu, Shouyang},
  year = 2025,
  month = sep,
  number = {arXiv:2509.06907},
  eprint = {2509.06907},
  primaryclass = {cs},
  publisher = {arXiv},
  doi = {10.48550/arXiv.2509.06907},
  urldate = {2025-12-26},
  archiveprefix = {arXiv}
}

@misc{INnovationsPlantVarIety,
    author = {{INVITE Consortium}},
    title = {{{INnovations}} in Plant {{VarIety Testing}} in {{Europe}} to Foster the Introduction of New Varieties Better Adapted to Varying Biotic and Abiotic Conditions and to More Sustainable Crop Management Practices \textbar{} {{INVITE}} \textbar{} {{Project}} \textbar{} {{Fact Sheet}} \textbar{} {{H2020}}},
    year   = {n.d.},
  journal = {CORDIS \textbar{} European Commission},
  urldate = {2025-12-19},
  howpublished = {https://cordis.europa.eu/project/id/817970},
  langid = {english}
}

@misc{ToolsMethodsExtended,
  
  author = {{PHENET Consortium}},
    title = {Tools and Methods for Extended Plant {{PHENotyping}} and {{EnviroTyping}} Services of {{European Research Infrastructures}} \textbar{} {{PHENET}} \textbar{} {{Project}} \textbar{} {{Fact Sheet}} \textbar{} {{HORIZON}}},
    year   = {n.d.},
  journal = {CORDIS \textbar{} European Commission},
  urldate = {2025-12-19},
  howpublished = {https://cordis.europa.eu/project/id/101094587},
  langid = {english}
}

\appendix
\section{Additional material}

\setcounter{table}{0}
\setcounter{algorithm}{0}
\setcounter{figure}{0}

\renewcommand{\thetable}{\Alph{section}.\arabic{table}}
\renewcommand{\thealgorithm}{\Alph{section}.\arabic{algorithm}}

    \begin{algorithm}[htbp]
       \caption{Algorithm to define the main axis of spike based on the skeleton.}
       \label{alg:main_axis}
        \begin{algorithmic}
           \STATE {\bfseries Input:} skeleton pixel mask $S$, starting point $p$
           \STATE $path \gets [p]$
           \STATE Remove $p$ from $S$
           \STATE Remove $N(p)$ from $S$
           \STATE $subpaths \gets []$
           \FOR{$\widetilde{p} \in N(p)$}
               \STATE $tmp \gets$ {\bfseries Find\_Main\_Axis}($S$, $\widetilde{p}$)
               \STATE Append $tmp$ to $subpaths$
           \ENDFOR
           \STATE $maxpath \gets \arg\max_{z \in subpaths} \text{length}(z)$
           \STATE $path \gets \text{concatenate}(path, maxpath)$
           \STATE \textbf{return} $path$
        \end{algorithmic}
    \end{algorithm}

    In Algorithm~\ref{alg:main_axis}, $N(p)$ denotes the set of 8-connected neighbors of point $p$ within $S$. Specifically, for a point $p = (i, k)$, we define $N(p) = \{\widetilde{p} \in S : \widetilde{p} \in (i \pm 1, k \pm 1)\}$.

\begin{table*}[htbp]
    \centering
    \caption{Train, validation, and test performance of the best-performing MLP, LSTM, and Transformer models evaluated on six images.}
    \label{tab:metrics_mlp_lstm_transformer}
    \begin{tabularx}{\textwidth}{
    @{}
    >{\raggedright\arraybackslash}p{0.18\textwidth}
    *{3}{>{\centering\arraybackslash}X}
    @{\hspace{10pt}}
    *{3}{>{\centering\arraybackslash}X}
    @{\hspace{10pt}}
    *{3}{>{\centering\arraybackslash}X}
    @{}
}

    \toprule
    & \multicolumn{3}{c}{\textbf{MLP}} 
    & \multicolumn{3}{c}{\textbf{LSTM}} 
    & \multicolumn{3}{c}{\textbf{Transformer}} \\
    \cmidrule(r){2-4} \cmidrule(lr){5-7} \cmidrule(l){8-10}
    \textbf{Metric} 
        & Train & Val & Test
        & Train & Val & Test
        & Train & Val & Test \\
    \midrule

    \textbf{Correlation} 
        & 0.97 & 0.94 & 0.94
        & 0.98 & 0.95 & 0.95
        & 0.92 & 0.93 & 0.91 \\

    \textbf{$R^2$} 
        & 0.94 & 0.88 & 0.88
        & 0.95 & 0.90 & 0.89
        & 0.84 & 0.85 & 0.80 \\

    \textbf{MAPE [$\%$]} 
        & 5.10 & 7.60 & 6.69
        & 4.62 & 6.81 & 6.31
        & 8.43 & 8.34 & 8.71 \\

    \textbf{MAE [$\mathrm{mm}^3$]} 
        & 237.22 & 358.57 & 301.57
        & 212.22 & 333.63 & 289.70
        & 390.35 & 410.27 & 397.91 \\

    \bottomrule
    \end{tabularx}
\end{table*}

    \begin{table*}[htbp]
        \centering
        \caption{Baseline models and fitted curve estimates based on the training set for each respective number of images.}
        \label{tab:baseline_fitted}
        \begin{tabularx}{\textwidth}{@{}lX@{}}
        \toprule
        \textbf{Baseline Model} & \textbf{Fitted Curve} \\
        \midrule
        Area baseline (one image)      & $f(x) = -1566.5864903085 + 0.0234685774\,x - 1.21 \times 10^{-8}\, x^{2}$ \\
        Area baseline (two images)     & $f(x) = -1741.5529764335 + 0.0222985968\,x - 8.9 \times 10^{-9}\, x^{2}$ \\
        Area baseline (four images)     & $f(x) = -1767.2998506379 + 0.0223104925\,x - 8.7 \times 10^{-9}\, x^{2}$ \\
        Area baseline (six images)     & $f(x) = -1541.7191738870 + 0.0206986053\,x - 6.3 \times 10^{-9}\, x^{2}$ \\

        Geometric baseline (one images)        & $f(x) = -240.723620 + 1.199311\,x - 5.3 \times 10^{-5}\, x^{2}$ \\
        Geometric baseline (two images)       & $f(x) = -359.778449 + 1.077189\,x - 3.9 \times 10^{-5}\, x^{2}$ \\
        Geometric baseline (four images)       &  $f(x) = -505.270175 + 1.116801\,x - 4.2 \times 10^{-5}\, x^{2}$ \\
        Geometric baseline (six images)        & $f(x) = -369.823033 + 1.060307\,x - 3.7 \times 10^{-5}\, x^{2}$ \\
        \bottomrule
        \end{tabularx}
    \end{table*}

    \begin{figure*}
        \centering
        \begin{subfigure}[t]{0.23\textwidth}
            \centering
            \includegraphics[width=1\linewidth]{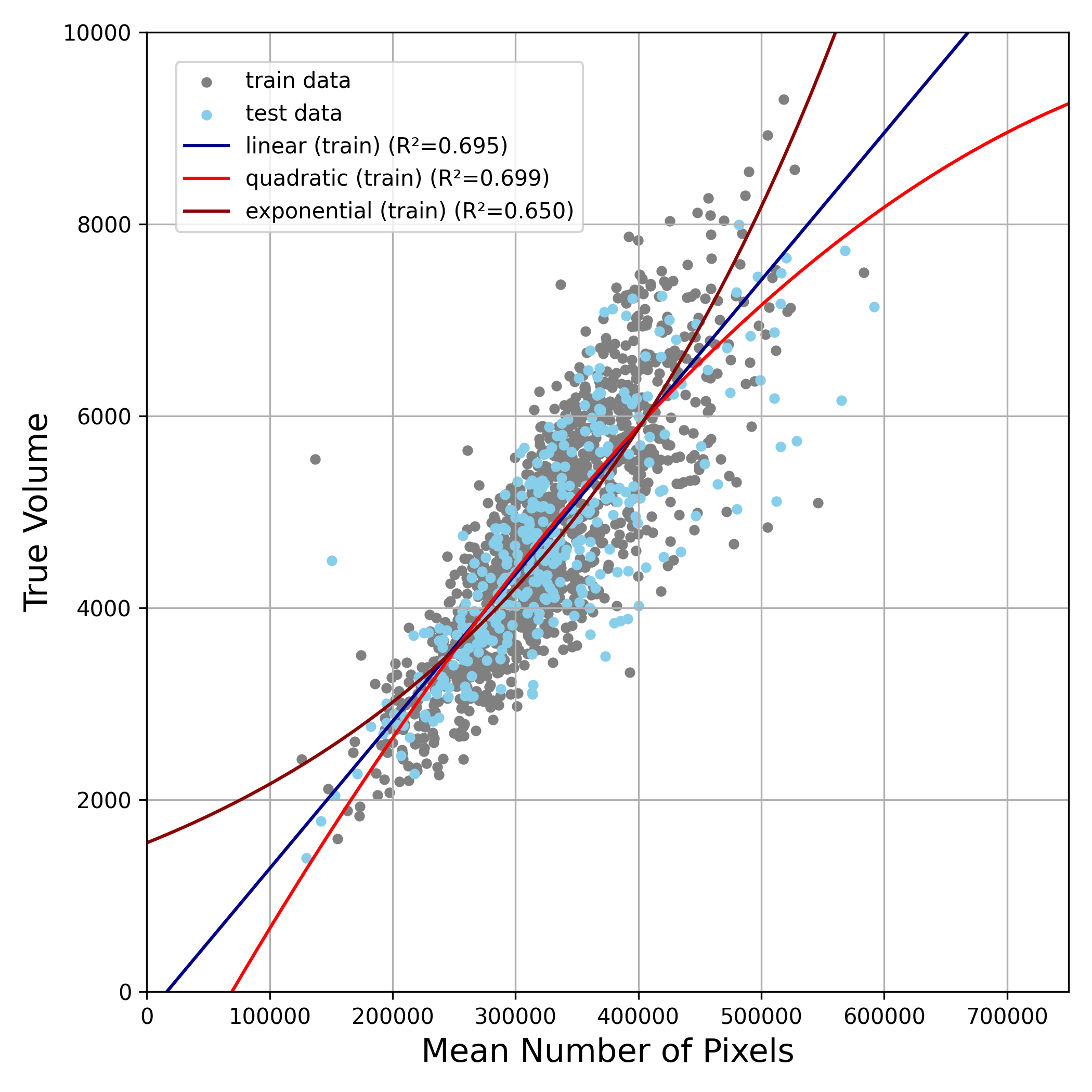}
            \caption{}
        \end{subfigure}
        \vspace{0.02\textwidth}
        \begin{subfigure}[t]{0.23\textwidth}
            \centering
            \includegraphics[width=1\linewidth]{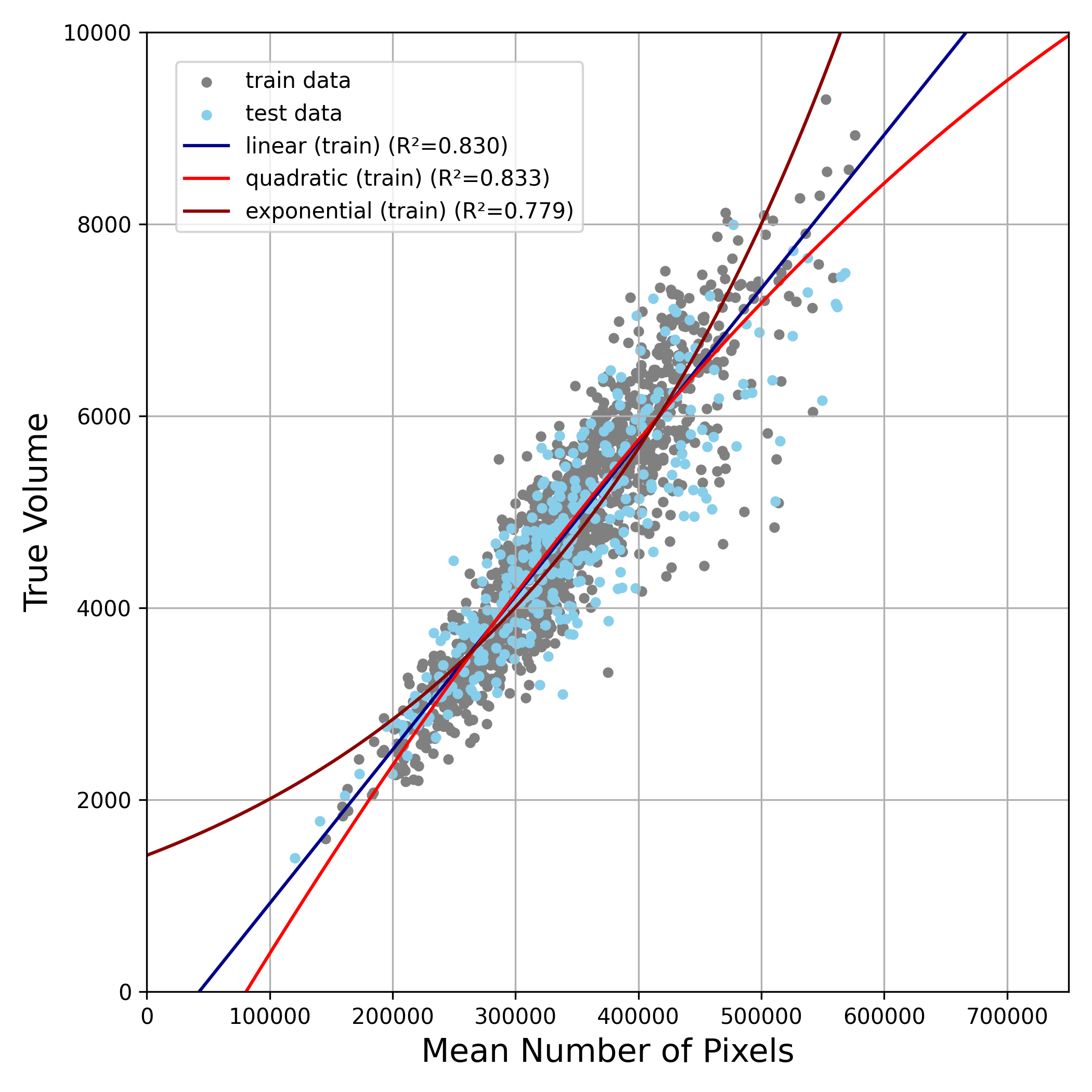}
            \caption{}
        \end{subfigure}
        \vspace{0.02\textwidth}
        \begin{subfigure}[t]{0.23\textwidth}
            \centering
            \includegraphics[width=1\linewidth]{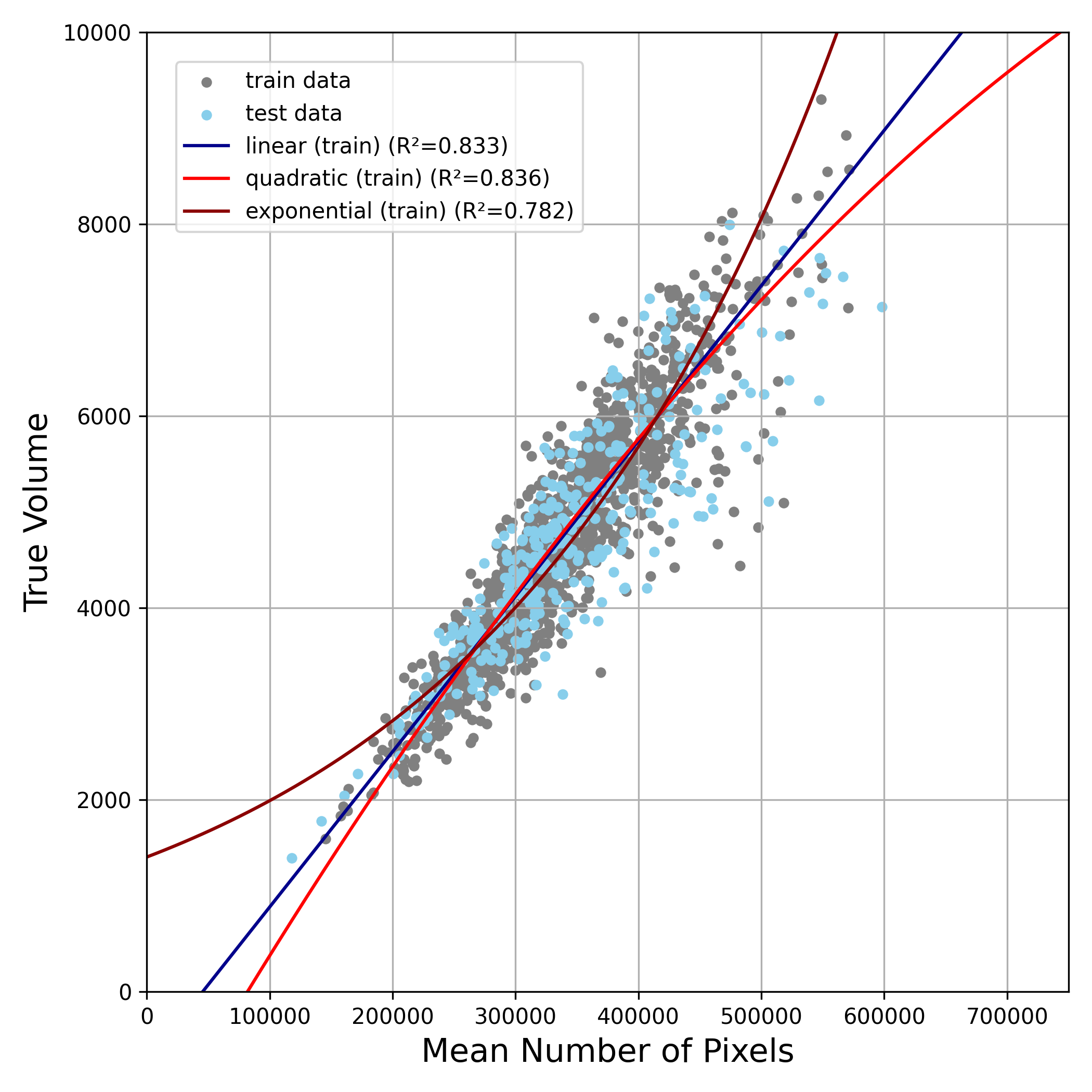}
            \caption{}
        \end{subfigure}
        \vspace{0.02\textwidth}
        \begin{subfigure}[t]{0.23\textwidth}
            \centering
            \includegraphics[width=1\linewidth]{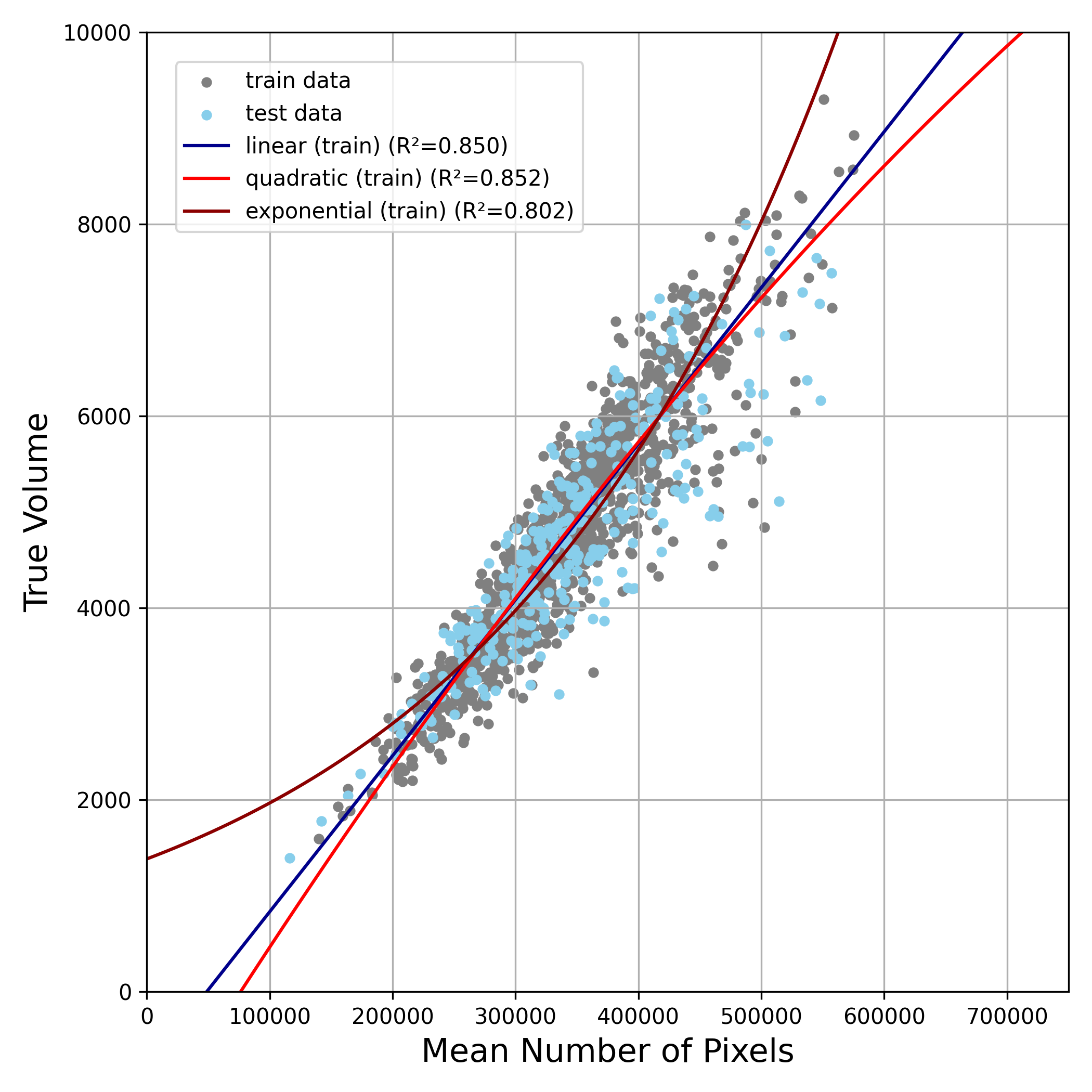}
            \caption{}
        \end{subfigure}
        \caption{Linear, quadratic, and exponential curve fitted to the number of spike pixels of the area baseline and the measured volume of the training dataset (grey), and their respective $R^2$ score. Test dataset (blue) was added for visualization purpose. Separate curves were fitted for estimations based on one image (a), two images (b), four images(c), and six images (d).}
        \label{fig:train_curves_area}
    \end{figure*}

    \begin{figure*}
        \centering
        \begin{subfigure}[t]{0.23\textwidth}
            \centering
            \includegraphics[width=1\linewidth]{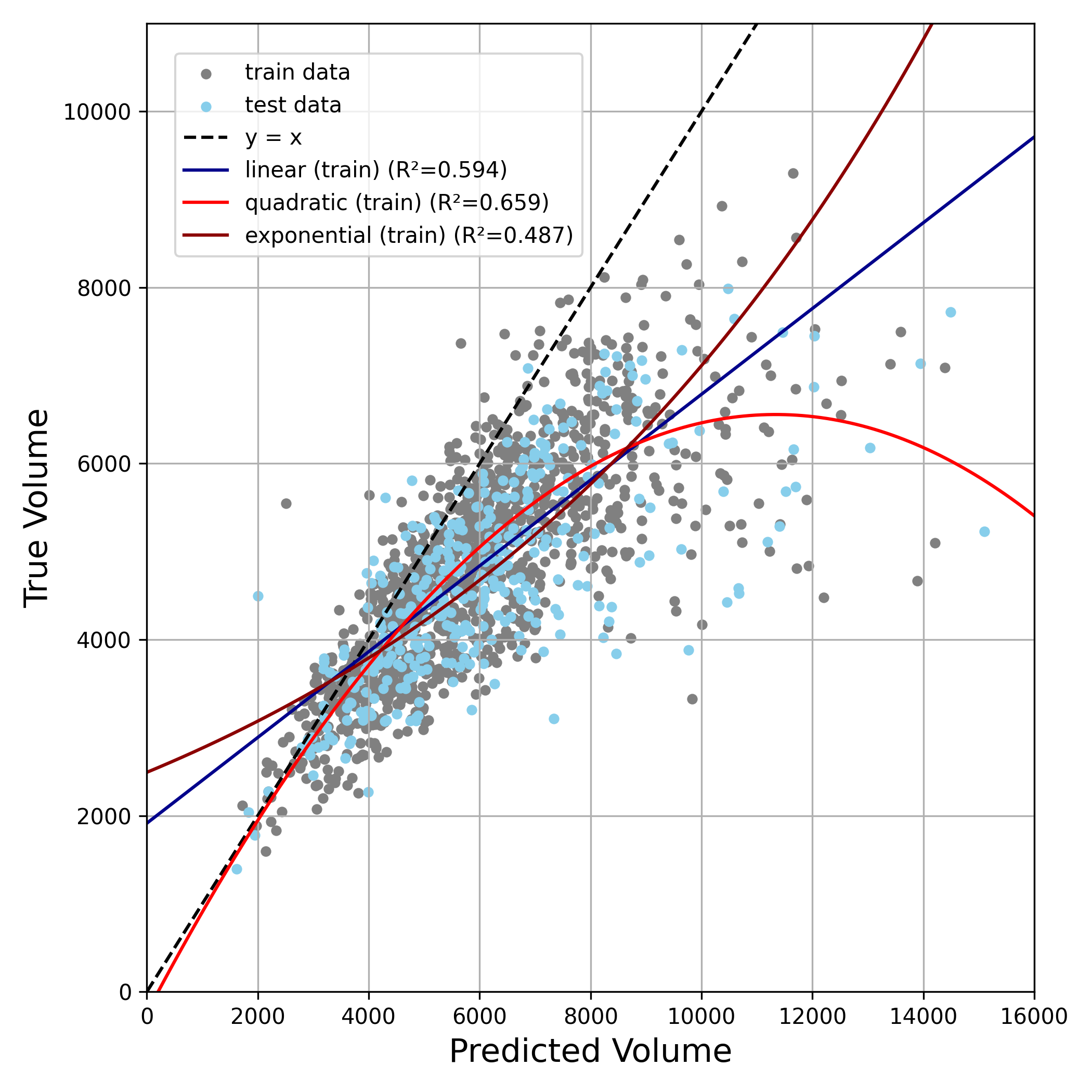}
            \caption{}
        \end{subfigure}
        \vspace{0.02\textwidth}
        \begin{subfigure}[t]{0.23\textwidth}
            \centering
            \includegraphics[width=1\linewidth]{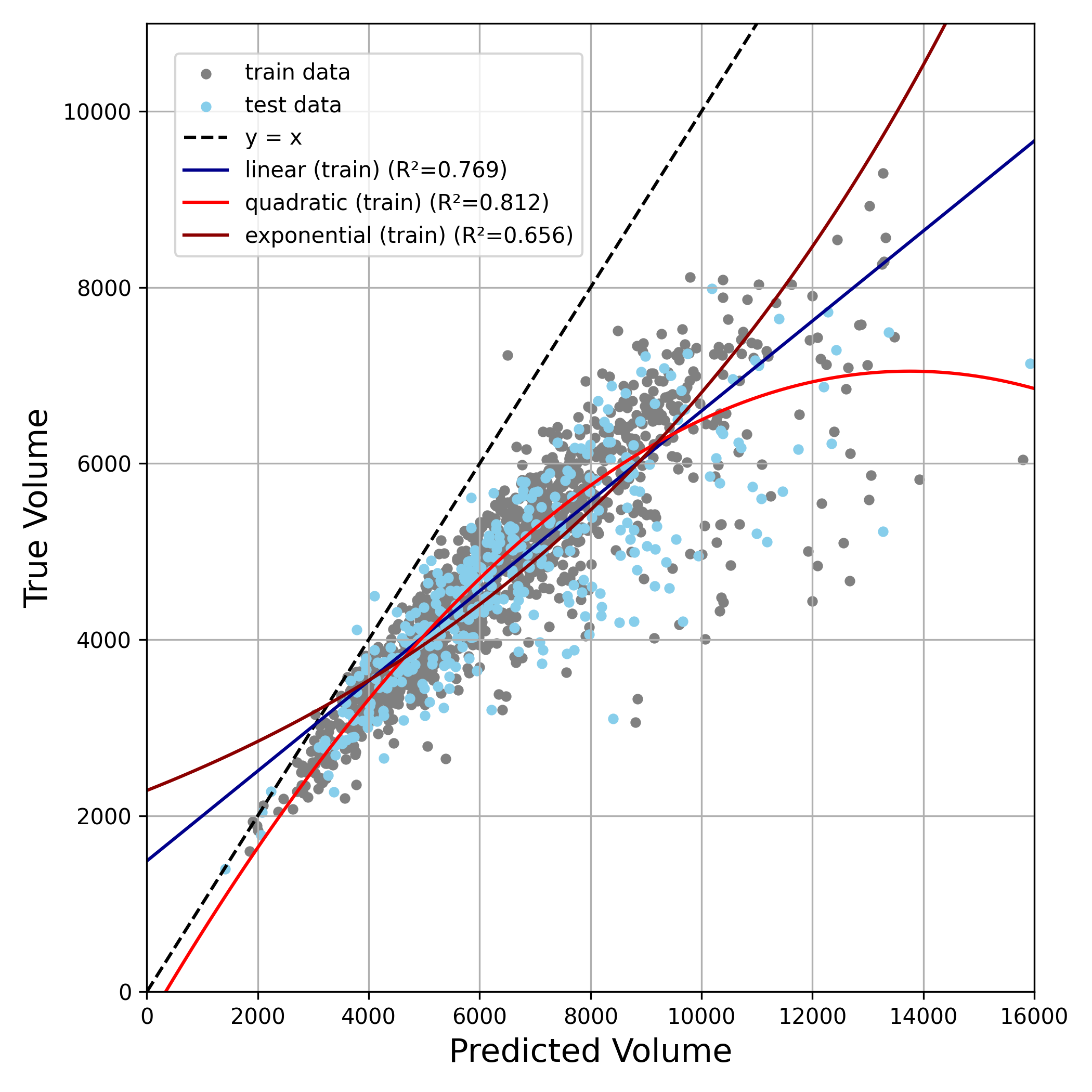}
            \caption{}
        \end{subfigure}
        \vspace{0.02\textwidth}
        \begin{subfigure}[t]{0.23\textwidth}
            \centering
            \includegraphics[width=1\linewidth]{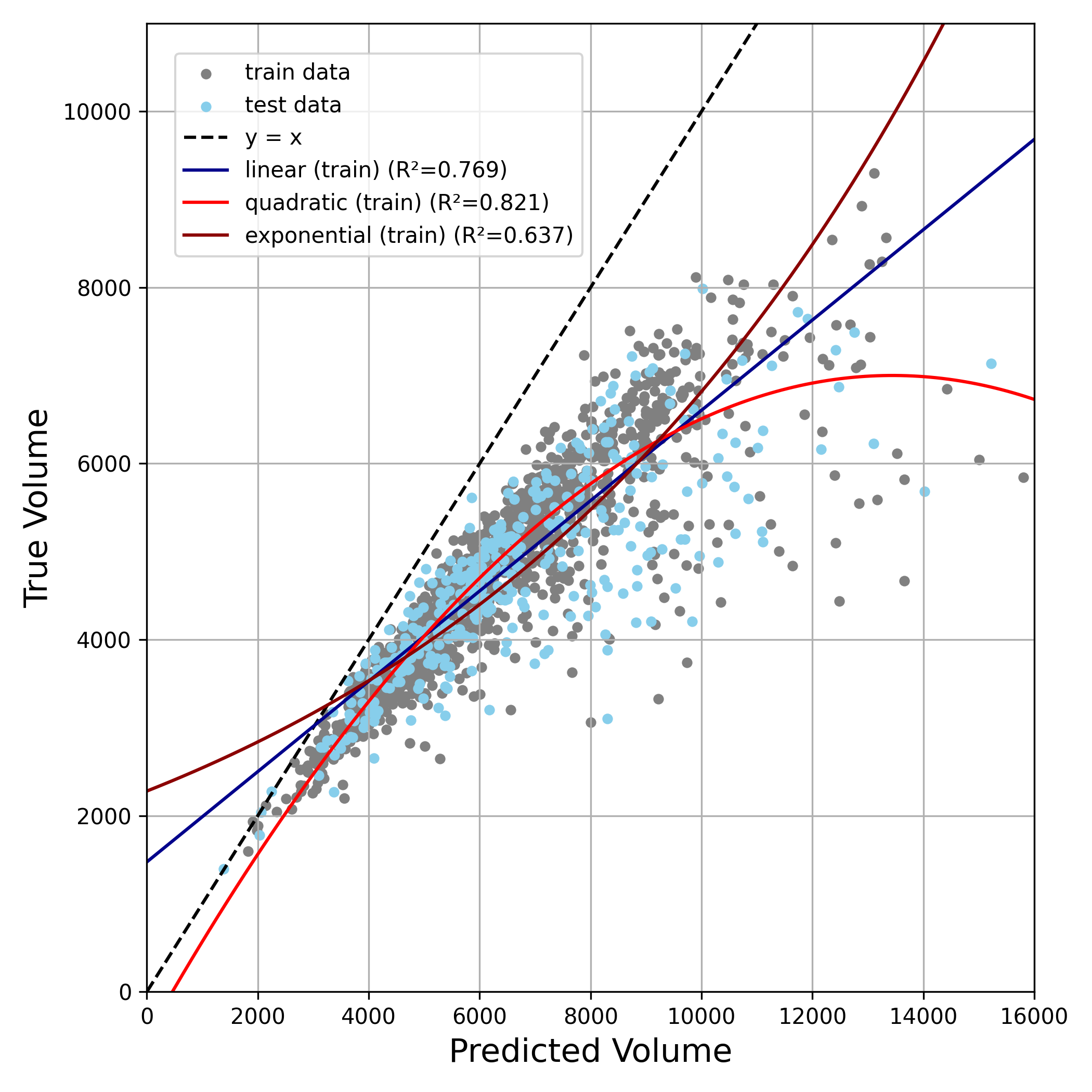}
            \caption{}
        \end{subfigure}
        \vspace{0.02\textwidth}
        \begin{subfigure}[t]{0.23\textwidth}
            \centering
            \includegraphics[width=1\linewidth]{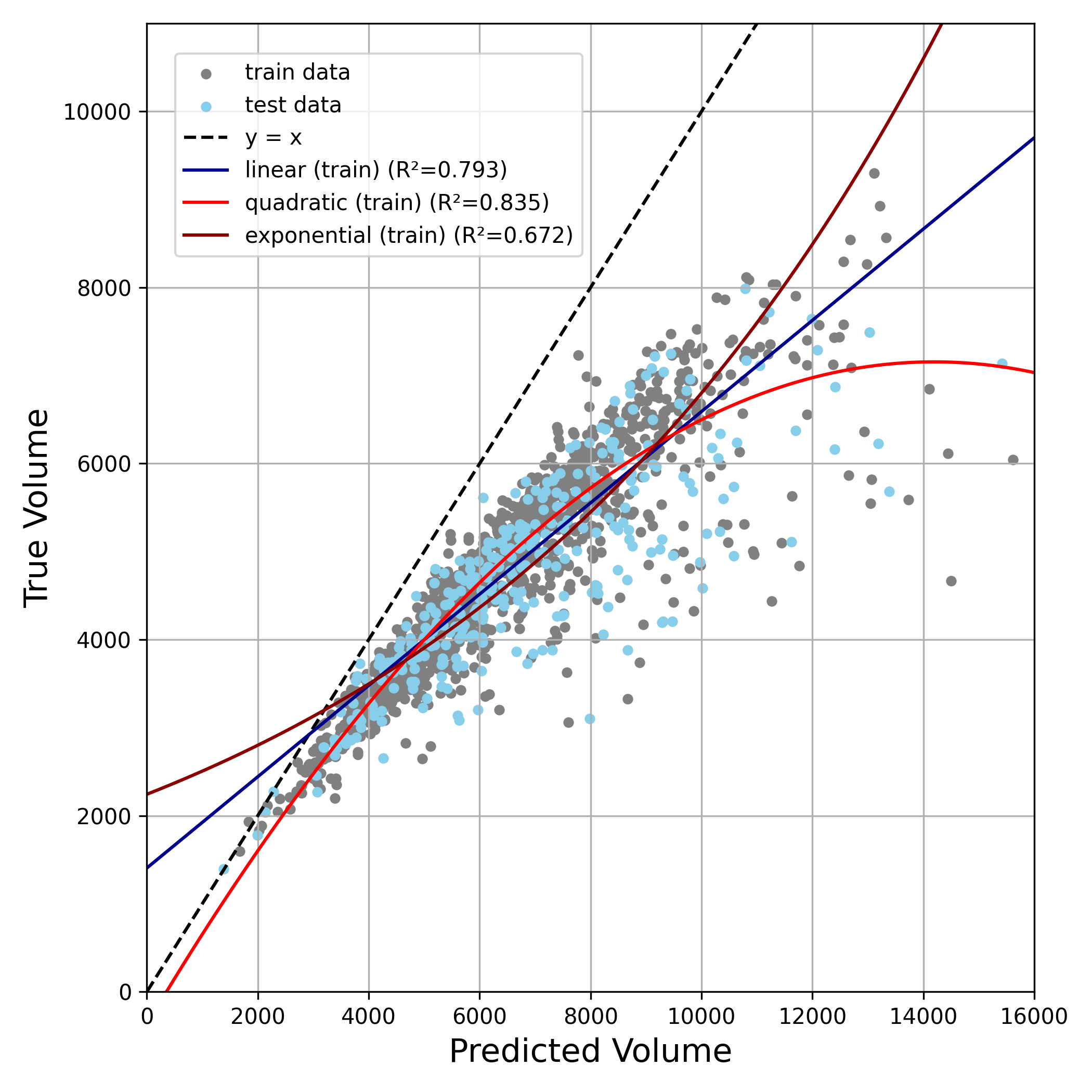}
            \caption{}
        \end{subfigure}
        \caption{Linear, quadratic, and exponential curve fitted to the estimated spikes based on the geometric baseline and the measured volume of the training dataset (grey), and their respective $R^2$ score. Test dataset (blue) was added for visualization purpose. Separate curves were fitted for estimations based on one image (a), two images (b), four images(c), and six images (d).}
        \label{fig:train_curves_geo}
    \end{figure*}

     \begin{figure*}
        \vskip 0.2in \centering{\includegraphics[width=\textwidth]{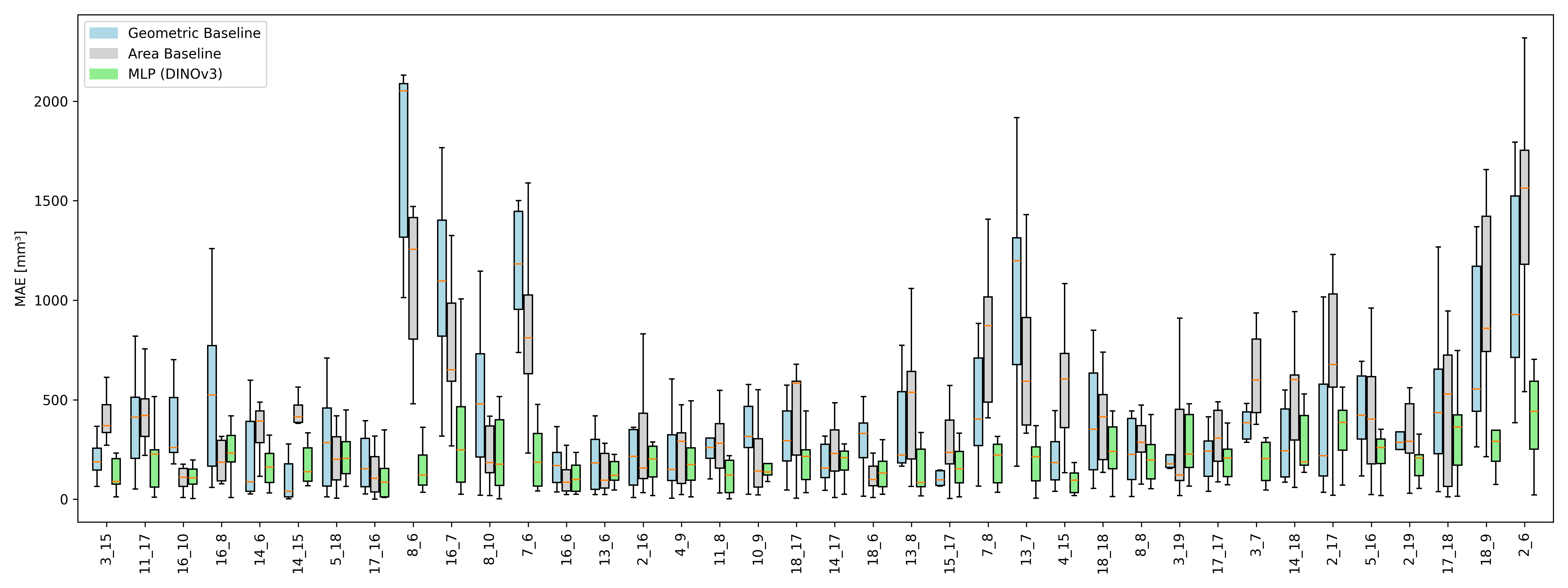}} 
        \caption{Mean absolut error across testset genotypes for the geometric and area baseline and the fine-tuned DINOv3 evaluated on six images.}
        \label{fig:genotype_error}
        \vskip -0.2in
    \end{figure*}

     \begin{figure}
        \vskip 0.2in \centering{\includegraphics[width=0.25\textwidth]{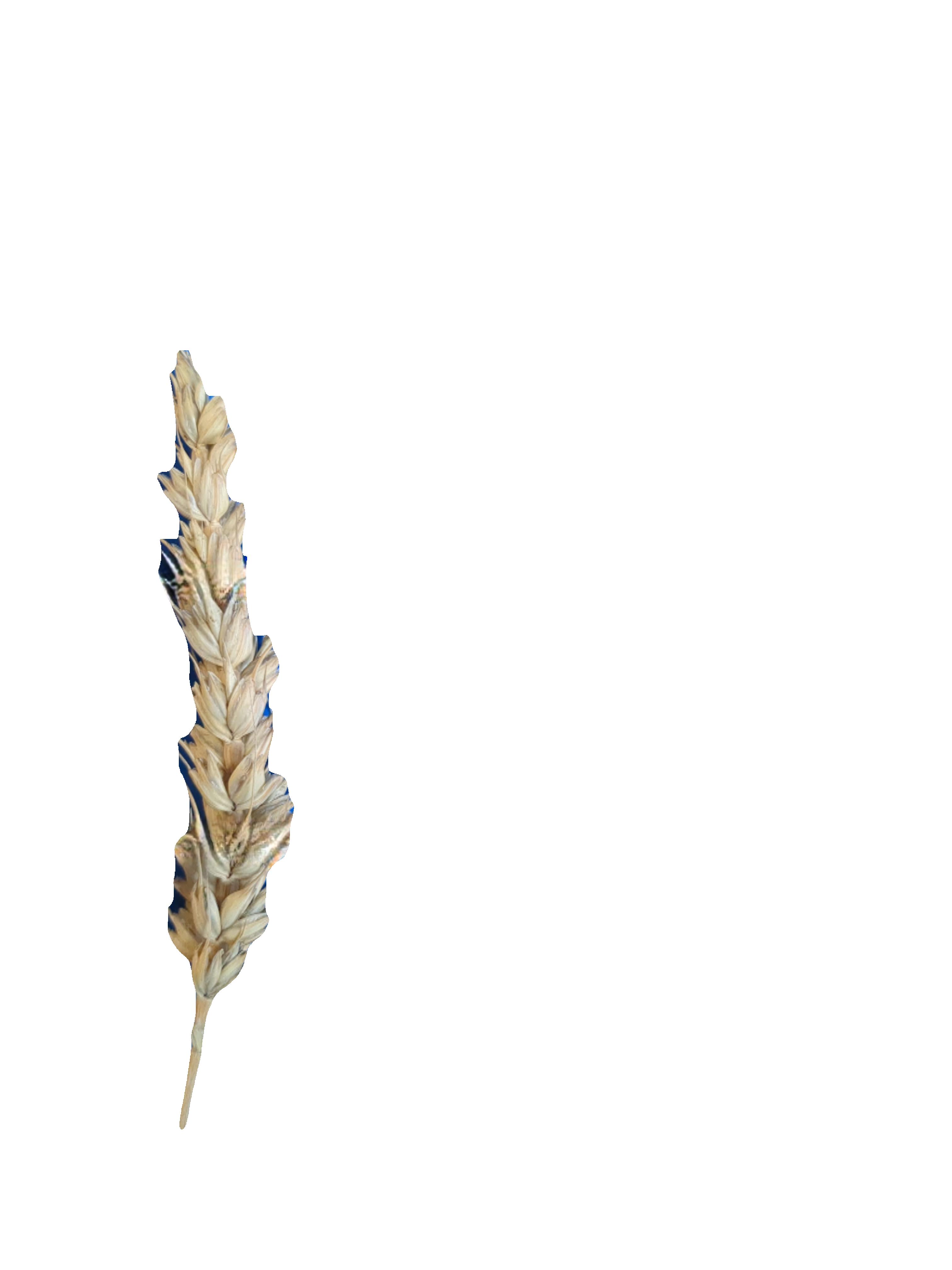}} 
        \caption{Example of a spike resulting in a high volume estimation error for the geometric baseline.}
        \label{fig:high_error_spike}
        \vskip -0.2in
    \end{figure}

\end{document}